\documentclass[lettersize,journal]{IEEEtran}
\usepackage{amsmath,amsfonts}
\usepackage{algorithmic}
\usepackage{algorithm}
\usepackage{balance}
\usepackage{array}
\usepackage[caption=false,font=normalsize,labelfont=sf,textfont=sf]{subfig}
\usepackage{textcomp}
\usepackage{stfloats}
\usepackage{url}
\usepackage{verbatim}
\usepackage{graphicx}
\usepackage{cite}
\usepackage{bigstrut}
\usepackage{array}
\usepackage{multirow} 
\usepackage{tabularx} 
\usepackage{tabu}
\usepackage{tabularx}
 \usepackage{booktabs}
 \usepackage{rotating}
 \usepackage[algo2e]{algorithm2e} 
\usepackage[subfigure]{tocloft}
\usepackage{tikz,xcolor}
\usepackage{bbding}
\usepackage{color}
 \usepackage{threeparttable}
\usepackage[implicit=false]{hyperref}
\hypersetup{hidelinks,
	colorlinks=true,
	allcolors=black,
	pdfstartview=Fit,
	breaklinks=true}
\definecolor{lime}{HTML}{A6CE39}
\DeclareRobustCommand{\orcidicon}{
\begin{tikzpicture}
\draw[lime, fill=lime] (0,0)
circle[radius=0.16]
node[white]{{\fontfamily{qag}\selectfont \tiny \.{I}D}};
\end{tikzpicture}
\hspace{-2mm}
}
\foreach \x in {A, ..., Z}{%
\expandafter\xdef\csname orcid\x\endcsname{\noexpand\href{https://orcid.org/\csname orcidauthor\x\endcsname}{\noexpand\orcidicon}}
}

\begin{document}

\title{Unifying Lane-Level Traffic Prediction from a Graph Structural Perspective: Benchmark and Baseline}

\author{
\thanks{Manuscript received August 3, 2024; revised April 25, 2025; accepted June 9, 2025.  This work was supported by Hong Kong Research Grants Council Grant 16202722, Grant T22-607/24-N, Grant T43-513/23N-1. It was partially conducted in JC STEM Lab of Data Science Foundations funded by The Hong Kong Jockey Club Charities Trust. (Corresponding author: Weidong Yang.)}
Shuhao Li\orcidA{}, ~\IEEEmembership{Student Member, ~IEEE}, Yue Cui\orcidB{}, Jingyi Xu, Libin Li, Lingkai Meng, Weidong Yang\orcidD{}, Fan Zhang, and Xiaofang Zhou\orcidC{},~\IEEEmembership{Fellow,~IEEE}
\thanks{Shuhao Li, Jingyi Xu and Weidong Yang are with the School of Computer Science, Fudan University, Shanghai 200082, China (shli23@m.fudan.edu.cn, jyxu22@m.fudan.edu.cn, wdyang@fudan.edu.cn)

Yue Cui and Xiaofang Zhou are with The Hong Kong University of Science and Technology, Hong Kong SAR 999077, China (ycuias@cse.ust.hk, zxf@cse.ust.hk)

Libin Li and Fan Zhang are with Guangzhou University, Guangzhou 510006, China (lilibin@e.gzhu.edu.cn, zhangf@gzhu.edu.cn)

Lingkai Meng is with Shanghai Jiao Tong University, Shanghai 200052, China (mlk123@sjtu.edu.cn)}
}

\markboth{Journal of \LaTeX\ Class Files,~Vol.~14, No.~8, August~2021}%
{Shell \MakeLowercase{\textit{et al.}}: A Sample Article Using IEEEtran.cls for IEEE Journals}

\IEEEpubid{0000--0000/00\$00.00~\copyright~2021 IEEE}

\maketitle

\begin{abstract}
Traffic prediction has long been a focal and pivotal area in research, witnessing both significant strides from city-level to road-level predictions in recent years. With the advancement of Vehicle-to-Everything (V2X) technologies, autonomous driving, and large-scale models in the traffic domain, lane-level traffic prediction has emerged as an indispensable direction. However, further progress in this field is hindered by the absence of comprehensive and unified evaluation standards, coupled with limited public availability of data and code. In this paper, we present the first systematic classification framework for lane-level traffic prediction, offering a structured taxonomy and analysis of existing methods. We construct three representative datasets from two real-world road networks, covering both regular and irregular lane configurations, and make them publicly available to support future research. We further establishes a unified spatial topology structure and prediction task formulation, and proposes a simple yet effective baseline model, GraphMLP, based on graph structure and MLP networks. This unified framework enables consistent evaluation across datasets and modeling paradigms. We also reproduce previously unavailable code from existing studies and conduct extensive experiments to assess a range of models in terms of accuracy, efficiency, and applicability, providing the first benchmark that jointly considers predictive performance and training cost for lane-level traffic scenarios. All datasets and code are released at https://github.com/ShuhaoLii/LaneLevel-Traffic-Benchmark.
\end{abstract}


\section{Introduction}
\IEEEPARstart{L}ane-level traffic prediction has gradually become an essential component of multi-granularity traffic prediction with the rapid development of intelligent vehicles and refined urban management. In the hierarchical structure of traffic prediction, city-level studies primarily address large-scale urban transportation issues using methods such as transfer learning \cite{ouyang2023citytrans,lu2022spatio} and multi-task learning \cite{zhang2021multi}, providing foundational support for urban planning and smart city infrastructure \cite{zhou2020riskoracle}. Regional-level prediction focuses on enhancing the understanding of traffic patterns in specific areas \cite{cui2023roi,ji2023spatio}, such as analyzing traffic flow, travel time, and vehicle types in commercial, residential, or industrial zones \cite{pan2019urban}. Road-level traffic prediction is one of the key focuses of current research. Its core lies in leveraging models such as graph convolutional networks (GCNs) \cite{zhang2023automated,jiang2023spatio,li2021spatial}, recurrent neural networks (RNNs) \cite{li2017diffusion}, and attention mechanisms to capture the spatio-temporal dependencies at the road level \cite{yan2021learning,xu2020spatial}. This line of research aims to accurately predict traffic states, alleviate congestion, and identify high-risk accident zones.

\begin{figure}[t]
	\centering
		\includegraphics[width=1\linewidth]{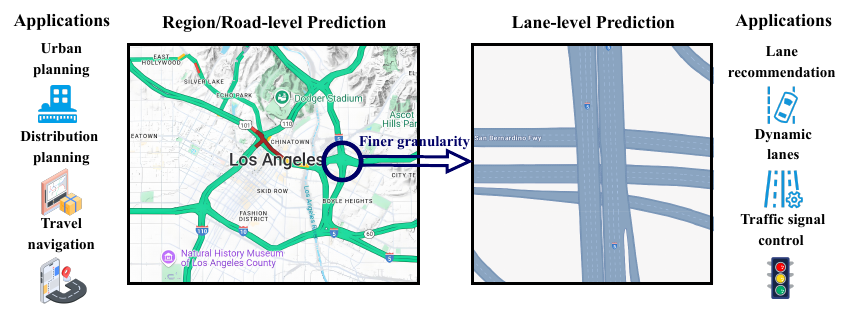}
  \vspace{-0.25cm}
  \caption{\label{fig:intro_mov}Examples of comparisons between regional/road-level and lane-level traffic prediction and their supported application scenarios: lane-level traffic prediction plays a critical role in more refined traffic management and applications.}
  \vspace{-0.35cm}
\end{figure}

As illustrated in \textbf{Figure~\ref{fig:intro_mov}}, lane-level traffic prediction further refines the research perspective by concentrating on real-time monitoring and prediction of parameters at the single-lane level, such as traffic flow and vehicle speed. Compared to other levels, lane-level prediction requires processing higher-frequency and more granular data while simultaneously considering vehicle interactions and lane-changing behaviors. Research at this level is particularly significant for optimizing intelligent transportation systems \cite{tian2018connected,song2017enhancing}. For example, it can assist autonomous vehicles in selecting lanes with lower traffic volumes, reduce the need for frequent lane changes by freight vehicles, and improve traffic safety. Additionally, it provides critical technical support for the dynamic adjustment of traffic signals and flexible management of reversible lanes.

\IEEEpubidadjcol

Despite having a relatively long research history, lane-level traffic prediction remains underdeveloped compared to other granularities of traffic prediction. Existing research findings are relatively fragmented, resembling the early development of road-level traffic prediction. Initial studies primarily relied on basic statistical methods and simple flow models. With technological advancements, more complex machine learning, and deep learning methods have gradually been introduced into the research. However, the application of these methods still faces numerous challenges. For instance, issues such as inconsistent lane counts and complex multi-branch intersections on urban expressways significantly increase modeling difficulty.

Although lane-level traffic prediction has accumulated some research, its development has significantly lagged behind that of other granularities of traffic prediction. Existing findings are relatively scattered, following a research trajectory similar to the early stages of road-level traffic prediction. Early studies primarily relied on basic statistical methods and simple flow models \cite{clark2003traffic,guo2014adaptive}, which, while computationally inexpensive, struggled to effectively capture the complex interactions and dynamic characteristics between lanes. Moreover, traditional statistical models often assumed linear traffic flow relationships, overlooking the highly nonlinear and time-varying nature of real-world traffic flows, resulting in limited prediction accuracy. With advancements in technology, more complex machine learning and deep learning methods have been gradually introduced \cite{shen2021st,zheng2022lane}. These methods leverage the strong modeling capabilities of neural networks to capture the spatio-temporal dependencies in lane-level traffic flows. However, their practical application still faces numerous challenges. For instance, inconsistencies in lane counts and the complex multi-branch layouts of urban expressways significantly increase modeling difficulty. Additionally, lane-level traffic prediction requires handling massive, high-frequency, high-dimensional data, which not only significantly raises computational costs but also imposes higher demands on data collection and processing capabilities.

\begin{table}[t]
  \centering
  \caption{The current state of public accessibility for codes and datasets in existing lane-level research studies.}
  \begin{threeparttable}
    \begin{tabular}{ccc}
    \toprule
    \textbf{Models} & \textbf{Code Available} & \textbf{Dataset Available} \\
    \midrule
    Cat-RF-LSTM \cite{zhao2022hybrid} & \XSolidBrush & \XSolidBrush \\
    CEEMDAN-XGBoost \cite{lu2020hybrid} & \XSolidBrush & \XSolidBrush \\
    STMGG \cite{zeng2022modeling} & \XSolidBrush & \XSolidBrush \\
    TM-CNN \cite{ke2019two} & \XSolidBrush & \XSolidBrush \\
    MDL \cite{lu2020lane} & \CheckmarkBold & \XSolidBrush \\
    CNN-LSTM \cite{ma2020multi} & \XSolidBrush & \XSolidBrush \\
    HGCN \cite{zhou2022lane} & \XSolidBrush & \XSolidBrush \\
    DGCN \cite{wang2021lane} & \XSolidBrush & \XSolidBrush \\
    ST-AFN\cite{shen2021st} & \CheckmarkBold & \CheckmarkBold \\
    FDL \cite{gu2019short} & \XSolidBrush & \XSolidBrush \\
    GCN-GRU \cite{li2023dynamic} & \XSolidBrush & \XSolidBrush \\
    STA-ED \cite{zheng2022lane} & \XSolidBrush & \XSolidBrush \\
    \bottomrule
    \end{tabular}%
    \begin{tablenotes}
    \item Note: 'Available' here refers to the explicit mention of accessible code or dataset URLs in their respective research papers.
    \end{tablenotes}
\end{threeparttable}
  \label{tab:available}%
  \vspace{-0.45cm}
\end{table}%

Several unresolved issues remain in this field, such as how to rapidly and accurately respond to changes in lane traffic conditions, and how to integrate lane traffic prediction across different scenarios to achieve a universal and unified solution for lane-level prediction. Furthermore, there is currently no comprehensive review or benchmark research, making it difficult for new researchers entering the field to quickly grasp its core problems and latest advancements. As shown in \textbf{Table~\ref{tab:available}}, most existing studies are based on private datasets, with relevant code and datasets not publicly available \cite{lu2020hybrid,gu2019short,ke2019two,zhou2022lane,wang2021lane,zheng2022lane,zeng2022modeling}. This lack of sharing and openness severely limits the comparison, validation, and improvement of research methods. To advance the field of lane-level traffic prediction, it is imperative to establish public standardized datasets and benchmarks and to encourage the research community to share code and data resources openly. This would not only improve the quality of research but also accelerate technological progress across the entire field.

We commence with a detailed review and classification of current research in lane-level traffic prediction. This review emphasizes the development of spatial topology and the techniques for modeling spatial-temporal dependencies. Additionally, we conduct an in-depth analysis of the advantages and limitations of these methods. Building upon this groundwork, our approach to traffic prediction incorporates a macro perspective and the application of graph structures. We effectively establish lane-level network topologies using graph structure. Utilizing this framework, we develop an advanced, efficient solution and a novel baseline model. We then extract data from three real-world lane networks, comprising two with regular lane configurations and one with irregular lane configurations. To enhance the evaluation of model performance, we introduce training cost as a new metric. This metric seeks a balance between model effectiveness and efficiency, both of which are vital in lane-level traffic prediction. It evaluates the models based on both prediction accuracy and training duration, ensuring a thorough assessment. Finally, we conduct extensive and impartial testing of existing lane-level traffic prediction models and our proposed methods using these datasets and the training cost metric. These tests confirm our model's efficacy and highlight its potential in practical scenarios, particularly in managing complex traffic patterns and enabling efficient real-time predictions, as well as identifying critical lane segments, recurrent congestion points, and anomalous traffic patterns through fine-grained modeling. Our research contributes novel insights and tools to the field of lane-level traffic prediction, establishing a robust foundation and direction for future scholarly and practical endeavors.

In summary, our main contributions can be outlined as follows:
\begin{itemize} 
\item We introduce the first systematic classification framework for lane-level traffic prediction and provide a comprehensive analysis of the technical characteristics, advantages, and limitations of existing methods. 
\item We propose a unified topology modeling approach, clarify the tasks and challenges of lane-level traffic prediction, and design a simple yet effective baseline model to address these challenges. 
\item We extract and construct three reliable datasets from two real-world road networks, including datasets with regular and irregular lane counts, to enhance the diversity of testing scenarios. These datasets are made publicly available to the research community. 
\item We reproduce and release previously unavailable code from existing studies, conduct extensive and fair testing of both existing models and the proposed baseline model, and evaluate their performance using dual metrics of effectiveness and efficiency. 
\end{itemize}

The rest of this paper is organized as follows: In \textbf{Section~\ref{section 2}}, we provide a comprehensive classification and analysis of existing work based on spatial topology construction and spatio-temporal dependency modeling. Subsequently, \textbf{Section~\ref{section 3}} defines a unified problem of lane-level traffic prediction. Motivated by these challenges, details of our proposed unified graph-structured approach and a simple baseline are introduced in \textbf{Section~\ref{section 4}}. Next, in \textbf{Section~\ref{section 5}}, we evaluate our model on three datasets using multiple metrics, presenting a benchmark for lane-level traffic prediction, including studies on effectiveness and efficiency, as well as long-term predictive performance. Finally, we conclude the paper and discuss the future implications of lane-level traffic prediction in \textbf{Section~\ref{section 6}}.

\begin{figure}[t]
	\centering
		\includegraphics[width=1\linewidth]{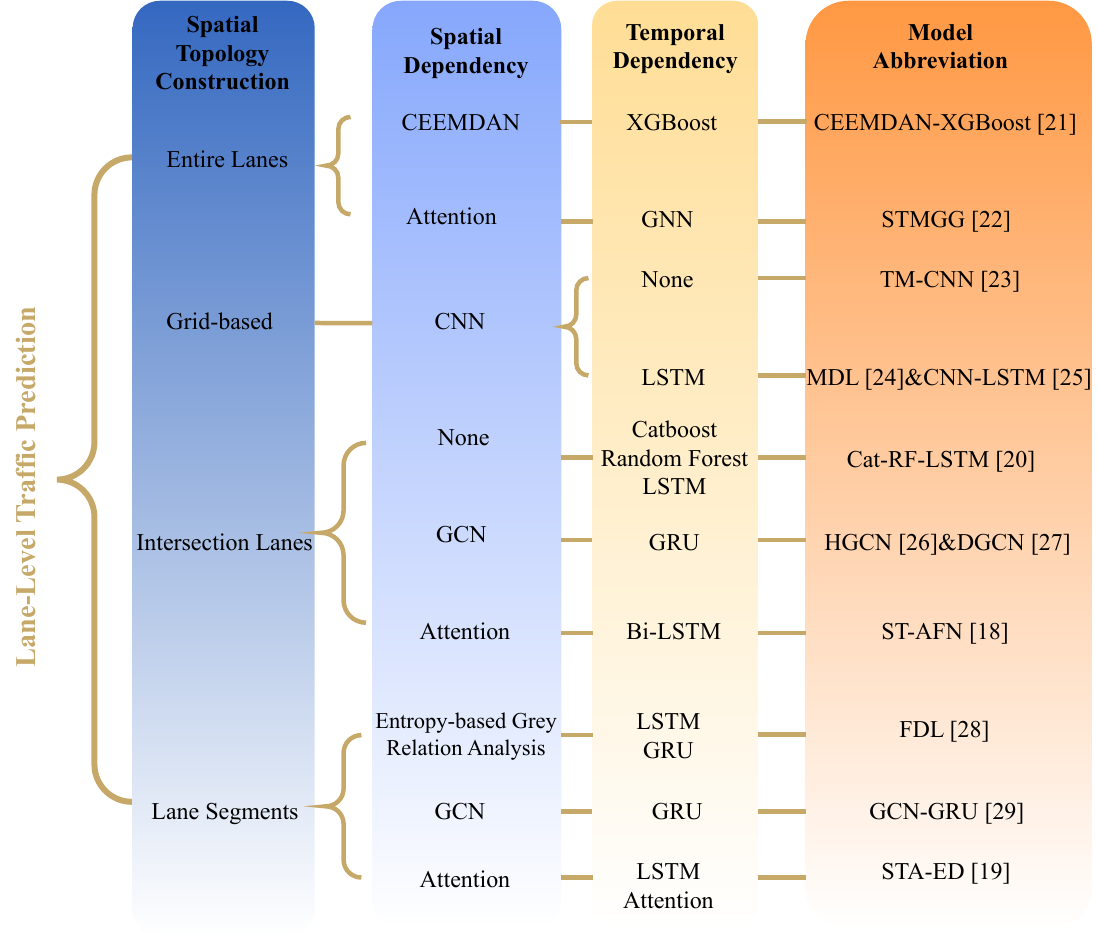}
  \caption{\label{fig:survey}Categorization of existing lane-level traffic prediction research based on spatial topology construction, spatial dependency modeling techniques, and temporal dependency modeling techniques.}
  \vspace{-0.35cm}
\end{figure}

\section{Literature Review and Classification} \label{section 2}
Lane-level traffic prediction is an integral part of intelligent transportation systems. The challenge in this field lies in accurately capturing and understanding the complex spatio-temporal dependencies in traffic flow. Spatial dependencies refer to how traffic flow on different roads or lanes affects each other, while temporal dependencies focus on the patterns of traffic flow over time. These models usually need to predict traffic volume accurately in highly dynamic and multi-variable urban traffic networks, requiring not only the capture of complex traffic dynamics but also a rapid response to real-time changes in traffic conditions. As illustrated in \textbf{Figure~\ref{fig:survey}}, based on existing research, we categorize the modeling approaches for spatial topology (modeling objectives), spatial dependency modeling techniques, and temporal dependency modeling techniques. Through \textbf{Table~\ref{tab:comparison}}, we provide a detailed analysis of the core principles, advantages, and limitations of various modeling techniques, and we evaluate the strengths and weaknesses of existing research from three dimensions: flexibility, robustness, and efficiency. Specifically, flexibility assesses the applicability and transferability of the model, robustness evaluates the model's ability to handle high dynamics and sudden changes, and efficiency focuses on the technical implementation and the scale of training parameters. Our goal is to offer a comprehensive perspective on the current state and future development directions of lane-level traffic prediction.

\begin{table}[t]
  \centering
  \caption{Overall comparison of the strengths and weaknesses of existing lane-level traffic prediction models across three dimensions.}
    \begin{tabular}{cccc}
    \toprule
    \textbf{Models} & \textbf{Flexibility} & \textbf{Robustness} & \textbf{Efficiency} \\
    \midrule
    Cat-RF-LSTM \cite{zhao2022hybrid} & \XSolidBrush & \XSolidBrush & \CheckmarkBold \\
    CEEMDAN-XGBoost \cite{lu2020hybrid} & \XSolidBrush & \XSolidBrush & \CheckmarkBold \\
    FDL \cite{gu2019short} & \CheckmarkBold & \XSolidBrush & \CheckmarkBold \\
    STMGG \cite{zeng2022modeling} & \XSolidBrush & \CheckmarkBold & \XSolidBrush \\
    TM-CNN \cite{ke2019two} & \XSolidBrush & \XSolidBrush & \CheckmarkBold \\
    MDL \cite{lu2020lane} & \XSolidBrush & \XSolidBrush & \XSolidBrush \\
    CNN-LSTM \cite{ma2020multi} & \XSolidBrush & \XSolidBrush & \XSolidBrush \\
    HGCN \cite{zhou2022lane} & \XSolidBrush & \CheckmarkBold & \CheckmarkBold \\
    DGCN \cite{wang2021lane} & \XSolidBrush & \CheckmarkBold & \CheckmarkBold \\
    ST-AFN\cite{shen2021st} & \XSolidBrush & \CheckmarkBold & \XSolidBrush \\
    GCN-GRU \cite{li2023dynamic} & \CheckmarkBold & \CheckmarkBold & \XSolidBrush \\
    STA-ED \cite{zheng2022lane} & \CheckmarkBold & \CheckmarkBold & \XSolidBrush \\
    \bottomrule
    \end{tabular}%
  \label{tab:comparison}%
\end{table}%

\subsection{Spatial Topology Construction}
The construction of spatial topology determines the modeling objectives and is a critical factor influencing the modeling of spatio-temporal dependencies and prediction accuracy. Compared to coarse-grained traffic prediction at the road or city level, lane-level traffic prediction requires capturing finer-grained dynamic data, significantly increasing the complexity and diversity of modeling targets. To systematically summarize existing research, we classified lane-level traffic prediction modeling objects into four main types based on different modeling requirements and application scenarios, as shown in \textbf{Figure \ref{fig:Topology}}: Entire Lanes, Grid-based Models, Intersection Lanes, and Lane Segments. Below is a detailed explanation and analysis of the characteristics of each type:
\begin{figure*}[htbp]
\vspace{-0.45cm}
	\centering
		\includegraphics[width=1\linewidth]{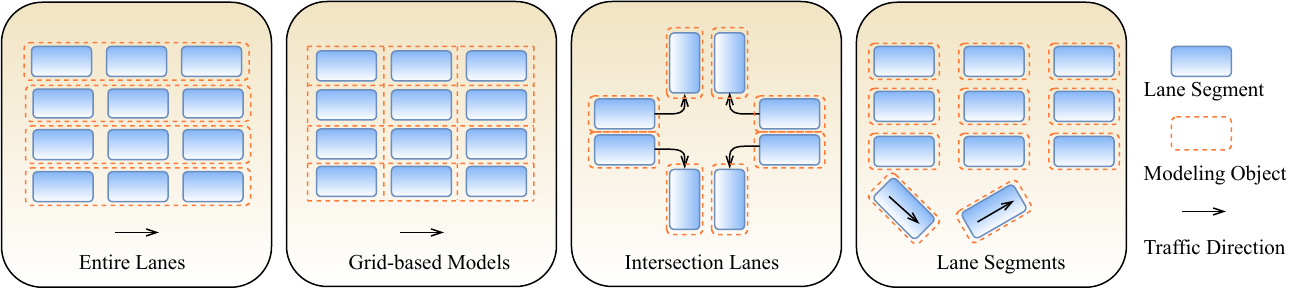}
  \vspace{-0.25cm}
  \caption{\label{fig:Topology}Four types of spatial topological structures in lane-level modeling, where the longer edges of lane segments are adjacent to neighboring lane segments, and the shorter edges correspond to the preceding and following lane segments.}
  \vspace{-0.35cm}
\end{figure*}

\subsubsection{Entire Lanes} The entire lane modeling approach treats each lane as a whole modeling object. Input data typically include overall lane-level traffic states, such as traffic flow, average speed, and lane density. The output targets are traffic predictions for each lane, such as future total traffic flow or average speed \cite{lu2020hybrid,zeng2022modeling}. A key advantage of this method is its simplicity: the number of modeling objects corresponds directly to the number of lanes, significantly reducing model complexity and improving computational efficiency. This approach is particularly suitable for scenarios with uniform traffic characteristics and minimal inter-lane interactions, such as unidirectional lanes on highways. However, it fails to capture interactions between lanes, making it less effective for complex segments like intersections or densely interactive urban roads. Consequently, its ability to model overall road network traffic patterns is limited.

\subsubsection{Grid-based Models} Grid-based methods divide lanes into fixed-sized uniform grids, treating each grid as a unit akin to an image pixel, allowing CNNs to extract local spatial features  \cite{ke2019two,lu2020lane,ma2020multi}. Input data typically includes traffic state information within grids, such as traffic density and local speed, while output targets are traffic predictions for each grid, such as future traffic flow or congestion levels. These methods perform well in simple road structures, effectively capturing dependencies between adjacent grid units. However, fixed grid divisions cannot adapt to roads with varying numbers of lanes, such as those with entrances or exits. This limitation reduces their effectiveness in dynamically changing traffic scenarios.

\subsubsection{Intersection Lanes} Intersection modeling methods focus on the most complex areas of traffic flow and serve as a crucial approach for modeling key traffic nodes. In both road-level and lane-level traffic prediction, intersections remain one of the most challenging research subjects. Road-level intersection prediction typically relies on aggregated road-level traffic flow information, incorporating overall traffic movement patterns or leveraging traffic signal control to enhance the accuracy of regional traffic forecasts \cite{li2020real, xu2023signal}. In contrast, lane-level intersection modeling requires more granular input data, including lane-specific traffic volume, speed, vehicle turning behavior, and traffic signal control states. The output targets encompass lane-level traffic distribution prediction, traffic signal optimization parameter computation, and congestion evaluation \cite{zhao2022hybrid,zhou2022lane,wang2021lane,shen2021st}. These methods play a significant role in optimizing traffic signal control, alleviating congestion, and enhancing driving safety. However, most current lane-level studies still predominantly depend on road-level datasets, failing to fully capture the fine-grained dynamic characteristics of intersections. The lack of high-resolution data and the limited adaptability of models in capturing cross-lane traffic flows constrain prediction accuracy, thereby affecting their applicability in dynamic and evolving traffic environments.

\subsubsection{Lane Segments} 
Lane segment modeling divides each lane into multiple segments, treating each segment as an independent modeling unit, providing a more flexible spatial topology for complex traffic scenarios \cite{gu2019short,li2023dynamic,zheng2022lane}. Input data typically include segment-level metrics such as flow, speed, density, and interaction features between adjacent segments. Output targets include segment-level flow predictions, congestion trends, or dynamic relationships between segments. This method is especially suited to multi-entry and multi-exit roads and other complex structures, capturing finer-grained dynamics. For example, in intersection traffic modeling, each lane segment can serve as an independent modeling object, significantly improving adaptability to dynamic scenarios. However, the increased number of modeling objects imposes higher computational demands and requirements for spatial dependency modeling techniques, potentially resulting in high computational costs in high-frequency, real-time traffic predictions.

\subsubsection{Comprehensive Analysis} Each existing spatial topology modeling method has its specific advantages and use cases, but also significant limitations. The entire lane modeling approach is simple and efficient, reducing model complexity and improving computational efficiency. However, it fails to effectively capture interactions between lanes, making it less suitable for complex traffic scenarios. Grid-based methods perform well in extracting local spatial features and are effective in simple road structures. Nevertheless, their reliance on fixed grid divisions limits their ability to adapt to dynamic topological changes, such as roads with varying lane counts. Intersection lane modeling provides a targeted approach for predicting traffic in critical areas with complex flows, such as intersections. While it enhances prediction accuracy and traffic signal optimization, its reliance on coarse data and limited ability to model fine-grained cross-lane interactions restricts its broader applicability. Lane segment modeling, on the other hand, excels at capturing detailed dynamics and adapting to complex road structures, making it particularly effective in scenarios with multi-entry and multi-exit roads. However, the significantly increased number of modeling objects leads to higher computational demands, limiting its feasibility for high-frequency, real-time applications.

These limitations underscore the need for a unified spatial topology structure that can accommodate diverse scenarios and flexibly adapt to dynamic changes. Such a structure should integrate various modeling objects, including entire lanes, intersection lanes, and lane segments, within a single framework to enhance the generalizability and practicality of lane-level traffic prediction models.

\subsection{Spatial Dependency Modeling}
Spatial dependency modeling is a crucial component of lane-level traffic prediction, aiming to accurately capture the complex interactions between lanes and road segments. Compared to traditional road-level prediction, lane-level traffic prediction deals with higher-resolution dynamic characteristics and diverse traffic scenarios. The design of efficient spatial dependency modeling techniques directly determines the performance of prediction models. Based on their technical principles, existing spatial dependency modeling methods can be broadly categorized into the following types: methods without specific spatial modeling \cite{zhao2022hybrid}, traditional methods based on signal processing \cite{lu2020hybrid,gu2019short}, and advanced methods based on deep learning \cite{ke2019two,lu2020lane,ma2020multi,zhou2022lane,wang2021lane,li2023dynamic,shen2021st,zheng2022lane,zeng2022modeling}.

\subsubsection{Methods Without Specific Spatial Modeling}
These methods do not explicitly model the topology of the traffic network, treating lane-level flow prediction as a time series problem. Regression models (e.g., linear regression, support vector machines) or tree-based models (e.g., random forests, gradient boosting trees) are commonly employed to model lane flow directly based on historical data \cite{zhao2022hybrid}. While computationally efficient, these methods rely solely on historical data from individual lanes and neglect spatial dependencies between lanes. Consequently, they perform poorly in capturing local interactions and global topological relationships. Their predictive performance is particularly limited in complex scenarios, such as multi-lane interactions or competitive flows at intersections.

\subsubsection{Signal Processing Techniques}Traditional methods based on signal processing use mathematical decomposition and statistical analysis to extract local spatial dependencies between lanes. For instance, the CEEMDAN-XGBoost model applies Complete Ensemble Empirical Mode Decomposition with Adaptive Noise (CEEMDAN) to decompose lane flow data into components of different frequencies for feature extraction and prediction \cite{lu2020hybrid}. Similarly, the FDL model utilizes entropy-based grey relational analysis to identify lane segments highly correlated with the target lane flow\cite{gu2019short}.

These methods excel at analyzing local lane associations and perform well in structured scenarios like unidirectional highways. However, they rely on static feature extraction and lack adaptability to dynamic topological changes. Moreover, their inability to capture global dependencies in complex road networks—such as those with multiple entrances or dynamically changing lane topologies—limits their applicability in more complex scenarios.

\subsubsection{Deep Learning-Based Methods}
Deep learning models adopt data-driven approaches to learn complex nonlinear relationships in lane-level traffic flows, significantly enhancing predictive performance. These methods can be further divided into the following categories:

\noindent\textbf{Convolutional Methods:} CNNs and GCNs are two mainstream approaches for spatial modeling. CNNs excel at extracting local features from regular grid structures, while GCNs flexibly handle non-Euclidean structures like lane networks. For example, the TM-CNN model partitions lane flow data into uniform grids and uses CNNs to extract local dependencies \cite{ke2019two}, while GCN models leveraging data-driven adjacency matrices capture global spatial dependencies between lanes \cite{zhou2022lane,wang2021lane,li2023dynamic,li2024seeing,li2024st}.

While CNN-based methods perform well in modeling interactions among adjacent lanes, they assume a regular grid structure, limiting their flexibility. On the other hand, GCNs can adapt to complex topologies but incur high computational costs during matrix operations. Their efficiency and adaptability in multi-entry or irregular topological scenarios still require improvement.

\noindent\textbf{Hybrid Deep Learning Methods:} Hybrid models integrate multiple network structures (e.g., convolutional layers, recurrent neural networks, dense layers) to capture both spatio-temporal dependencies. For instance, the MDL model combines ConvLSTM and dense layers to extract temporal dynamics and spatial dependencies through convolutional operations \cite{lu2020lane,ma2020multi}. These models are well-suited for modeling multi-lane interactions in complex road networks. However, their high computational complexity and resource demands make them less suitable for real-time applications with stringent latency requirements.

\noindent\textbf{Attention-based Methods:} Attention mechanisms dynamically assign weights to emphasize important spatio-temporal relationships\cite{zheng2022lane,zeng2022modeling}. For example, the ST-AFN model combines spatial and temporal attention mechanisms to flexibly capture complex dynamic interactions between lanes \cite{shen2021st}. These methods excel in highlighting critical interactions, such as lane-changing behaviors or entry flows. However, introducing attention mechanisms significantly increases model complexity, posing challenges for model training and parameter tuning.

\subsubsection{Comprehensive Analysis}
The above classification reveals the strengths and limitations of various approaches to spatial dependency modeling. Methods without specific spatial modeling fail to capture topological relationships between lanes, limiting their ability to model complex interactions. Signal processing techniques are effective for local dependency modeling but struggle in dynamic scenarios. Deep learning-based models address these shortcomings with data-driven adaptability, achieving higher performance in complex lane networks. Within deep learning-based approaches, differences in effectiveness are also apparent. Convolution-based and graph convolution-based methods excel at capturing local and global dependencies but require improved adaptability for dynamic scenarios. Hybrid models offer comprehensive spatio-temporal feature extraction but are computationally expensive. Attention mechanism-based methods enhance the focus on critical dependencies but face limitations in scalability due to their complexity.

To address these challenges, future research should aim to balance modeling flexibility, computational efficiency, and real-time adaptability, ensuring robust performance across diverse traffic scenarios.

\subsection{Temporal Dependency Modeling}
Temporal dependency modeling is another crucial component of lane-level traffic prediction, aiming to capture the dynamic evolution of traffic flow over time. This process involves characterizing not only short-term fluctuations but also long-term temporal dependencies and potential nonlinear variations. Existing methods for temporal modeling differ significantly in their technical approaches and can be broadly categorized into machine learning-based methods, RNN-based deep learning methods (including attention-enhanced variants), and emerging GNN-based methods. Below, these approaches are summarized and evaluated from the perspectives of theoretical implementation and practical effectiveness.

\subsubsection{Machine Learning-based Methods} Machine learning-based methods rely on traditional statistical and machine learning techniques to predict traffic flow by extracting temporal features (e.g., time-of-day effects, trends, and periodicity) through feature engineering. Common approaches include CatBoost, Random Forest (RF), and Gradient Boosting Trees (XGBoost)\cite{zhao2022hybrid,lu2020hybrid}. Compared to deep learning models, these methods excel in computational efficiency and model interpretability, making them suitable for scenarios with small datasets or relatively simple temporal dependencies. For instance, analyzing the importance of features enables a clear explanation of the time-driven factors that influence traffic flow.

However, these methods are constrained by their linear assumptions and limited feature representation capabilities. They struggle to handle complex temporal dependencies and nonlinear dynamic variations, particularly in lane-level traffic prediction. These limitations make them inadequate for capturing long-term correlations in traffic flow or responding effectively to high-dynamic changes caused by sudden events, such as traffic accidents.

\subsubsection{RNN-based Methods} RNNs and their extensions are core technologies in time series modeling and have been widely applied in lane-level traffic prediction in recent years. Long Short-Term Memory (LSTM) networks and Gated Recurrent Units (GRU) effectively capture both short-term and long-term dependencies in time series through the use of gating mechanisms\cite{ke2019two,lu2020lane,ma2020multi,gu2019short,zhou2022lane,wang2021lane,li2023dynamic}. For example, Bidirectional LSTM (Bi-LSTM) enhances the modeling of traffic flow dynamics by considering both historical and future time segments via bidirectional information flow\cite{zheng2022lane}.

Attention-enhanced temporal modeling methods further improve model flexibility by dynamically assigning weights to focus on the most critical time segments for prediction. These approaches are particularly effective in capturing traffic surges caused by short-term events, such as traffic signal changes. However, these methods face several challenges. First, the sequential training mechanism of RNNs and their extensions results in lower computational efficiency, especially for long-time series, making the training process time-intensive. Second, the increased model complexity raises resource demands and limits scalability for large-scale lane flow data. While attention mechanisms improve interpretability to some extent, the hidden states within RNNs make the overall model difficult to interpret comprehensively \cite{NEURIPS2023_ee57cd73, 10.1145/3459637.3482000}.

\subsubsection{GNN-based Methods} GNN-based methods offer an innovative approach to temporal dependency modeling. Unlike traditional time series methods, these approaches transform time series into graph structures, representing time points as graph nodes and temporal dependencies as edges between nodes. For example, the Visibility Graph technique constructs a visualized spatial relation network from time series data and leverages GCNs to learn dependency relationships\cite{zeng2022modeling}.

Compared to traditional methods, GNN-based approaches can unify spatial and temporal information, which is particularly beneficial for lane-level scenarios where both dimensions interact closely. However, the efficiency of GNN-based temporal modeling remains a concern. Transforming temporal dependencies into graph relationships may lead to suboptimal performance in some complex time series tasks compared to specialized time series models. Additionally, constructing dynamic temporal graphs requires advanced modeling techniques and significant computational resources, raising the application threshold for these methods.

\subsubsection{Comprehensive Analysis}
Temporal dependency modeling methods have distinct characteristics, and their selection typically depends on the specific prediction task and data properties. Machine learning-based methods excel in computational efficiency and interpretability but are limited in their adaptability to complex temporal dynamics. RNN-based methods, particularly those enhanced with attention mechanisms, leverage deep networks to capture intricate temporal dependencies, making them suitable for dynamic traffic data with long-term correlations. In contrast, GNN-based methods introduce a spatio-temporal unified modeling perspective through graph structures, offering a novel approach to handling complex interactions in lane-level traffic flows.

In practical applications, selecting a method requires balancing prediction accuracy, computational efficiency, and modeling complexity. Future research should focus on exploring more efficient deep learning architectures, particularly in lightweight modeling and multimodal data integration, to develop more accurate and efficient temporal dependency modeling solutions for lane-level traffic prediction.

\section{Preliminaries and Problem Formulation \label{section 3}}
Based on the aforementioned literature review, we contend that treating each lane within road segments as an independent modeling object is crucial for enhancing the universality and flexibility of lane-level traffic prediction. As shown in \textbf{Figure~\ref{fig:Topology}}, we have appropriately transformed the four spatial topology construction methods to align with lane segments, allowing for flexible usage of individual or combined lane segments to meet diverse spatial topology requirements. In this approach, the lane network is conceptualized as an undirected graph $G = (V, E, A)$, where each vertex $l_{i,j} \in V^{N}$ denotes a lane segment of the $j$th lane of the total $J_i$ lanes on the $i$th road segments of the total $I$ road segments, where $\sum_{i=0}^I J_i = N$, and each edge $e \in E$ indicates whether the lane segments is connected.$A \in \mathbf{R^{N \times N}}$ denotes the static adjacency matrix of graph $G$. Furthermore, adjacency relationships are defined by both sequential lane connections and permitted lane changes, meaning adjacent lanes are not connected where lane changes are prohibited. At intersections, connectivity is determined by valid turning movements, ensuring an accurate representation of spatial dependencies. This graph-based formulation not only facilitates accurate prediction but also enables mining interpretable patterns and dependencies across lane segments.

Given a time interval $t$, we use $x_t^{l_{i,j}}$ to represent the average traffic state of a lane segment $l_{i,j}$, Towards the whole lane network, we use a state vector, denoted by $X_t = [x^{l_{1,1}}_t,x^{l_{1,2}}_t , ..., x^{l_{I,J_I}}_t]$, to represent the traffic information of all lane segments named supplementary lane segments at the time interval $t$. 

\textbf{Problem.} Given $X$ of window size $T$, $X = \{X_1, X_2, ..., X_T\}$, represent all kinds of the historical state of all the nodes on the lane network over the past $T$ time slices, Lane-level traffic prediction task aims to predict the future lane segments sequences $\hat{Y} = \{X_H|H = t+1, ...,t+z\}$, where $z$ is the number of time intervals to be predicted, which can be represented as follows:
\begin{equation}
    f[X, G] \rightarrow \hat{Y}
\end{equation}
where \( f \) represents the model used for prediction, and \( G \) is an optional, non-essential input.

\section{Graph Construction and Simple Baseline \label{section 4}}
We propose a structured framework to model various lane networks as a unified graph structure, integrating coarse-grained (e.g., road-level) and fine-grained (e.g., lane-level) traffic prediction and modeling. This approach not only unifies the spatial topology of road-level traffic prediction with lane-level and even city-level predictions but also enhances the understanding of complex interactions within traffic networks. Utilizing this unified spatial topology, lane-level traffic prediction can be formulated as a spatiotemporal graph prediction problem. This transformation enables seamless integration and conversion of traffic prediction models across different granularities, from coarse to fine.

Graph-based models capture the nuances of traffic flow on specific lanes while maintaining a broader perspective of the overall traffic network, thereby achieving more accurate and comprehensive traffic predictions. This unified approach facilitates deeper analysis of traffic dynamics, offering valuable insights for traffic management and planning.
\begin{figure*}[htbp]
\vspace{-0.45cm}
	\centering
		\includegraphics[width=0.9\linewidth]{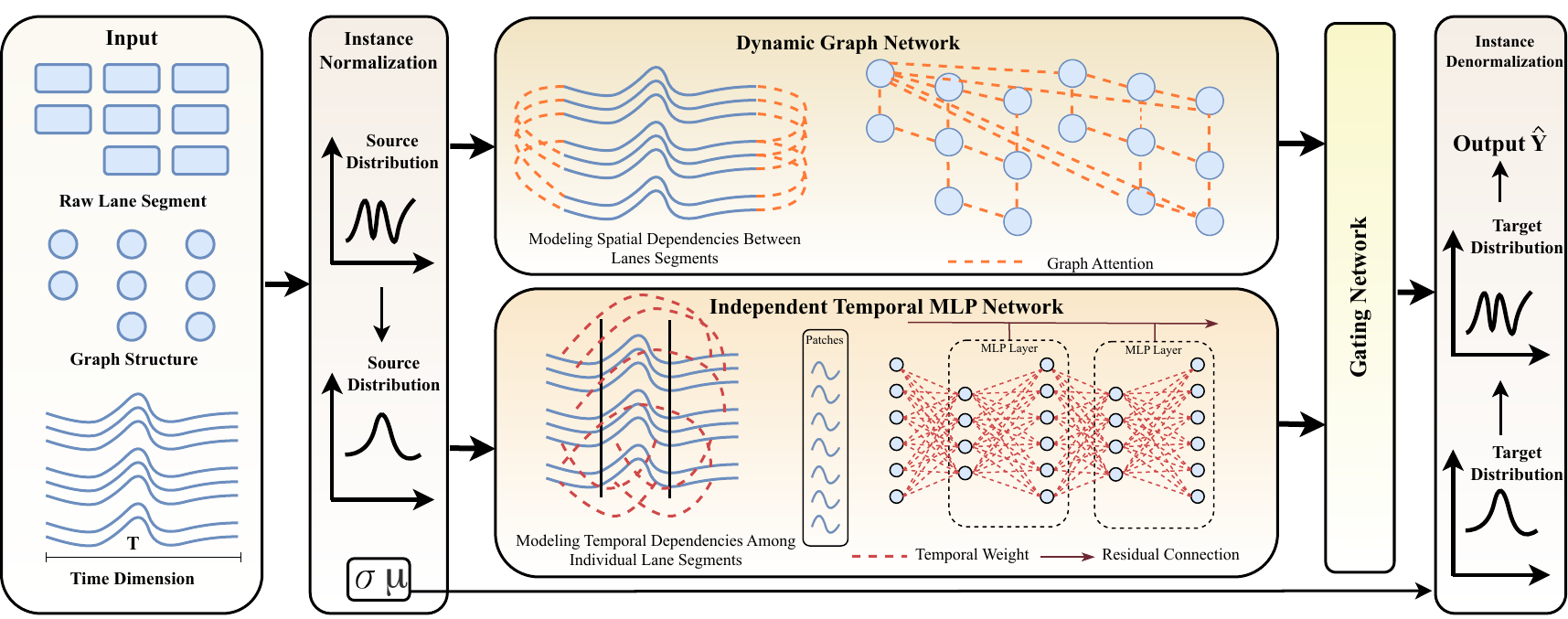}
  \vspace{-0.25cm}
    \caption{\label{fig:model_overview}Overview of GraphMLP model structure and data flow direction. $\mu$ and $\sigma$ represent the mean and standard deviation of lane segments, respectively.}
    \vspace{-0.25cm}
\end{figure*}
\subsection{Construction of Graph Structure for Lane Network}
The adoption of spatio-temporal graph-based algorithms as the mainstream approach in coarse-grained traffic prediction is primarily attributed to their ability to overcome the limitations of CNNs, which are typically confined to Euclidean spaces. spatio-temporal graph models, by contrast, are more adept at processing non-Euclidean structures, such as traffic networks, offering powerful and flexible methods to analyze and predict complex traffic flow patterns. These models provide a fresh perspective for traffic flow prediction by capturing the spatial dependencies and temporal dynamics among network nodes.

As illustrated in \textbf{Figure~\ref{fig:Topology}}, we demonstrate the segmentation of lane networks to encompass all existing topological structures studied. In exploring potential coarse-grained models that might be adaptable for fine-grained, lane-level traffic prediction, we consider the following approaches based on spatio-temporal graphs:
\subsubsection{Distance-based Graph}The construction method for distance graphs involves calculating the similarity between pairs of nodes using a distance metric modulated by a thresholded Gaussian kernel function. This process results in the formation of a weighted adjacency matrix: 
\begin{equation}
\alpha_{l_{i,j},l_{a,b}} =
\begin{cases} 
exp (- \frac{dist(l_{i,j},l_{a,b})^2}{\sigma^2}),  & \text{if } dist(l_{i,j},l_{a,b}) \leq k. \\
0, & \text{otherwise. }
\end{cases}
\end{equation}
where  $dist(l_{i,j},l_{a,b})$ denotes the lane network distance from lane segment $l_{i,j}$ to lane segment $l_{a,b}$ . $\sigma$ is the standard deviation of distances and $k$ is the threshold.

In this context, the Gaussian kernel function serves to evaluate the degree of similarity or connectivity between nodes (representing traffic entities like lane segments or intersections) based on their distances. The threshold applied ensures that only node pairs within a certain proximity influence each other significantly, thereby creating a more meaningful and structured representation of spatial relationships in the traffic network. This weighted adjacency matrix $A$ then becomes a crucial component in graph-based traffic prediction models, as it defines the network structure upon which various algorithms operate to understand and predict traffic behaviors.
\subsubsection{Binary Graph} The construction of a binary graph utilizes spatial adjacency relationships to define the adjacency matrix \( W \). Unlike road-level adjacency, which typically involves just the front and rear neighboring segments, the topology of lane-level networks is more complex. In lane-level networks, we define neighboring lanes in four directions: front, back, left, and right as follows:

\begin{equation}
\alpha_{l_{i,j},l_{a,b}} =
\begin{cases} 
1,  & \text{if adjacent. } \\
0, & \text{otherwise. }
\end{cases}
\end{equation}

This approach acknowledges the intricate nature of lane interactions within traffic systems. In urban settings, especially at intersections or multi-lane roads, the traffic flow in one lane can be significantly influenced by the adjacent lanes in all four directions. By considering these multidirectional relationships, the binary graph more accurately represents the real-world dynamics of lane-level traffic. The adjacency matrix \( A \) in this context will thus consist of binary values, indicating whether lanes are adjacent (or neighbors) in any of the four specified directions. This matrix becomes a key component in the analysis, enabling more sophisticated and realistic traffic flow modeling and prediction at the lane level.
\subsubsection{Adaptive Graph}
The predefined adjacency matrix is static and its construction methods are limited, which is a natural shortcoming in describing complex road networks. To compensate for this "lost view," the concept of an adaptive adjacency matrix has been proposed. It enables the learning of adjacency matrix parameters through the training process, thereby enhancing performance. Specifically, the adjacency relationships are determined by the learned distances between lane segment features. This approach makes lane-level traffic modeling more flexible. The formulation of this method is as follows:
\begin{equation}
    \alpha_{l_{i,j},l_{a,b}} = \text{exp}\left(-\frac{||f(x^{l_{i,j}}) - f(x^{l_{a,b}})||^2}{\sigma^2}\right)
\end{equation}
where \( f(x^{l_{i,j}}) \) and \( f(x^{l_{a,b}}) \) represent the results of processing the feature vectors \( x^{l_{i,j}} \) and \( x^{l_{a,b}} \) of lane segments \( l_{i,j} \) and \( l_{a,b} \) through the mapping function \( f \), respectively. The symbol \( \sigma \) is a tunable scale parameter used to control the degree of influence exerted by distances.

By allowing the adjacency matrix to adapt based on the data, it becomes possible to capture the dynamic and complex interactions within a traffic network more accurately. This adaptability is particularly valuable in traffic systems where the importance of certain connections may change over time due to factors like traffic conditions, roadworks, or accidents. The adaptive approach ensures that the model remains responsive to these changes, leading to more accurate predictions and analyses. This flexibility in modeling lane-level traffic provides a more nuanced and accurate representation of real-world traffic conditions.

\subsection{A Simple but Effective New Baseline}
The aforementioned graph-based modeling approach not only offers a graph-structured alternative design for existing models but also can be applied to resource-constrained edge devices. Despite its potential, it does not fully address the unique challenges of lane-level prediction. The complexity of lane-level traffic prediction is primarily manifested in more detailed and complex spatial dependencies, high demands for real-time data processing, and the need for dynamic predictive capabilities.

These challenges stem from the rapid changes in lane-level data and the need for more granular analysis. For instance, models are required to capture and analyze interactions between lanes, lane-changing behaviors, and lane-specific flow and speed patterns. Additionally, lane-level prediction necessitates that models adapt to various complex traffic scenarios, such as intersections, acceleration lanes, and exit lanes.

Given these unique challenges, we have designed a simple baseline model, GraphMLP, for lane-level traffic prediction. This model aims to provide a basic yet effective framework for capturing the essential aspects of lane-level traffic dynamics. GraphMLP leverages graph-based methods to model spatial dependencies and integrate MLP (Multilayer Perceptron) components to handle the nonlinear characteristics of traffic flow. This hybrid approach allows for accommodating the nuanced nature of lane-level traffic while maintaining computational efficiency, making it a suitable starting point for exploring more advanced lane-level traffic prediction models.

\subsubsection{GraphMLP}
Recently, various lightweight models have emerged, employing feature ID\cite{shao2022spatial}, channel independence\cite{wang2023st}, and sampling strategies\cite{zhang2022less} in the temporal dimension to reduce parameter counts and training time while achieving commendable prediction accuracy. However, these models largely overlook spatial dependency modeling.  

To address this gap, we introduce GraphMLP, a lightweight model that comprehensively addresses the accuracy and training cost issues in lane-level traffic prediction from a hybrid spatial graph and independent temporal perspective. \textbf{Figure~\ref{fig:model_overview}} illustrates the data flow and key components of GraphMLP, which consists of five critical elements: instance normalization and denormalization, dynamic graph network, independent temporal MLP network, and a gating network. Initially, input data is processed from perspectives of multivariate time series and graph structure. Then, instance normalization is applied to remove statistical information from each instance. Subsequently, both normalized data sets are fed into the dynamic graph network and the independent temporal MLP network for parallel processing, enhancing model efficiency. The outputs of both networks are aggregated through the gating network to yield a comprehensive spatio-temporal analysis result. The predictions obtained from the gating network are then subjected to instance denormalization, restoring the statistical information previously removed, to produce the final prediction output. Next, we delve into the details of the model’s components.

\subsubsection{Instance Normalization}In our task, we identified the issue of distribution shift, where time series prediction models are affected by the statistical properties of data, such as mean and variance, which can change over time. This results in a discrepancy between the distribution of training and testing data, thereby reducing prediction accuracy. To address this challenge, we have adopted and enhanced the solution proposed in \cite{kim2021reversible}, tailoring it to effectively handle lane-level data issues. Specifically, we use the mean and standard deviation of specific instances to normalize each instance 
$x^{l_{i,j}}$ of the input data. The mean and standard deviation are calculated as $\mu(x^{l_{i,j}}) = \frac{1}{T}\sum^T_{t = 1}x_t^{l_{i,j}}$ and $\sigma^2(x^{l_{i,j}}) =  \frac{1}{T}\sum^T_{t = 1}(x^{l_{i,j}}_t - \mu(x^{l_{i,j}}))^2$, respectively, allowing the model to adapt to changes in the data's statistical properties over time. Using these statistical measures, we normalize the input data $x^{l_{i,j}}$ as follows:
\begin{equation}
\dot{x}^{l_{i,j}} = \psi\big(\frac{x^{l_{i,j}}-\mu(x^{l_{i,j}})}{\sqrt{\sigma^2(x^{l_{i,j}})}}\big) + \beta
\end{equation}

where $\psi$ and $\beta$ are learnable parameters reflecting the latest data distribution. 
This normalization ensures that the input sequences have more consistent means and variances, where non-stationary information is reduced. Consequently, the normalization layer allows the model to accurately predict the local dynamics within a sequence when receiving inputs with consistent distributions of means and variances.

\subsubsection{Dynamic Graph Network}Due to the spatial complexity of lane-level road networks, the configuration of adjacent lane segments may present in multiple directions, such as up, down, left, right, and their diagonal counterparts. This complex spatial arrangement can make traditional fixed-contact adjacency matrices less flexible in capturing relationships between lanes. Therefore, to more accurately reflect the spatial relationships between lane segments, we adopted a self-attention mechanism to dynamically identify adjacent lanes relevant to each lane segment. This approach allows the adjacency matrix to change dynamically over time, better capturing the complex and time-varying spatial connections between lanes. The self-attention mechanism generates an adjacency matrix by calculating the relationships between each node and all other nodes. Given the traffic state vector $\dot{x}^{l_{i,j}}$ of lane segment $l_{i,j}$, we first compute the Query ($Q$), Key ($K$), and Value ($V$) using learnable weight matrices:
\begin{equation}
\begin{aligned}
&Q^{l_{i,j}} = W^Q\dot{x}^{l_{i,j}} \\
&K^{l_{i,j}} = W^K\dot{x}^{l_{i,j}}\\
&V^{l_{i,j}} = W^V\dot{x}^{l_{i,j}}
\end{aligned}
\end{equation}
where $W^Q, W^K, W^V$ are the weight matrices for Query, Key, and Value, respectively. These weight matrices are trainable parameters that allow the model to learn specific patterns and relationships between different traffic states during training. The attention coefficient $\alpha_{i_{i,j},l_{a,b}}$ between lane segment $l_{i,j}$ and lane segment $l_{a,b}$ is then computed as:
\begin{equation}
\alpha_{i_{i,j},l_{a,b}} = softmax\big(LeakyReLU(\frac{Q^{l_{i,j}}(K^{l_{a,b}})^T}{\sqrt{d_k}})\big)
\end{equation}

In this equation, $d_k$ is the dimension of the key vector, which scales the dot product to avoid overly large values. The softmax function ensures that the sum of all output attention coefficients for a given lane segment $l_{i,j}$ equals 1. This normalization ensures that the relative importance of each neighboring lane segment is appropriately weighted, maintaining the consistency and interpretability of the attention mechanism.

Using this dynamic adjacency matrix, we further perform graph convolution operations, effectively combining the spatial information of adjacent lane segments. The integration of self-attention coefficients into the adjacency matrix allows the model to dynamically adapt to traffic patterns by prioritizing critical spatial dependencies. The new feature of each node is computed by aggregating the weighted sum of its own features and the features of its neighboring nodes. The new feature $\bar{x}^{l_{i,j}}$ of lane segment $l_{i,j}$ is calculated as follows:
\begin{equation}
\bar{x}^{l_{i,j}} = \rho\big(\sum_{l_{a,b}\in\mathcal{N}(l_{i,j})}\alpha_{l_{i,j},l_{a,b}}\dot{x}^{l_{a,b}}\big)
\end{equation}
where $\mathcal{N}(l_{i,j})$ represents the set of neighboring segments of lane segment $l_{i,j}$, $\rho$ is a nonlinear activation function, and $\alpha_{l_{i,j},l_{a,b}}$ are the attention coefficients obtained from the self-attention mechanism. This calculation ensures that the new feature of each node is a weighted combination of the features of its neighbors, where the weights are determined by the self-attention coefficients.

Additionally, considering the importance of training time, especially in real-time traffic management systems that require quick responses and updates, our model design aims to optimize computational efficiency. Through efficient self-attention mechanisms and graph convolution operations, the model can learn and adapt to changes in the traffic network in a shorter time, thereby enhancing the timeliness and accuracy of overall predictions.

\subsubsection{Independent Temporal MLP Network} Although various Transformer model variants \cite{zhou2021informer,zhou2022fedformer,liu2021pyraformer} have been adapted to suit the characteristics of time series data, it has been demonstrated that a remarkably simple linear model, DLinear\cite{zeng2023transformers}, can outperform most of these Transformer-based prediction models.  The current research provides a solid foundation for designing a simple yet efficient Independent MLP (Multilayer Perceptron) network layer suitable for temporal-dependency modeling in lane-level traffic networks.

Initially, we segment the state vector $\dot{x}^{l_{i,j}}$ of each individual lane segment into equally-sized patches $p$:
\begin{equation}
p_n = \dot{x}^{l_{i,j}}_{(n-1)k:nk}, \text{for } n = 1, ..., m
\end{equation}
where $m$ represents the total number of patches and $k = \frac{T}{m}$ is the length of each patch. We then use multi-layer MLPs with residual connections to model the dependencies both within and between these patches:
\begin{equation}
p_n' = MLP(p_n)
\end{equation}
\begin{equation}
x'^{l_{i,j}} = MLP(p_1', ..., p_m')
\end{equation}

The MLP components can be stacked to accommodate datasets of varying scales, with residual connections between these components $\hat{x}^{l_{i,j}} = x'^{l_{i,j}} + x''^{l_{i,j}} + x'''^{l_{i,j}}$.

\subsubsection{Gating Network} Compared to sequential processing of spatio-temporal relationships, parallel processing of these relationships among lanes offers higher efficiency. The gating network combines spatial and temporal dependencies through adaptive weight modulation to produce preliminary prediction results:
\begin{equation}
\bar{y}^{l_{i,j}} = \lambda\bar{x}^{l_{i,j}} + (1-\lambda)\hat{x}^{l_{i,j}}
\end{equation}
where gating weight $ \lambda$ can be used to adaptively adjust the contribution of dynamic graph network and independent temporal MLP network.

\begin{algorithm}[H]
\caption{Training Procedure for GraphMLP}
\label{alg:graphmlp}
\KwIn{Graph \(\mathcal{G} = (\mathcal{V}, \mathcal{E})\), feature matrix \(\mathbf{X}\), ground truth labels \(\mathbf{Y}\), number of epochs \(T_{train}\), learning rate \(\eta\), loss function \(\mathcal{L}\).}
\KwOut{Trained parameters \(\theta\).}

\textbf{Step 1: Model Initialization.} Initialize model parameters \(\theta\), including the dynamic graph module, independent temporal MLP, gating network, and normalization parameters.

\For{epoch \(t = 1\) to \(T_{train}\)}{

    \textbf{Step 2: Instance Normalization.}  
    Normalize the input feature \(\mathbf{X}\) for each lane segment \(l_{i,j}\) using:
    \[
    \dot{x}_{l_{i,j}} = \psi \cdot \frac{x_{l_{i,j}} - \mu(x_{l_{i,j}})}{\sqrt{\sigma^2(x_{l_{i,j}})}} + \beta
    \]
    where \(\mu(x_{l_{i,j}})\) and \(\sigma^2(x_{l_{i,j}})\) are the mean and variance over the time window for \(l_{i,j}\), and \(\psi\), \(\beta\) are learnable parameters.

    \textbf{Step 3: Spatial Dependency Modeling.}  
    Construct the dynamic adjacency matrix and compute attention coefficients:
    \[
    \alpha_{l_{i,j}, l_{a,b}} = \text{Softmax}\Big(\text{LeakyReLU}\Big(\frac{Q_{l_{i,j}} \cdot K_{l_{a,b}}^\top}{\sqrt{d_k}}\Big)\Big)
    \]
    Perform graph convolution to update spatial features:
    \[
    \bar{\mathbf{X}} = \rho\Big(\sum_{l_{a,b} \in \mathcal{N}(l_{i,j})} \alpha_{l_{i,j}, l_{a,b}} \cdot \dot{\mathbf{X}}\Big)
    \]

    \textbf{Step 4: Temporal Dependency Modeling.}  
    Divide the normalized temporal input into patches:
    \[
    p_n = \dot{\mathbf{X}}[(n-1)k:nk], \quad n = 1, \dots, m
    \]
    Process patches through the independent temporal MLP:
    \[
    p'_n = \text{MLP}(p_n), \quad \hat{\mathbf{X}} = \text{MLP}(p'_1, \dots, p'_m)
    \]

    \textbf{Step 5: Gating Network for Feature Fusion.}  
    Fuse spatial and temporal features:
    \[
    \bar{\mathbf{Y}} = \lambda \cdot \bar{\mathbf{X}} + (1 - \lambda) \cdot \hat{\mathbf{X}}
    \]
    where \(\lambda\) is the gating weight.

    \textbf{Step 6: Instance Denormalization.}  
    Restore the statistical properties removed during normalization for each lane segment \(l_{i,j}\) in \(\hat{Y}\) using:
    \[
    \hat{y}^{l_{i,j}} = \sqrt{\sigma(x^{l_{i,j}})}\left(\frac{\bar{y}^{l_{i,j}}-\beta}{\psi}\right)+\mu(x^{l_{i,j}})
    \]

    \textbf{Step 7: Loss Computation and Backpropagation.}  
    Compute loss:
    \[
    \mathcal{L} = \text{Loss}(\hat{\mathbf{Y}}, \mathbf{Y})
    \]
    Update parameters:
    \[
    \theta \gets \theta - \eta \cdot \nabla_{\theta} \mathcal{L}
    \]
}

\Return \(\theta\)\;
\end{algorithm}

\subsubsection{Instance Denormalizatiom} In the final stage, we leverage the statistical insights gained from instance normalization to reintegrate non-stationary elements back into the dataset. This process culminates in the derivation of the ultimate prediction outcome $y^{l_i,j}$, which can be mathematically represented as:

\begin{equation}
\hat{y}^{l_{i,j}} = \sqrt{\sigma(x^{l_{i,j}})}\left(\frac{\bar{y}^{l_{i,j}}-\beta}{\psi}\right)+\mu(x^{l_{i,j}})
\end{equation}

In this equation, the restored values account for the inherent variability within the data, enhancing the precision of our predictions.

Subsequently, the GraphMLP model undergoes training through the optimization of the Mean Squared Error (MSE). The corresponding loss function is structured as follows:
\begin{equation}
\mathcal{L}(\theta) = \frac{1}{N}\sum^j_{j=1}\sum^{i}_{i=1}(y^{l_{i,j}}-\hat{y}^{l_{i,j}})^2
\end{equation}
where $\theta$ represents the ensemble of trainable parameters within the GraphMLP framework. \textbf{Algorithm 1} presents the pseudocode for the training process of GraphMLP. GraphMLP offers a more lightweight and adaptable foundational solution for lane-level traffic prediction, capable of accommodating the spatial complexity and dynamics of lanes.

\section{Benchmark \label{section 5}}
In this benchmark, we first introduce three benchmark datasets that we have collected, processed, and made publicly available. These datasets represent regular and irregular freeways, namely the PeMS and PeMSF datasets, as well as the HuaNan dataset representing urban expressways. By treating lanes as a unified graph structure and utilizing the adjacency relationships constructed as described in \textbf{Section~\ref{section 4}}, we introduce spatio-temporal graph models to advance lane-level traffic prediction. Additionally, we conduct a unified experimental comparison of these models with lane-level prediction algorithms, providing more challenging benchmarks for the field of lane-level traffic prediction.

\subsection{Datasets}
We have collected and processed real-world lane-level traffic data from two representative sources: the public Performance Measurement System (PeMS\footnote{https://pems.dot.ca.gov/}) for the Santa Ana Freeway in Los Angeles, and sensor data from the HuaNan Expressway in Guangzhou, China. Based on these two sources, we constructed three distinct datasets to support our benchmark experiments. The PeMS dataset features a regular 5-lane configuration. The HuaNan dataset captures a typical 4-lane expressway in an urban environment. In contrast, the PeMSF dataset is an extended variant of PeMS that includes freeway segments with entrance ramps, where some road segments contain a sixth lane, representing irregular lane configurations.

\begin{figure}
\vspace{0.40cm}
\captionsetup[subfloat]{labelfont=scriptsize,textfont=scriptsize}
\centering
\subfloat[Distribution in the PeMS and PeMSF datasets.]{\includegraphics[width=3in]{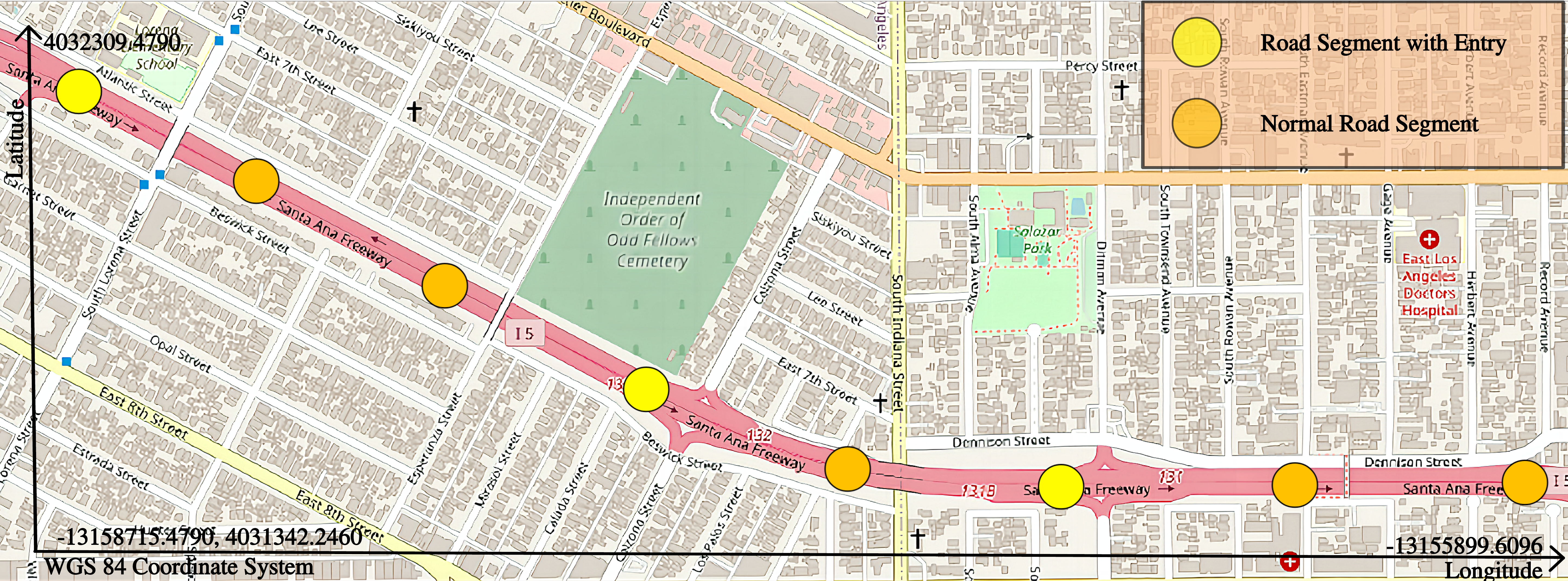}}
\newline
\centering
\subfloat[Distribution in the HuaNan dataset.]{
\hspace{-0.6cm}
\includegraphics[width=3in]{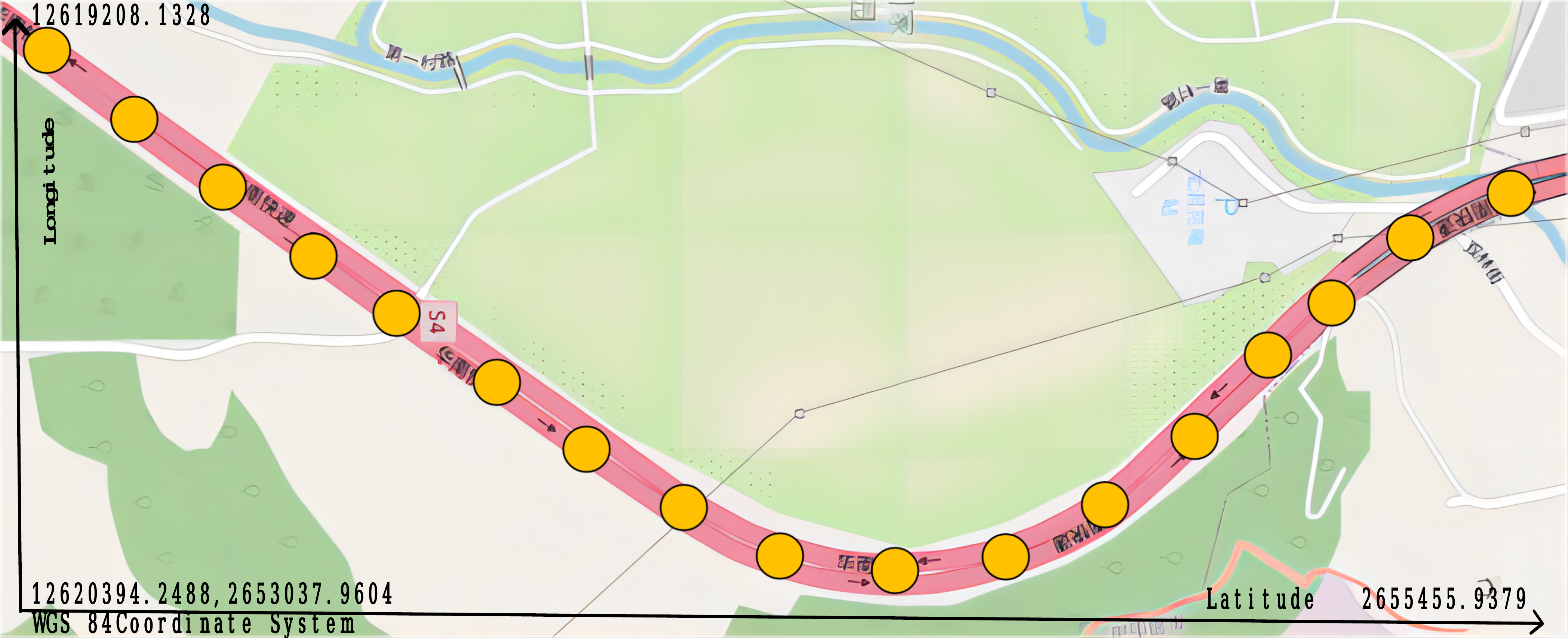}}
\caption{\label{fig:sensors} Distribution of sensors in the two data sources.} 
\end{figure}

This triplet of datasets is deliberately designed to encompass both regular and irregular lane conditions, enabling comprehensive evaluation of lane-level traffic prediction models. \textbf{Figure \ref{fig:sensors}} displays the sensor locations across both sources, along with examples of segments containing entrance lanes.

Evaluating the model on these datasets allows us to assess its effectiveness, accuracy, and generalizability across varying lane structures and real-world complexities. This ensures that the benchmark is not confined to simplified or idealized road scenarios, but rather reflects a wide spectrum of practical traffic conditions.

\begin{figure}
\captionsetup[subfloat]{labelfont=scriptsize,textfont=scriptsize}
\centering
\subfloat[PeMS Dataset]{
\hspace{-0.55cm}
\includegraphics[width=3.1in]{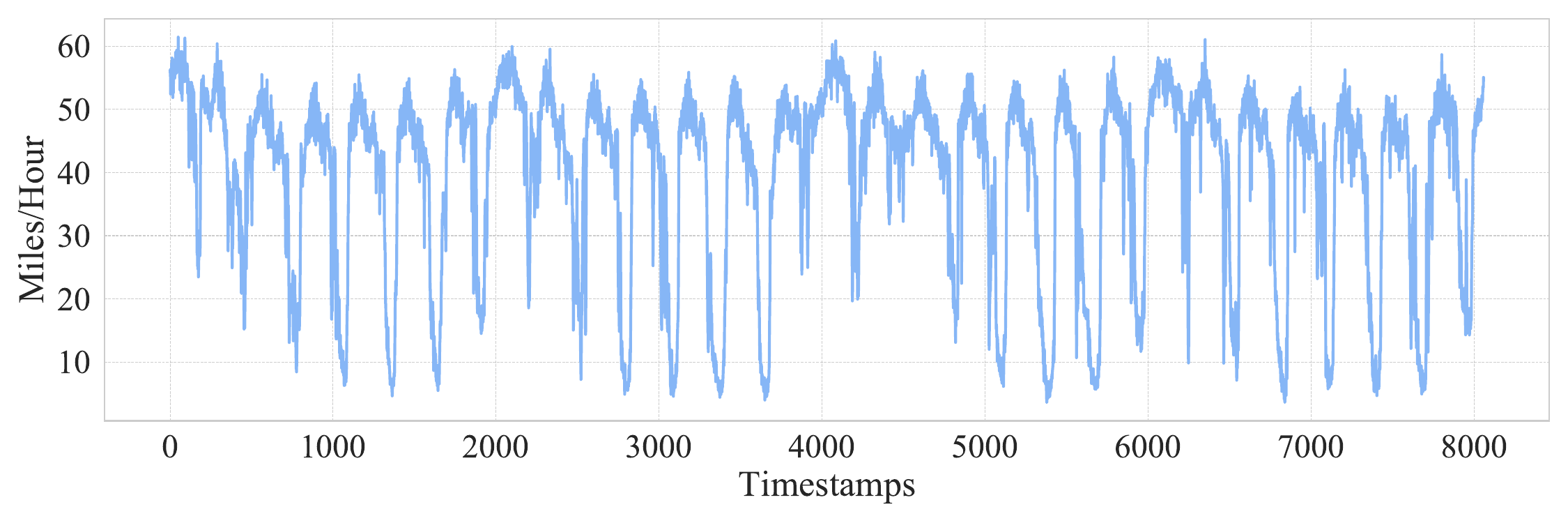}}
\newline
\subfloat[PeMSF Dataset]{
\hspace{-0.55cm}
\includegraphics[width=3.1in]{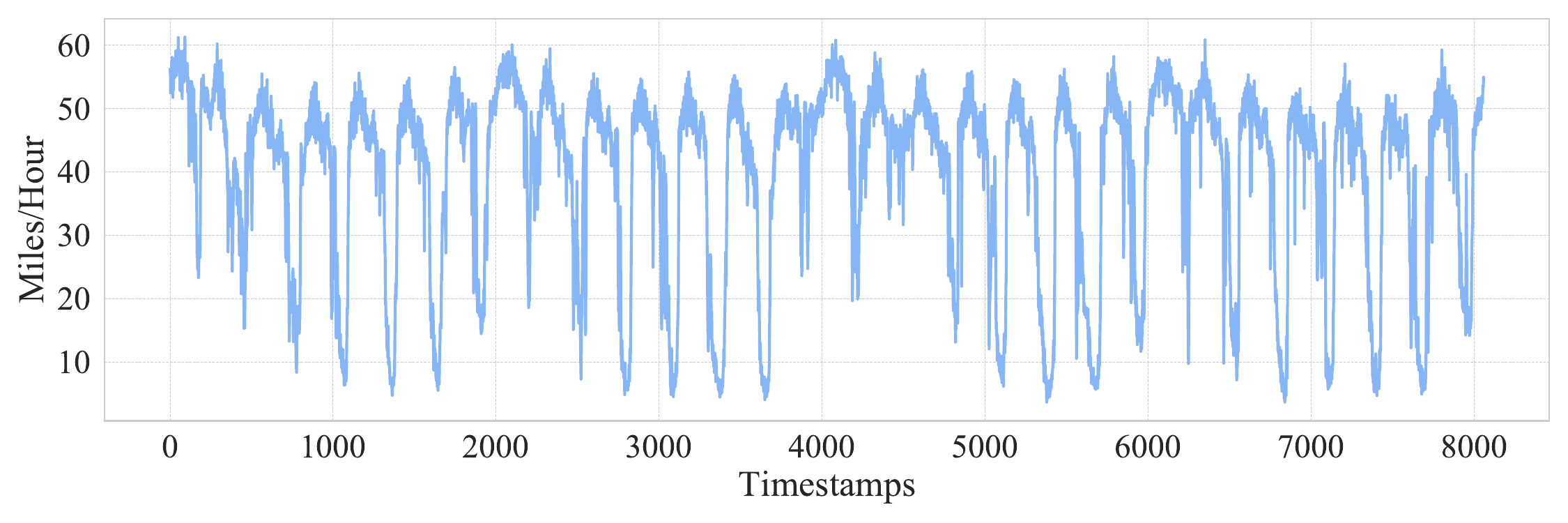}}
\newline
\subfloat[HuaNan Dataset]{
\hspace{-1cm}
\includegraphics[width=3.1in]{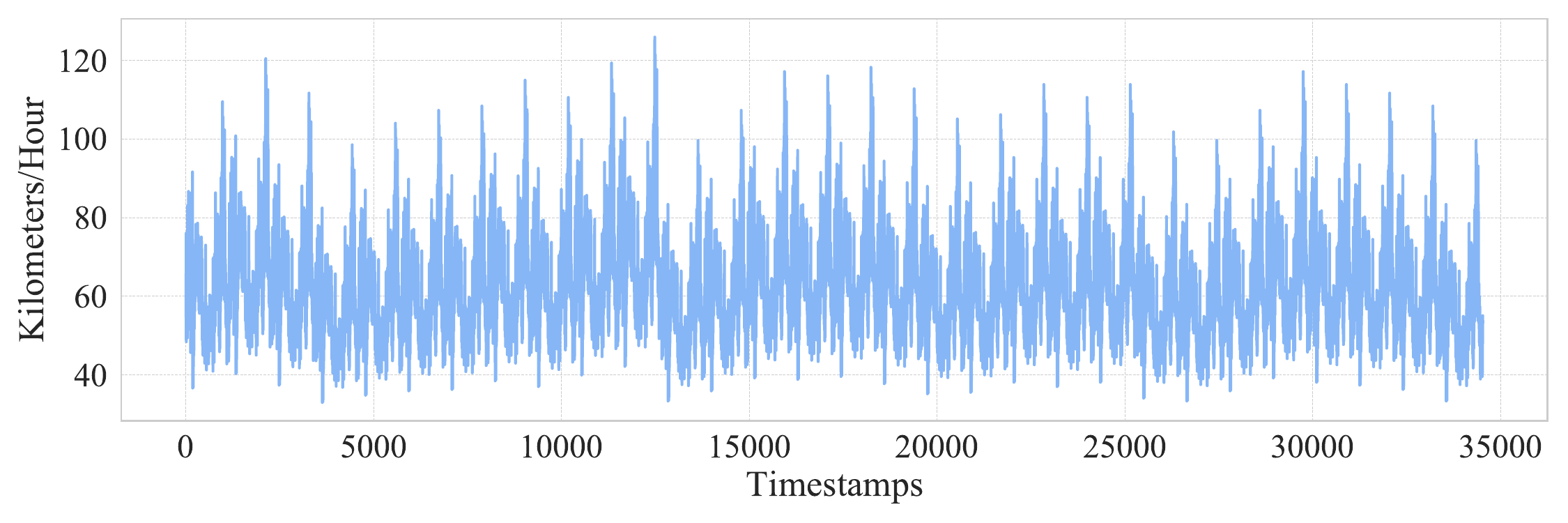}}
\caption{Average trends of all sensors across the three datasets.} 
\label{fig:DataMean} 
\end{figure} 

\subsubsection{PeMS Dataset} 
The PeMS dataset is a renowned open traffic data system managed by the California Department of Transportation, widely used in road-level traffic studies. To construct a high-quality lane-level dataset, we carefully selected 8 sensors from approximately 40,000 available detectors, based on data continuity and operational reliability. The resulting dataset covers the period from February 5 to March 5, 2017, including lane-level speed and flow data. To address rare cases of missing values, we applied mean imputation using adjacent time slots. The data shows clear daily periodicity, as visualized in \textbf{Figure \ref{fig:DataMean}(a)}, making it suitable for temporal traffic modeling. Researchers can also access the original PeMS platform for further exploration and extension.

\subsubsection{PeMSF Dataset}
The PeMSF dataset is an extended variant of the PeMS dataset that includes freeway segments with entrance ramps, where the number of lanes increases from 5 to 6 at certain locations. This variation reflects more realistic and irregular lane-level structures common in urban expressways, especially near merges or exits. PeMSF is designed to test the adaptability of prediction models to dynamic and non-uniform lane configurations, an essential capability for deployment in complex real-world traffic networks. \textbf{Figure \ref{fig:sensors}(a)} highlights these irregular segments.

\begin{figure}[t]
\captionsetup[subfloat]{labelfont=scriptsize,textfont=scriptsize}
  \subfloat[Sensor 1]{\includegraphics[width=0.24\textwidth]{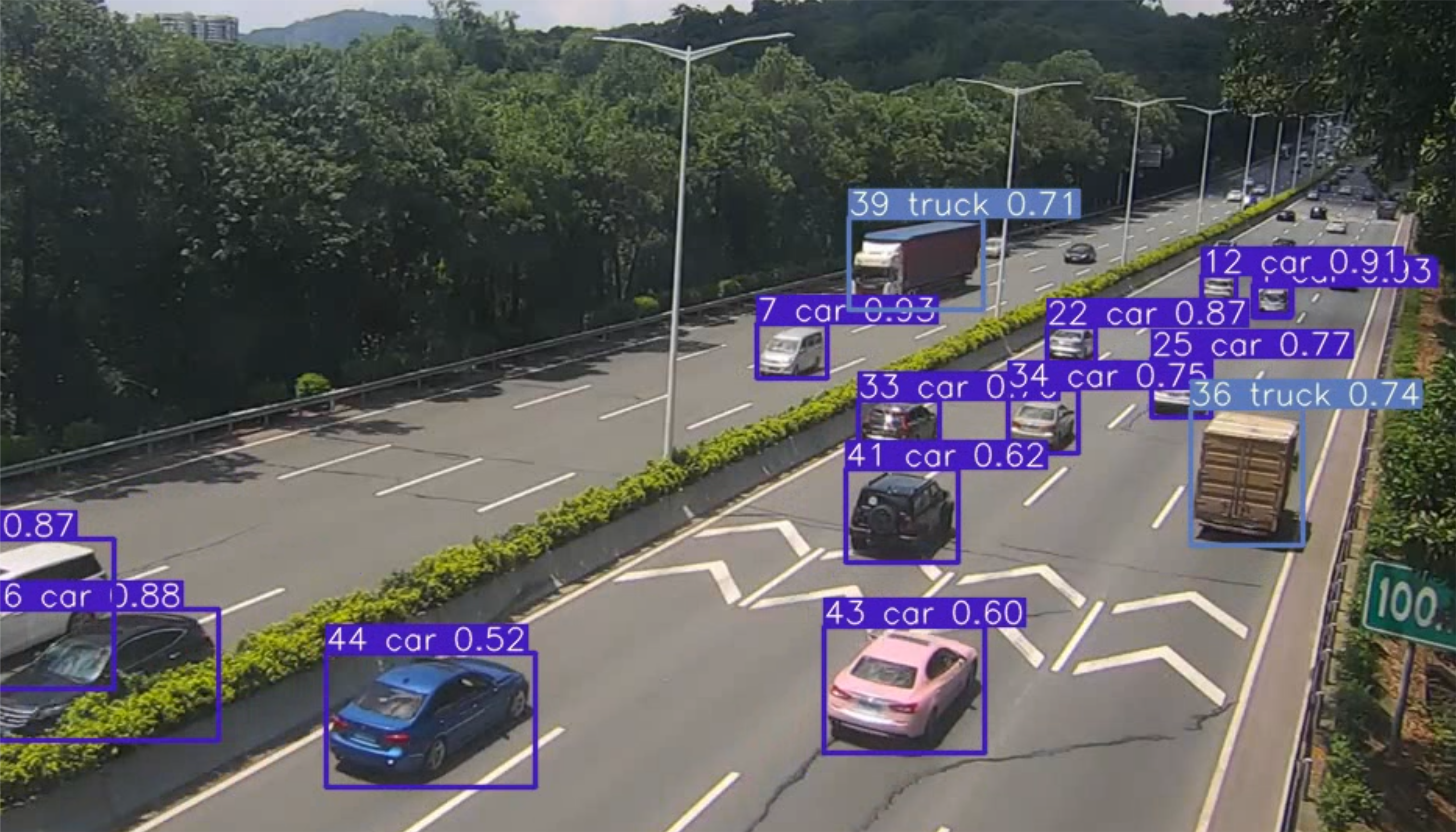}}
 \hfill 	
  \subfloat[Sensor 3]{\includegraphics[width=0.24\textwidth]{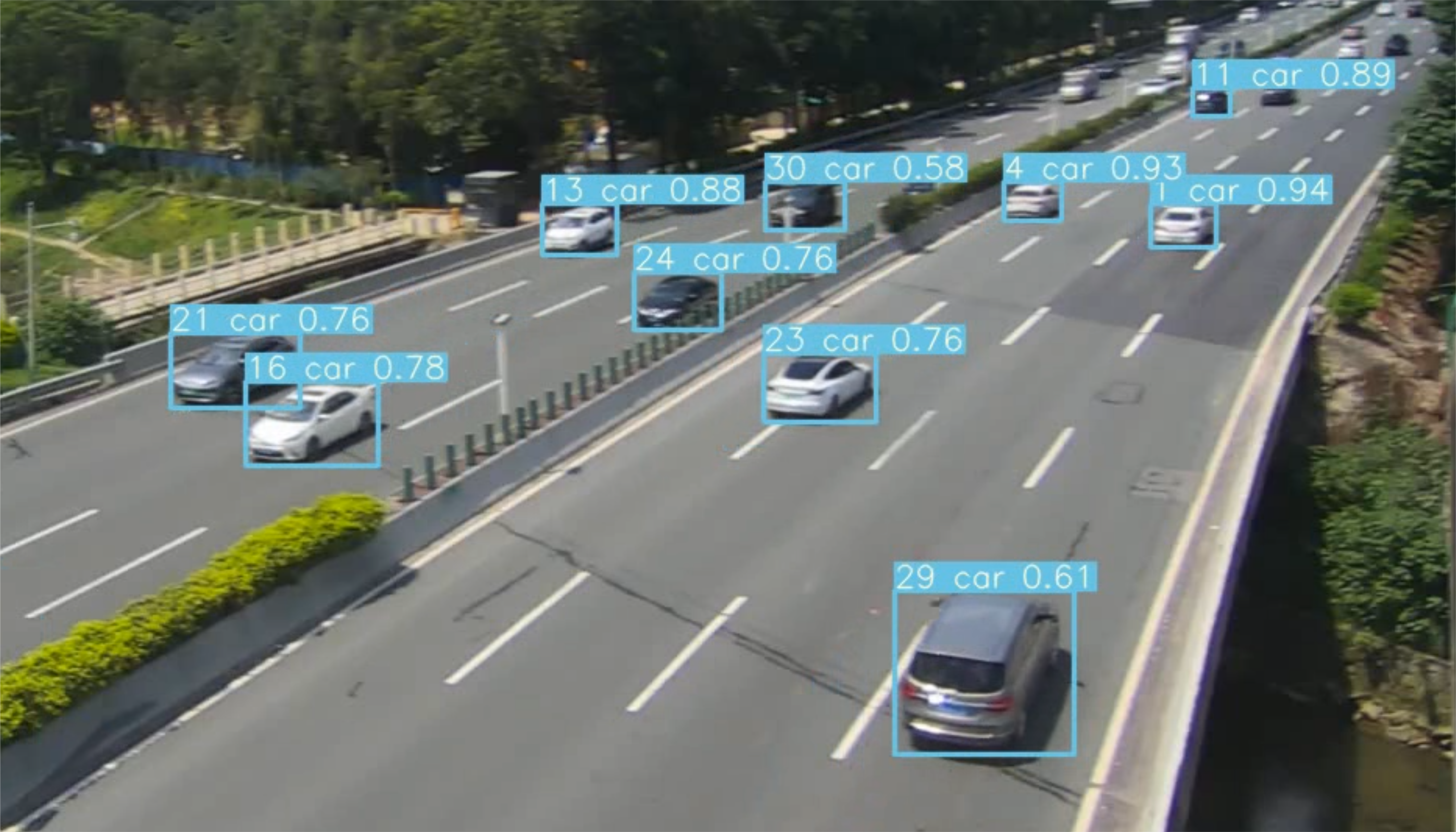}}
 \hfill	
  \newline
  \subfloat[Sensor 5]{\includegraphics[width=0.24\textwidth]{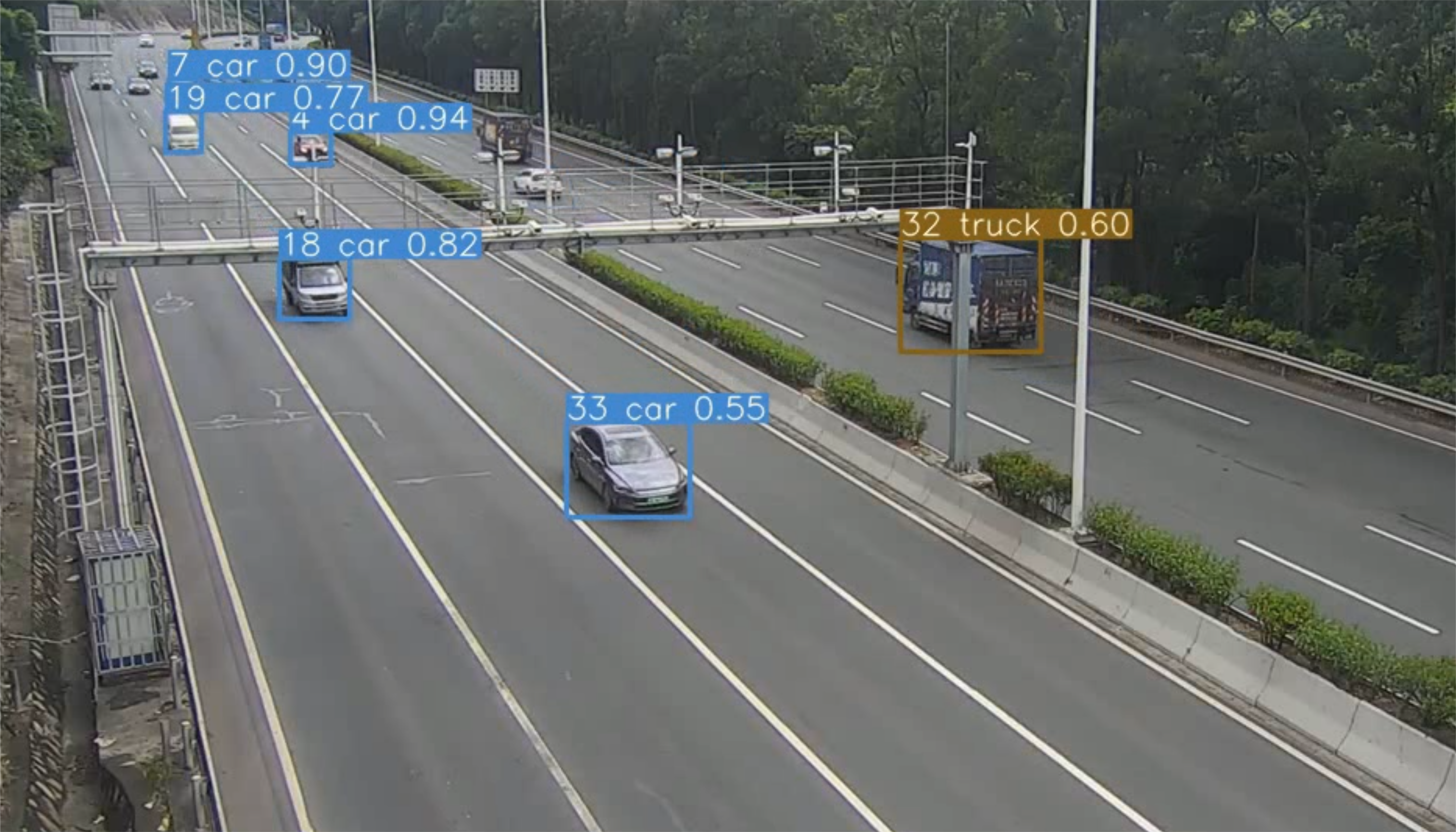}}
 \hfill 	
  \subfloat[Sensor 8]{\includegraphics[width=0.24\textwidth]{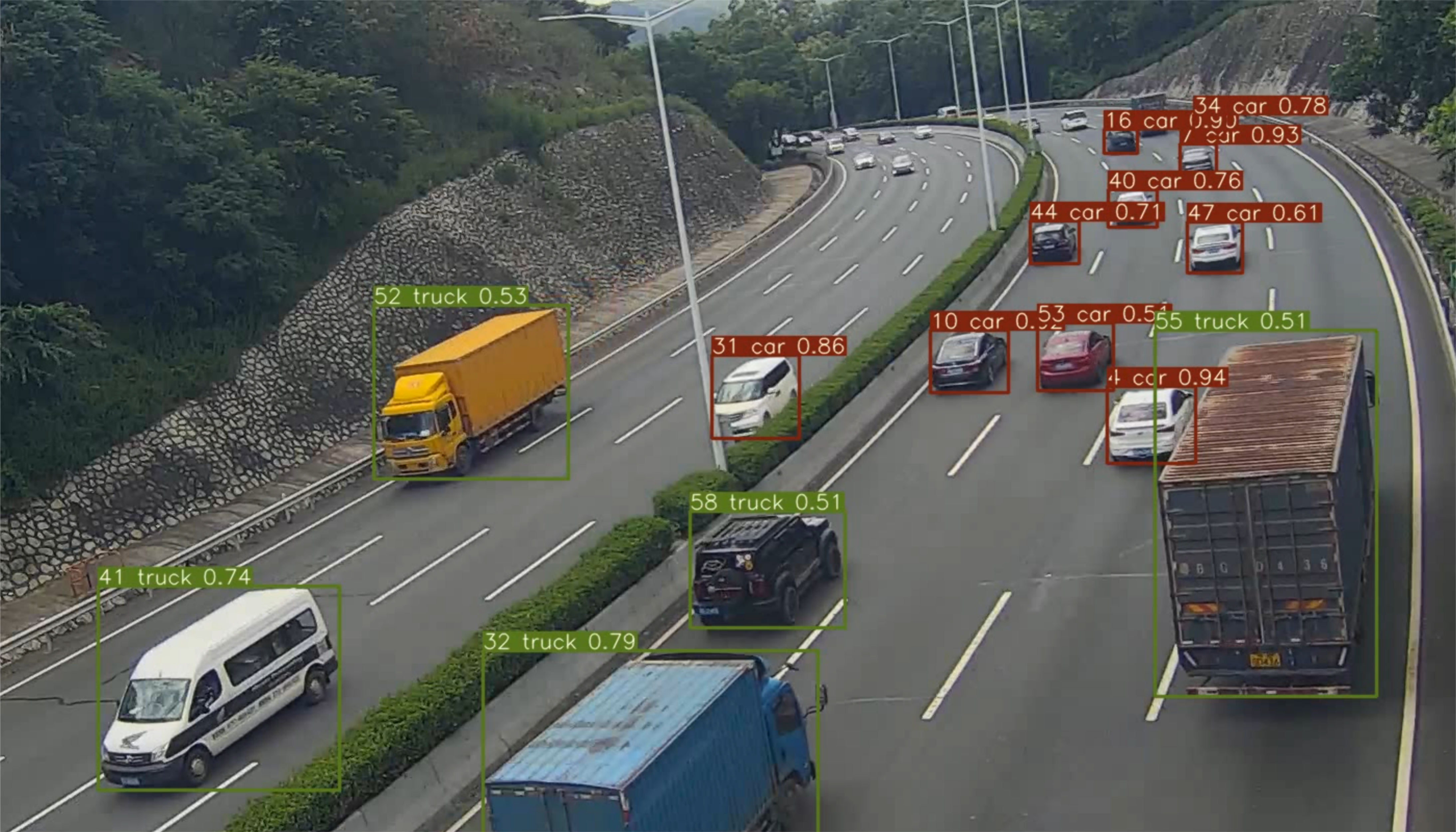}}
 \hfill	
\caption{\label{fig:sensors_huanan}The process of extracting the average speed of vehicles from some sensors using the Strongsort algorithm.}
\end{figure}

\subsubsection{HuaNan Dataset} The HuaNan dataset originates from the Huanan Expressway in Guangzhou, China, a major urban expressway in the city, which is particularly prone to congestion during rush hours. This makes it an ideal location for studying urban traffic flow and congestion patterns. To obtain comprehensive traffic flow data, we employed the advanced Strongsort tracking algorithm \cite{du2023strongsort}, extracting traffic speed and volume information from 18 video sensors installed on the expressway. The data covers the period from July 22 to August 22, 2022. The Strongsort algorithm is known for its efficient and accurate vehicle tracking capabilities, especially in processing video data of high-density traffic flows. As shown in \textbf{Figure \ref{fig:sensors_huanan}}, we detail the process of data extraction from the video sensors, ensuring the accurate extraction of key traffic parameters from the original video footage.

To ensure the quality and reliability of the dataset, we thoroughly cleaned and preprocessed the extracted data. This included identifying and correcting anomalies, filling in missing values, and performing necessary data normalization procedures. This process ensured the completeness and consistency of the dataset, laying a solid foundation for subsequent analysis and modeling. In \textbf{Figure \ref{fig:DataMean}}, we present a trend analysis of all sensors data. These trends reflect the traffic flow characteristics of the HuaNan Expressway during different periods, providing an intuitive perspective on the usage patterns of the road and the characteristics of peak traffic periods. Additionally, the chosen road segment, devoid of entrances and exits, makes this dataset particularly suitable as a benchmark for prediction models that are only applicable to roads with a regular number of lanes.
\begin{figure*}[ht]
\captionsetup[subfloat]{labelfont=scriptsize,textfont=scriptsize}
  \subfloat[PeMS Dataset]{\includegraphics[width=0.33\textwidth]{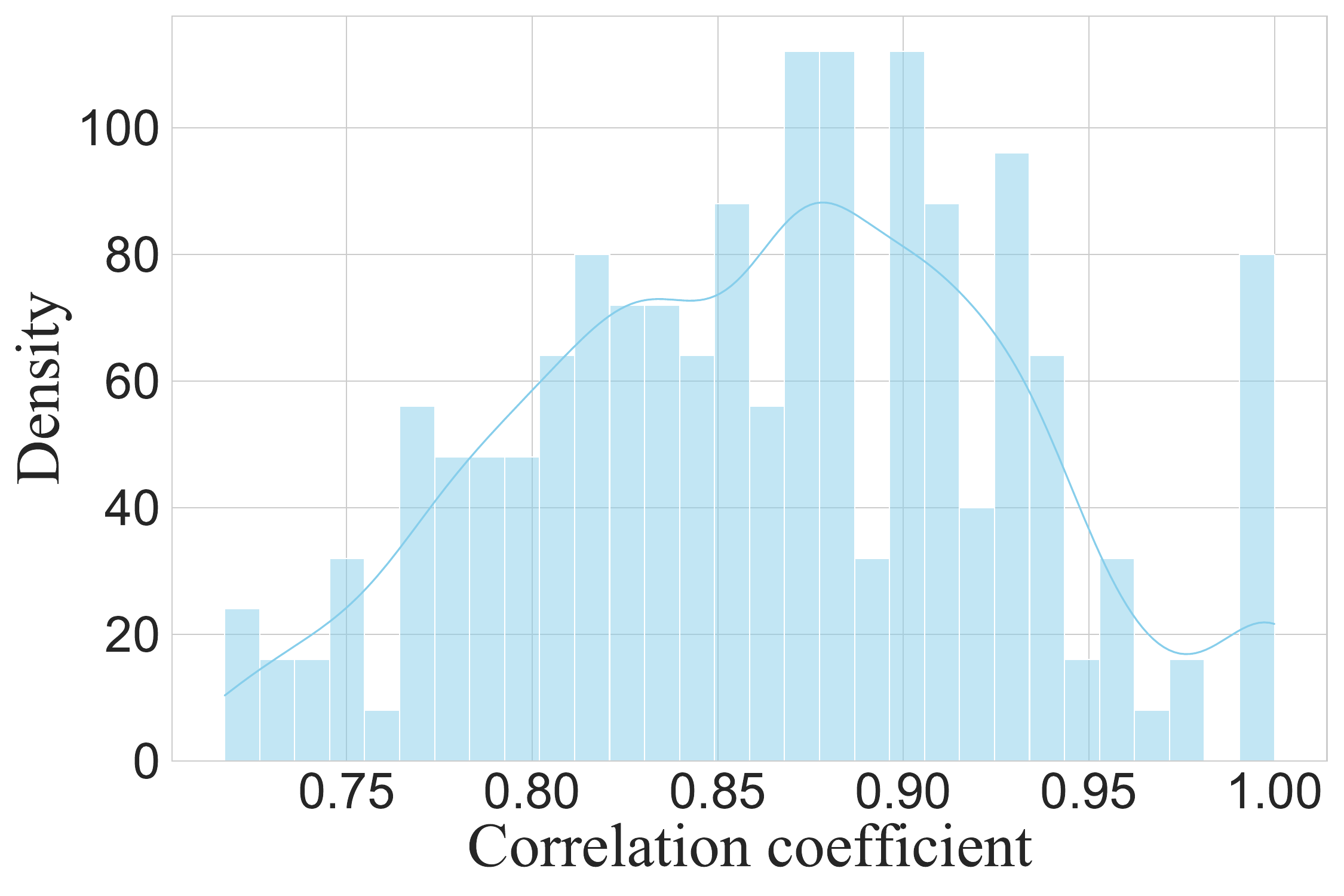}}
 \hfill 	
  \subfloat[PeMSF Dataset]{\includegraphics[width=0.33\textwidth]{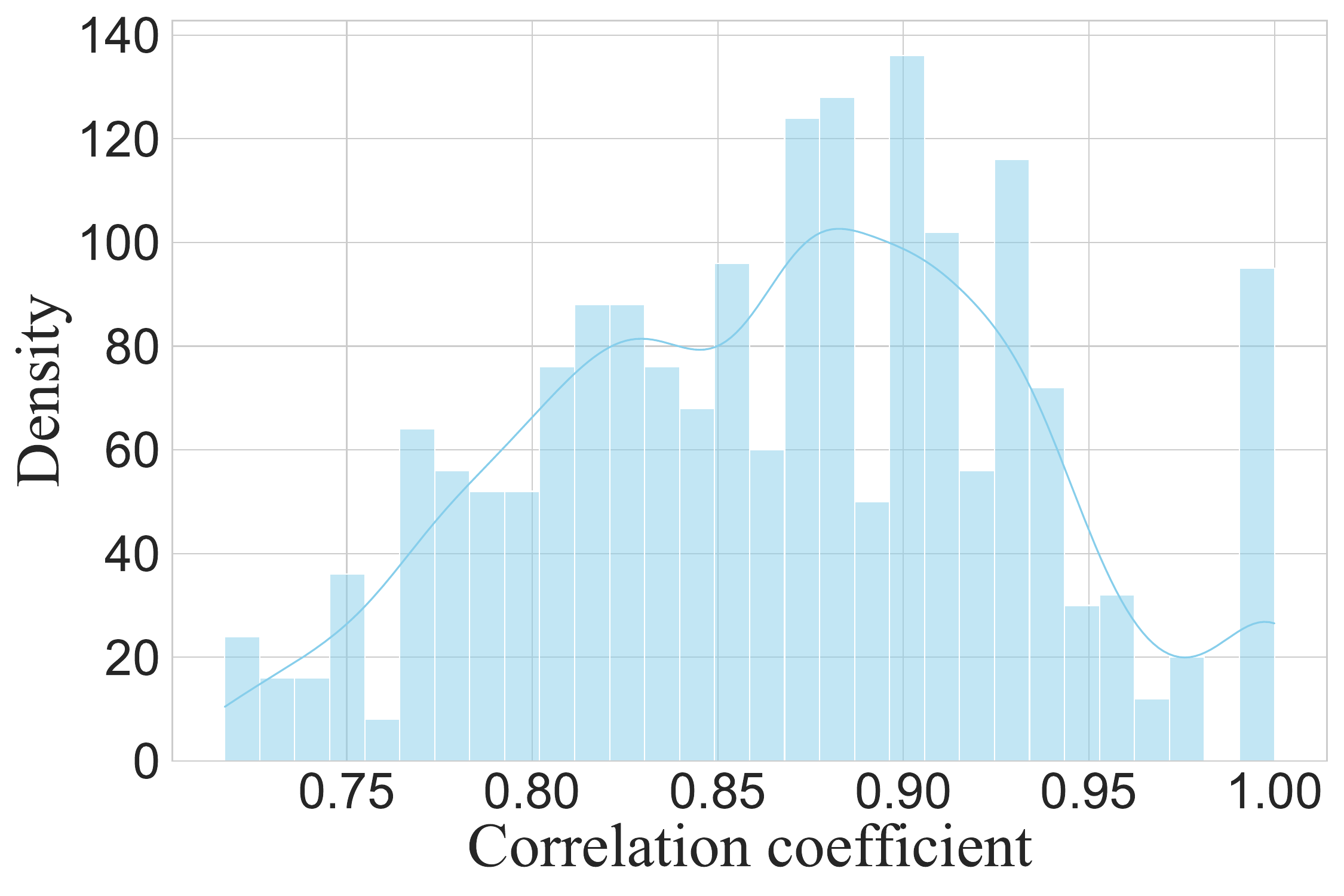}}
 \hfill	
  \subfloat[HuaNan Dataset]{\includegraphics[width=0.33\textwidth]{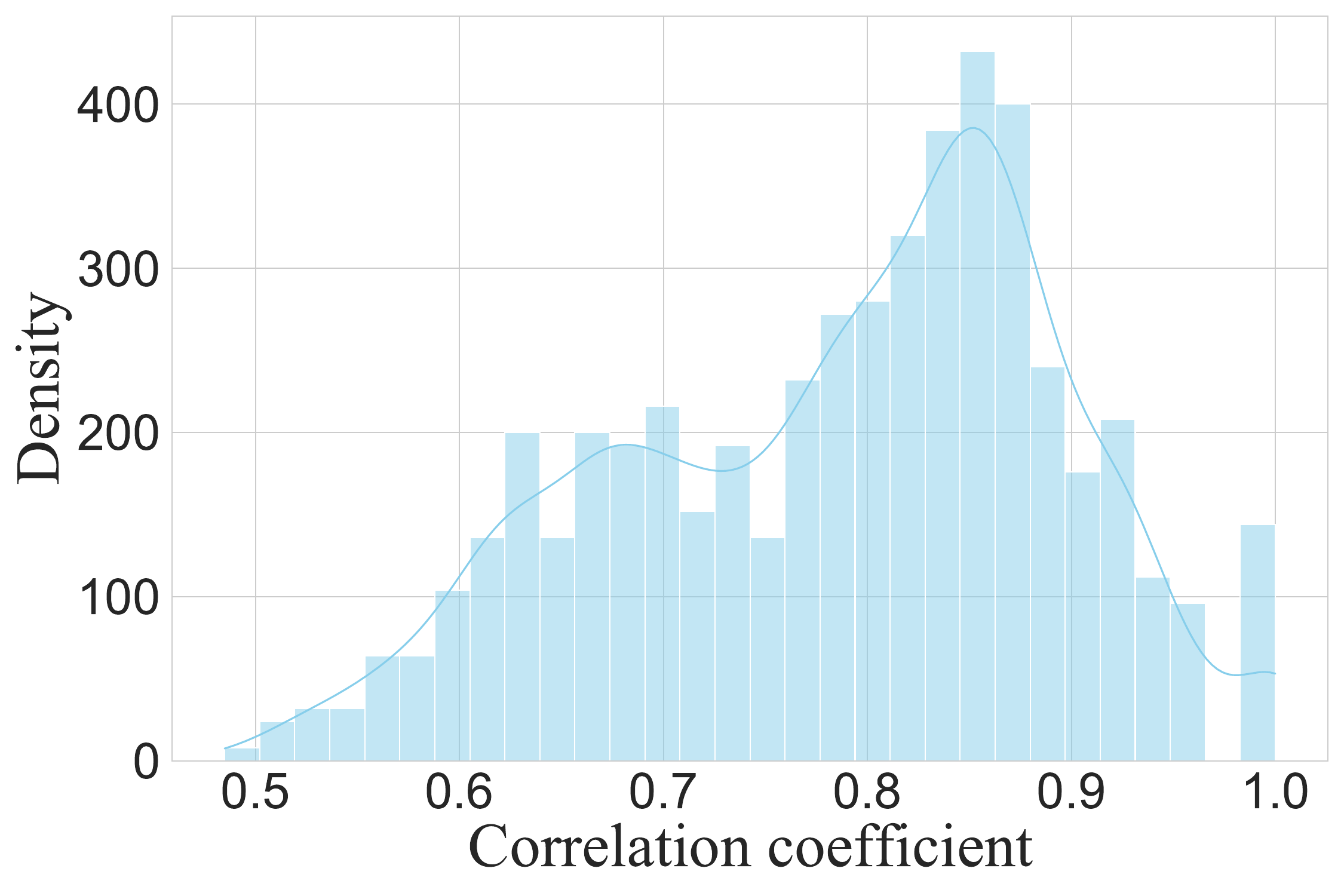}}
  \caption{\label{fig:Data_poisson}The Poisson correlation distributions and kernel density curves for the three datasets. }
\end{figure*}

\subsubsection{ Statistical Analysis} To assess the utility of the data, we analyzed the distribution of Poisson correlation coefficients between lane segment vectors and performed kernel density estimation (KDE) curve analysis. As depicted in \textbf{Figure \ref{fig:Data_poisson}}, \textbf{subfigures (a)} and \textbf{(b)} reveal that the lane segment pairs in both the PeMS and PeMSF datasets exhibit high correlation coefficients. Due to the predominance of identical points, the KDE curves for both datasets follow a similar trend. The PeMSF dataset, with its larger number of lane segments, demonstrates a higher density in the correlation distribution.

In \textbf{subfigure (c)}, the distribution of Poisson correlation coefficients and the KDE curve for the HuaNan dataset are presented. Compared to the datasets derived from the PeMS data source, the HuaNan dataset exhibits a more dispersed distribution of correlation coefficients, indicating a greater diversity in pattern distribution. However, the KDE curve suggests a higher concentration of density in the regions of high correlation. This analysis implies significant spatial correlations in all three datasets.

To visually present and compare the core characteristics of our publicly available datasets, we summarize the detailed information of the three datasets in \textbf{Table~\ref{tab:datasets}}, providing a systematic comparative analysis.

\begin{table}[htbp]
  \centering
  \caption{Statistics of Datasets}
    \begin{tabular}{cccc}
    \toprule
    Parameters  & PeMS & PeMSF & HuaNan \\
    \midrule
    Timespan & 2/5/2017- & 2/5/2017- & 7/22/2022- \\
        & 3/5/2017 & 3/5/2017 & 8/22/2022 \\
    Region & Los Angeles & Los Angeles & Guangzhou \\
    \# of Objects & 322,360 & 346,537 & 3,214,080 \\
    \# of Static Edges & 67  & 70  & 121 \\
    Road Type & Freeway & Freeway & Urban Expressway \\
    Unit & Miles/Hour & Miles/Hour & Kilometers /Hour \\
    Time Interval & 5 min & 5 min & 2min \\
    \bottomrule
    \end{tabular}%
  \label{tab:datasets}%
\end{table}%

\subsection{Metrics}
In the field of traffic prediction, particularly at the lane level, a model's accuracy is typically assessed using three key metrics: Root Mean Square Error (RMSE), Mean Absolute Error (MAE), and Mean Absolute Percentage Error (MAPE). However, beyond accuracy, the practicality of a model also hinges on the duration of its training. This is especially crucial in lane-level traffic prediction scenarios where rapid response and real-time updates are essential. Therefore, focusing solely on predictive accuracy while overlooking training time could limit the model's performance in practical applications.

Acknowledging this, we have included training expenditure as an evaluation metric, aiming to ensure that models consider both training time and predictive accuracy. The training Cost metric is calculated as the time required for each iteration of model training multiplied by 100, with the unit being seconds. The formulas for all the evaluation metrics used in our study are as follows:

\begin{equation}
MAE = \frac{1}{z}\sum^{z}_{t = 1}\frac{1}{N}\sum^{J}_{j = 1}\sum^{I}_{i = 1}\Big|y^{l_{i,j}}_t - \hat{y}^{l_{i,j}}_t \Big|
\end{equation}
\begin{equation}
RMSE = \sqrt{\frac{1}{z}\sum^{z}_{t = 1}\frac{1}{N}\sum^{J}_{j = 1}\sum^{I}_{i = 1}\Big(y^{l_{i,j}}_t - \hat{y}^{l_{i,j}}_t \Big)^2}
\end{equation}
\begin{equation}
MAPE = \frac{1}{z}\sum^{z}_{t = 1}\frac{1}{N}\sum^{J}_{j = 1}\sum^{I}_{i = 1}\Big|\frac{y^{l_{i,j}}_t - \hat{y}^{l_{i,j}}_t}{y^{l_{i,j}}_t} \Big|
\end{equation}
\begin{equation}
Cost = (\text{times per iteration})*10^{-2}
\end{equation}
where $\hat{y}^{l_{i,j}}_t$ is the predicted value at the lane segment $l_{i,j}$ at time $t$, $z$ is the predicted horizon, and $y^{l_{i,j}}_t$ is the corresponding ground truth.
\subsection{Baselines and Code Configuration} 
We conducted a comparative analysis of fourteen models, including the GraphMLP model, designed for lane-level traffic prediction, and nine graph-structured spatio-temporal prediction models across three datasets. Given the general lack of publicly available code for models proposed in the field of lane-level traffic prediction, this section not only introduces the baseline model but also briefly outlines the process of reproducing the experimental code and its source. The input and output of the reproduction framework remain consistent with those in \textbf{Algorithm 1}.

\subsubsection{Lane-level Traffic Prediction Model}
The majority of the baseline models were replicated by us, while a few were implemented using source code and configurations provided by the original authors. Should any authors feel that our replication is inappropriate, we welcome them to contact us for necessary modifications. Below, we provide a succinct overview of the lane-level model baselines and their code configurations.

\begin{itemize}
\item Cat-RF-LSTM \cite{zhao2022hybrid}: Combines Catboost for spatio-temporal feature construction, Random Forest for variance reduction, and LSTM for temporal trend extraction, using stacking ensemble for final prediction. Implemented using Catboost, scikit-learn, and PyTorch libraries; CatBoost with 1,000 iterations, depth 6, learning rate 0.1; Random Forest with 100 trees; LSTM with 64 hidden units, 2 layers.

\item CEEMDAN-XGBoost \cite{lu2020hybrid}: Combines CEEMDAN for data decomposition with XGBoost for prediction. Implemented using the XGBoost library and CEEMDAN class from PyEMD, with squared error loss and 100 estimators.

\item LSTM: A specialized form of RNN featuring input, output, and forget gates. Implemented using PyTorch, with a hidden layer dimension of 64 and two layers.

\item GRU: A simplified version of LSTM, excluding forget gates and incorporating update and reset gates. Implemented using PyTorch, with a hidden layer dimension of 64 and two layers.

\item FDL \cite{gu2019short}: Combines entropy-based gray correlation analysis with LSTM and GRU for lane-level prediction. Implemented using PyTorch; both LSTM and GRU have a hidden dimension of 64 and two layers.

\item TM-CNN \cite{ke2019two}: Transforms traffic speed and volume data into matrices for prediction. Implemented using PyTorch as a single-stream(only speed or volume), multi-channel convolutional network for fairness.

\item MDL\cite{lu2020lane}: Unites ConvLSTM, convolutional, and dense layers for lane-based dynamic traffic prediction. Implemented using the authors' code\footnote{https://github.com/lwqs93/MDL} with adjustments for fairness like TM-CNN.

\item CNN-LSTM\cite{ma2020multi}: Improves short-term traffic prediction using CNN for lane analysis. Implemented with PyTorch; hidden layer dimension of 16 and one LSTM layer.

\item HGCN\cite{zhou2022lane} and DGCN\cite{wang2021lane}: The two models in question, originating from the same author, utilize identical formulas and methodologies, integrate spatial dependency analysis, data fusion, and temporal attention. Implemented using PyTorch, excluding heterogeneous data for fairness.

\item GCN-GRU\cite{li2023dynamic}: Employs GCN with data-driven adjacency matrix and GRU. Implemented in PyTorch; GCN output dimension 16, GRU with 64 hidden units, 2 layers.

\item ST-AFN\cite{shen2021st}: Features a speed process network, spatial encoder, and temporal decoder with an embedded attention mechanism.  Implemented using the authors' code\footnote{https://github.com/MCyutou/ST-AFN}.

\item STA-ED\cite{zheng2022lane}: Uses LSTM in an encoder-decoder architecture with two-stage attention. Designed in PyTorch, with a 64-unit hidden layer.

\item STMGG\cite{zeng2022modeling}: Utilizes visibility graphs, spatial topological graphs, attention-based gated mechanism, and Seq2Seq for lane-level traffic prediction. Implemented using PyTorch with a 64-unit hidden layer.

\end{itemize}

\subsubsection{Graph-structured Spatio-temporal Prediction Models}
We selected graph-structured spatio-temporal prediction models with publicly available code, endeavoring to adhere to the original parameter configurations in our experiments as closely as possible.
\begin{itemize}
\item DCRNN\cite{li2017diffusion}: Simulates diffusion on traffic graphs, combining spatial-temporal dynamics, bidirectional walks, and an encoder-decoder structure. Implemented using the authors' code\footnote{https://github.com/liyaguang/DCRNN}.

\item STGCN\cite{yu2017spatio}: Efficiently models traffic networks on graphs with a fully convolutional structure for faster training and fewer parameters. Implemented using the authors' code\footnote{https://github.com/VeritasYin/STGCN\_IJCAI-18}.

\item MTGNN \cite{wu2020connecting}: Presents a graph neural network framework for multivariate time series, automatically extracting variable relations and capturing spatial and temporal dependencies through innovative layers. Implemented using the authors' code\footnote{https://github.com/nnzhan/MTGNN}.

\item ASTGCN \cite{guo2019attention}: Attention-based Spatio-Temporal Graph Convolutional Network (ASTGCN) models recent, daily, and weekly traffic dependencies using a space-time attention mechanism and graph convolution. Implemented using the authors’ code\footnote{ https://github.com/wanhuaiyu/ASTGCN}.

\item GraphWaveNet \cite{wu2019graph}: Employs an adaptive dependency matrix and node embedding to capture spatial data dependencies, processing long sequences with dilated one-dimensional convolutions. Implemented using the authors' code\footnote{https://github.com/nnzhan/Graph-WaveNet}.

\item STSGCN \cite{song2020spatial}: Captures localized spatio-temporal correlations using a synchronous modeling mechanism with modules for different time periods. Implemented using the authors' code\footnote{https://github.com/Davidham3/STSGCN}.

\item AGCRN\cite{jiang2023spatio}: Combines adaptive learning modules with recurrent networks to autonomously capture detailed spatial and temporal traffic correlations. Implemented using the authors' code\footnote{https://github.com/LeiBAI/AGCRN}.

\item STGODE \cite{fang2021spatial}: Captures spatio-temporal dynamics with tensor-based ODEs, constructing deeper networks that utilize these features. Implemented using the authors' code\footnote{https://github.com/square-coder/STGODE}.

\item MegaCRN \cite{jiang2023spatio}: Combines a meta graph learner with a GCRN encoder-decoder architecture, effectively handling varied road patterns and adapting to abnormal traffic conditions. Implemented using the authors' code\footnote{https://github.com/deepkashiwa20/MegaCRN}.

\end{itemize}
\begin{table*}[ht]
  \centering
  \vspace{-0.55cm}
  \caption{Comparisons on PeMS dataset}
    \begin{tabular}{cccccccccccccc}
    \toprule
    Model & Horizon & \multicolumn{4}{c}{3} & \multicolumn{4}{c}{6} & \multicolumn{4}{c}{12} \\
\cmidrule{2-14}    Type & Metric & MAE & RMSE & MAPE & Cost & MAE & RMSE & MAPE & Cost & MAE & RMSE & MAPE & Cost \\
    \midrule
    \multirow{14}[4]{*}{\begin{sideways}Lane-Level\end{sideways}} & Cat-RF-LSTM & 7.75  & 10.68  & 81.50\% & $-$ & 7.81  & 11.00  & 81.92\% & $-$ & 8.30  & 11.90  & 83.47\% & $-$ \\
        & CEEMDAN-XGBoost & 7.41  & 10.17  & 76.28\% & $-$ & 7.44  & 10.42  & 76.55\% & $-$ & 8.09  & 10.73  & 78.90\% & $-$ \\
        & LSTM & 7.11  & 9.94  & 54.93\% & 0.61  & 7.44  & 10.47  & 56.77\% & $\underline{0.73}$ & 7.93  & 11.08  & 59.04\% & 0.94  \\
        & GRU & 6.75  & 9.51  & 43.66\% & $\underline{0.52}$ & 7.14  & 10.10  & 45.08\% & 0.74  & 7.72  & 10.84  & 47.14\% & $\underline{0.83}$ \\
        & FDL & 6.79  & 9.59  & 43.56\% & 1.45  & 7.20  & 10.18  & 45.21\% & 2.52  & 7.84  & 11.02  & 47.34\% & 2.58  \\
        & TM-CNN & 4.77  & 7.54  & 23.19\% & 1.77  & 5.06  & 8.16  & 24.55\% & 1.80  & 5.86  & 9.39  & 28.73\% & 1.84  \\
        & MDL & $\underline{4.27}$ & $\underline{6.95}$ & 21.45\% & 4.60  & $\underline{4.87}$ & $\underline{7.95}$ & $\underline{24.21\%}$ & 4.75  & $\underline{5.55}$ & $\underline{9.03}$ & $\underline{28.46\%}$ & 4.79  \\
        & CNN-LSTM & 8.56  & 11.84  & 86.89\% & 2.33  & 8.69  & 11.98  & 89.42\% & 2.44  & 8.81  & 12.29  & 91.69\% & 2.58  \\
        & HGCN & 4.73  & 7.65  & 24.03\% & 2.48  & 5.22  & 8.49  & 26.03\% & 3.27  & 5.92  & 9.62  & 29.94\% & 3.47  \\
        & GCN-GRU & 4.70  & 7.65  & 23.36\% & 2.39  & 5.15  & 8.46  & 25.87\% & 2.69  & 6.14  & 9.81  & 31.80\% & 2.76  \\
        & ST-AFN & 4.43  & 7.35  & $\underline{21.38\%}$ & 2.15  & 5.11  & 8.36  & 25.53\% & 2.22  & 6.02  & 9.87  & 29.09\% & 2.85  \\
        & STA-ED & 6.84  & 9.78  & 44.40\% & 2.24  & 7.44  & 10.56  & 47.02\% & 2.54  & 7.77  & 10.98  & 47.35\% & 2.62  \\
        & STMGG & 6.93  & 9.10  & 42.75\% & 1.70  & 7.74  & 10.58  & 52.00\% & 1.98  & 7.87  & 11.14  & 55.47\% & 2.53  \\
\cmidrule{3-14}        & Mean & 6.40  & 9.23  & 47.17\% & 2.17  & 6.79  & 9.90  & 49.66\% & 2.50  & 7.36  & 10.72  & 52.50\% & 2.70  \\
    \midrule
    \multirow{10}[4]{*}{\begin{sideways}ST-Graph\end{sideways}} & DCRNN & 5.54  & 8.19  & 28.26\% & 9.56  & 6.79  & 9.83  & 32.74\% & 13.53  & 7.49  & 10.65  & 36.35\% & 17.41  \\
        & STGCN & 5.41  & 8.83  & 27.13\% & $\underline{1.81}$ & 6.16  & 10.07  & 30.53\% & $\underline{3.14}$ & 7.10  & 11.69  & 37.79\% & $\underline{3.53}$ \\
        & MTGNN & 4.80  & 7.85  & 23.36\% & 4.48  & 5.42  & 9.06  & 26.71\% & 4.60  & 6.33  & 10.44  & 32.53\% & 4.98  \\
        & ASTGCN & 5.23  & 7.46  & 22.05\% & 3.20  & 5.74  & 8.65  & 26.62\% & 3.21  & 6.75  & 9.79  & 33.98\% & 3.28  \\
        & GraphWaveNet & 4.20  & 6.25  & 17.48\% & 5.18  & 4.37  & 6.94  & $\underline{17.78\%}$ & 5.35  & 5.44  & 8.49  & 24.72\% & 5.40  \\
        & STSGCN & 4.16  & 6.07  & $\underline{17.28\%}$ & 5.95  & 4.44  & 6.81  & 18.83\% & 7.83  & 5.10  & 7.99  & $\underline{22.65\%}$ & 8.67  \\
        & AGCRN & 4.18  & 6.59  & 19.75\% & 9.69  & 4.75  & 7.66  & 22.98\% & 11.28  & 5.57  & 9.00  & 27.30\% & 12.06  \\
        & STGODE & $\underline{4.14}$ & $\underline{6.04}$ & 17.40\% & 8.53  & $\underline{4.37}$ & $\underline{6.74}$ & 18.57\% & 9.60  & $\underline{5.09}$ & $\underline{7.94}$ & 22.93\% & 12.32  \\
        & MegaCRN & 4.73  & 7.56  & 20.66\% & 6.61  & 5.65  & 9.13  & 28.70\% & 6.84  & 7.73  & 12.05  & 39.50\% & 7.36  \\
\cmidrule{3-14}        & Mean & 4.78  & 7.35  & 22.01\% & 6.65  & 5.41  & 8.52  & 25.71\% & 7.78  & 6.44  & 10.01  & 31.89\% & 8.94  \\
    \midrule
    New & GraphMLP & \textbf{3.74 } & \textbf{5.66 } & \textbf{13.78\%} & \textbf{1.89 } & \textbf{4.05 } & \textbf{6.29 } & \textbf{16.25\%} & \textbf{2.05 } & \textbf{4.81 } & \textbf{7.24 } & \textbf{20.44\%} & \textbf{2.15 } \\
    \bottomrule
    \end{tabular}%
  \label{tab:PEMS}%
  \vspace{-0.35cm}
\end{table*}%
\subsection{Experimental Setups}
All experiments were conducted on a computing platform equipped with an Intel (R) Xeon (R) Gold 6278C CPU @ 2.60 GHz and eight NVIDIA GeForce RTX 2080 GPUs. For the primary experiments, the input window was set to encompass 12 timestamps, with the prediction horizon lengths individually set at 3, 6, and 12 timestamps. The GraphMLP model, along with all comparative baseline models, employed the Adam optimizer and underwent up to 1,000 training iterations. To prevent overfitting, an early stopping strategy based on validation set performance was implemented. The initial learning rate for model training was set at 0.001, and starting from the 20th training epoch, the learning rate was reduced to half of its value every 10 epochs.

\subsection{Lane-Level Traffic Benchmark}
This experiment represents the first unified evaluation of lane-level traffic prediction models. We conducted a fair comparison of 14 models for lane-level traffic prediction, including GraphMLP, and 9 graph-structured spatio-temporal prediction models on the PeMS and HuaNan datasets. The average performance of both lane-level models and spatio-temporal graph models was calculated, with the best-performing models in each category highlighted in \underline{underline} , and the overall best-performing model across all categories emphasized in \textbf{bold}. It should be noted that the Cat-RF-LSTM and CEEMDAN-XGBoost models, which require phased training, were not suitable for inclusion in the Cost metric analysis.

In \textbf{Table \ref{tab:PEMS}},  we present the experimental results on the freeway dataset PeMS. From the average values, we observe that the predictive performance of graph-structured models generally surpasses the average of lane-level models, indicating the feasibility of transferring graph-structured models to lane-level traffic prediction. However, the average training time for lane-level traffic prediction models is less than that for graph-structured models. This is attributed to the typically simpler structure of lane-level models, designed to adapt to the highly dynamic nature of lane traffic, presenting challenges for the direct transfer of graph-structured models to lane-level prediction. Among lane-level prediction models, MDL and ST-AFN, which utilized the authors' code and experimental configurations, performed relatively well. Unexpectedly, the FDL model performed worse than LSTM and GRU, most likely due to the ineffective computation of its entropy-based grey relational algorithm. For graph-structured models, STGODE, STSGCN, GraphWaveNet, and AGCRN all demonstrated commendable performance. Notably, STSGCN performed slightly better in the MAPE metric, indicating its effectiveness in predicting smaller actual values. STGCN was the fastest training model among graph-structured models, benefiting from its fully convolutional architecture. Overall, the GraphMLP model exhibited the best performance, surpassing the most effective STGODE model among all baselines in terms of prediction accuracy. Although its training time was not as short as some lane-level models, its higher predictive accuracy demonstrated its superior predictive effectiveness and efficiency.
\begin{table*}[htbp]
  \centering
  \vspace{-0.65cm}
  \caption{Comparisons on HuaNan dataset}
  \vspace{-0.15cm}
   \begin{tabular}{cccccccccccccc}
    \toprule
    Model & Horizon & \multicolumn{4}{c}{3}         & \multicolumn{4}{c}{6}         & \multicolumn{4}{c}{12} \\
\cmidrule{2-14}    Type  & Metric & MAE   & RMSE  & MAPE  & Cost  & MAE   & RMSE  & MAPE  & Cost  & MAE   & RMSE  & MAPE  & Cost \\
    \midrule
    \multirow{14}[4]{*}{\begin{sideways}Lane-Level\end{sideways}} & Cat-RF-LSTM & 18.17  & 22.71  & 93.50\% & $-$   & 18.30  & 22.89  & 93.67\% & $-$   & 18.67  & 23.21  & 99.23\% & $-$ \\
          & CEEMDAN-XGBoost & 17.01  & 21.24  & 81.31\% & $-$   & 17.09  & 21.58  & 80.48\% & $-$   & 17.69  & 22.51  & 85.50\% & $-$ \\
          & LSTM  & 17.04  & 21.72  & 80.98\% & 1.26  & 17.12  & 21.91  & 82.14\% & 1.30  & 17.21  & 22.14  & 82.60\% & 1.45  \\
          & GRU   & 16.41  & 21.03  & 72.80\% & $\underline{0.89}$ & 16.56  & 21.32  & 74.12\% & $\underline{1.14}$ & 16.60  & 21.34  & 74.79\% & $\underline{1.26}$ \\
          & FDL   & 16.36  & 20.98  & 72.48\% & 5.17  & 16.73  & 21.58  & 75.09\% & 4.21  & 16.73  & 21.69  & 76.21\% & 1.47  \\
          & TM-CNN & 13.68  & 18.74  & 51.23\% & 4.91  & 14.12  & 19.56  & 56.23\% & 5.19  & 15.74  & 21.32  & 74.70\% & 5.36  \\
          & MDL   & $\underline{3.63}$ & $\underline{5.86}$ & $\underline{18.74\%}$ & 5.22  & $\underline{5.86}$ & $\underline{9.01}$ & $\underline{29.63\%}$ & 5.31  & $\underline{8.37}$ & $\underline{12.55}$ & $\underline{31.79\%}$ & 5.77  \\
          & CNN-LSTM & 12.52  & 18.34  & 64.03\% & 8.03  & 12.78  & 18.46  & 66.60\% & 8.34  & 13.79  & 19.48  & 73.98\% & 8.75  \\
          & HGCN  & 11.00  & 15.46  & 34.73\% & 2.65  & 11.24  & 15.68  & 36.04\% & 2.79  & 11.67  & 16.12  & 37.71\% & 3.12  \\
          & GCN-GRU & 10.98  & 15.39  & 34.58\% & 2.58  & 11.28  & 15.76  & 36.30\% & 2.63  & 11.74  & 16.23  & 37.70\% & 2.87  \\
          & ST-AFN & 5.11  & 9.19  & 19.91\% & 3.00  & 7.59  & 12.21  & 32.39\% & 3.34  & 9.08  & 13.82  & 35.54\% & 3.54  \\
          & STA-ED & 13.55  & 18.36  & 62.28\% & 2.86  & 14.30  & 18.97  & 64.47\% & 3.45  & 14.98  & 19.56  & 67.22\% & 3.62  \\
          & STMGG & 14.24  & 17.88  & 69.66\% & 4.92  & 15.06  & 19.03  & 73.77\% & 7.84  & 16.05  & 19.84  & 75.62\% & 8.85  \\
\cmidrule{3-14}          & Mean  & 13.84  & 18.42  & 61.46\% & 4.06  & 14.35  & 19.08  & 64.27\% & 4.44  & 15.00  & 19.77  & 68.40\% & 4.48  \\
    \midrule
    \multirow{10}[4]{*}{\begin{sideways}ST-Graph\end{sideways}} & DCRNN & 13.85  & 18.32  & 57.04\% & 11.79  & 14.95  & 19.98  & 67.18\% & 24.09  & 16.18  & 21.24  & 72.14\% & 12.81  \\
          & STGCN & 13.70  & 18.31  & 55.69\% & $\underline{3.66}$ & 13.99  & 19.19  & 57.47\% & $\underline{4.81}$ & 15.08  & 20.70  & 54.96\% & $\underline{5.07}$ \\
          & MTGNN & 9.11  & 13.38  & 29.18\% & 4.82  & 10.95  & 14.90  & 40.42\% & 5.19  & 11.26  & 15.55  & 34.44\% & 5.49  \\
          & ASTGCN & 12.26  & 16.18  & 46.72\% & 5.06  & 12.74  & 16.70  & 50.65\% & 5.36  & 14.87  & 19.51  & 58.06\% & 5.63  \\
          & GraphWaveNet & 10.27  & 15.19  & 35.91\% & 6.11  & 11.38  & 16.75  & 41.39\% & 6.84  & 12.25  & 18.05  & 45.15\% & 7.10  \\
          & STSGCN & 10.85  & 15.28  & 37.17\% & 8.56  & 11.76  & 16.28  & 40.49\% & 8.75  & 12.47  & 17.05  & 44.70\% & 89.53  \\
          & AGCRN & $\underline{4.11}$ & $\underline{6.08}$ & $\underline{16.72\%}$ & 10.47  & $\underline{5.28}$ & $\underline{7.49}$ & $\underline{21.70\%}$ & 24.04  & $\underline{6.47}$ & $\underline{10.75}$ & $\underline{24.81\%}$ & 13.22  \\
          & STGODE & 10.47  & 14.84  & 35.67\% & 11.07  & 11.14  & 15.56  & 37.87\% & 12.85  & 11.83  & 16.27  & 41.36\% & 14.04  \\
          & MegaCRN & 13.15  & 18.32  & 42.60\% & 9.57  & 14.27  & 20.30  & 48.49\% & 10.28  & 16.00  & 23.05  & 58.99\% & 13.68  \\
\cmidrule{3-14}          & Mean  & 11.71  & 16.23  & 42.50\% & 8.43  & 12.65  & 17.46  & 47.99\% & 12.17  & 13.74  & 18.93  & 51.22\% & 20.19  \\
    \midrule
    New   & GraphMLP & \textbf{2.99 } & \textbf{4.03 } & \textbf{11.53\%} & \textbf{2.97 } & \textbf{3.88 } & \textbf{7.77 } & \textbf{17.03\%} & \textbf{3.09 } & \textbf{4.88 } & \textbf{9.00 } & \textbf{21.20\%} & \textbf{3.28 } \\
    \bottomrule
    \end{tabular}%
  \label{tab:HuaNan}%
  \vspace{-0.35cm}
\end{table*}%

\textbf{Table \ref{tab:HuaNan}} demonstrates the performance of various models on the urban expressway HuaNan dataset. Due to unit differences, the actual values in the HuaNan dataset are higher than those in the PeMS dataset, resulting in a significant amplification of errors. Observing the average performance, similar to the PeMS dataset, spatio-temporal graph-structured models outperform lane-level prediction models in terms of average prediction accuracy, yet require longer training times. The magnified errors highlight the distinctions between models more conspicuously. Among lane-level prediction models, MDL and ST-AFN remain the top performers, with MDL showing particularly notable performance. In contrast to the PeMS dataset, spatio-temporal graph-structured models exhibited some variances, where the AGCRN model surpassed STGODE, STSGCN, and GraphWave Net models across all three prediction horizons and metrics, emerging as the top-performing model. The MTGNN model also demonstrated commendable performance, while STGCN continued to have the least training time consumption. In a comprehensive assessment, GraphMLP maintained superior performance over all other baselines, exceeding the prediction accuracy of both the MDL and AGCRN models while requiring less training time. Moreover, GraphMLP showed more stable long-term predictive performance at horizons of 6 and 12, with its advantage in training time efficiency becoming even more pronounced. Compared to the results on the PeMS dataset, the higher computational cost reflects the greater complexity of urban roads, which demands more refined spatio-temporal modeling capabilities from the models.

\begin{table}[htbp]
  \centering
  \caption{Comparisons on PeMSF dataset}
    \begin{tabular}{lcccc}
    \toprule
    \multicolumn{1}{c}{Metric} & MAE & RMSE & MAPE & Difference \\
    \midrule
    Cat-RF-LSTM & 7.84  & 11.02  & 82.52\% & 0.28\% \\
    CEEMDAN-XGBoost & 7.48  & 10.51  & 77.56\% & 0.57\% \\
    LSTM & 7.42  & 10.45  & 55.40\% & -0.29\% \\
    GRU & 7.11  & 10.03  & 43.71\% & -0.44\% \\
    FDL & 7.17  & 10.02  & 45.12\% & -0.37\% \\
    HGCN & 5.20  & 8.49  & 26.30\% & -0.32\% \\
    GCN-GRU & 5.10  & $\underline{8.27}$  & 25.65\% & -1.15\% \\
    ST-AFN & $\underline{5.03}$  & 8.29  & $\underline{23.56\%}$ & -1.67\% \\
    STA-ED & 7.28  & 10.29  & 45.79\% & $\underline{-2.28\%}$ \\
    STMGG & 7.83  & 10.82  & 52.44\% & 1.13\% \\
\cmidrule{2-5}    Mean & 6.72  & 9.81  & 47.54\% & -0.45\% \\
    \midrule
    DCRNN & 6.87  & 10.05  & 33.02\% & 1.13\% \\
    STGCN & 6.26  & 10.16  & 33.09\% & 1.64\% \\
    MTGNN & 5.29  & 8.66  & 25.61\% & $\underline{-2.32\%}$ \\
    ASTGCN & 5.95  & 8.97  & 29.25\% & 3.57\% \\
    GraphWaveNet & $\underline{4.32}$  & 6.79  & $\underline{17.63\%}$ & -1.15\% \\
    STSGCN & 5.33  & 8.27  & 26.54\% & 1.67\% \\
    AGCRN & 4.67  & 7.65  & 22.79\% & -1.64\% \\
    STGODE & 4.39  & $\underline{6.75}$  & 18.89\% & 0.28\% \\
    MegaCRN & 5.68  & 9.20  & 28.21\% & 0.57\% \\
\cmidrule{2-5}    Mean & 5.33  & 8.37  & 25.75\% & 0.42\% \\
    \midrule
    GraphMLP & \textbf{3.90}  & \textbf{6.10}  & \textbf{15.04\%} & \textbf{-3.93\%} \\
    \bottomrule
    \end{tabular}%
  \label{tab: PeMSF}%
   \vspace{-0.25cm}
\end{table}%

\subsection{Comparison of Irregular Lane Performance}
Lane-level traffic prediction models should be capable of operating on roads with irregular lane configurations, as the majority of urban expressways in real-world scenarios exhibit such irregularities. Therefore, we conducted experiments on the complete PeMSF dataset with a prediction horizon of 6 to investigate the models' adaptability to roads with irregular lane numbers, employing the Difference metric to quantify performance disparities between the two datasets, thereby illustrating the models' adaptability to irregular lane data. The Difference metric is calculated as  \((\text{MAE}_{\text{PeMSF}} - \text{MAE}_{\text{PeMS}}) / \text{MAE}_{\text{PeMSF}}\). As depicted in \textbf{Figure \ref{fig:sensors}}, the PeMSF dataset includes three additional entrance lanes compared to the PeMS dataset. Regrettably, grid-based models such as TM-CNN and MDL cannot predict on datasets with irregular lane numbers.  

\textbf{Table \ref{tab: PeMSF}} showcases the performance of models that possess this capability. From the Difference index, we observe that most lane-level models exhibit stronger adaptability to scenarios with multiple lanes, with an average Difference index of -0.45\%, indicating an improvement in performance on the irregular lane dataset compared to regular lane numbers. The average Difference index for spatio-temporal graph-structured models is 0.42\%, suggesting that the average MAE on the irregular lane dataset, PeMSF, is larger compared to regular roads, indicating that some spatio-temporal graph models lack sufficient adaptability to irregular lane configurations. However, GraphMLP achieved a 3.93\% reduction in MAE, surpassing the lane-level model STA-ED, which reduced by 2.28\%, and the spatio-temporal graph-structured model MTGNN, which reduced by 2.32\%, making it the best-performing model. This improvement is likely due to the additional vehicles providing more valuable adjacency information to other lane segments within the dynamic graph network of the GraphMLP model.

\begin{figure}
 \vspace{-0.25cm}
\captionsetup[subfloat]{labelfont=scriptsize,textfont=scriptsize}
\centering
\hspace*{-0.7cm} 
\subfloat[MAE on PeMS dataset]{
\hspace{0.7cm}
\includegraphics[width=3.1in]{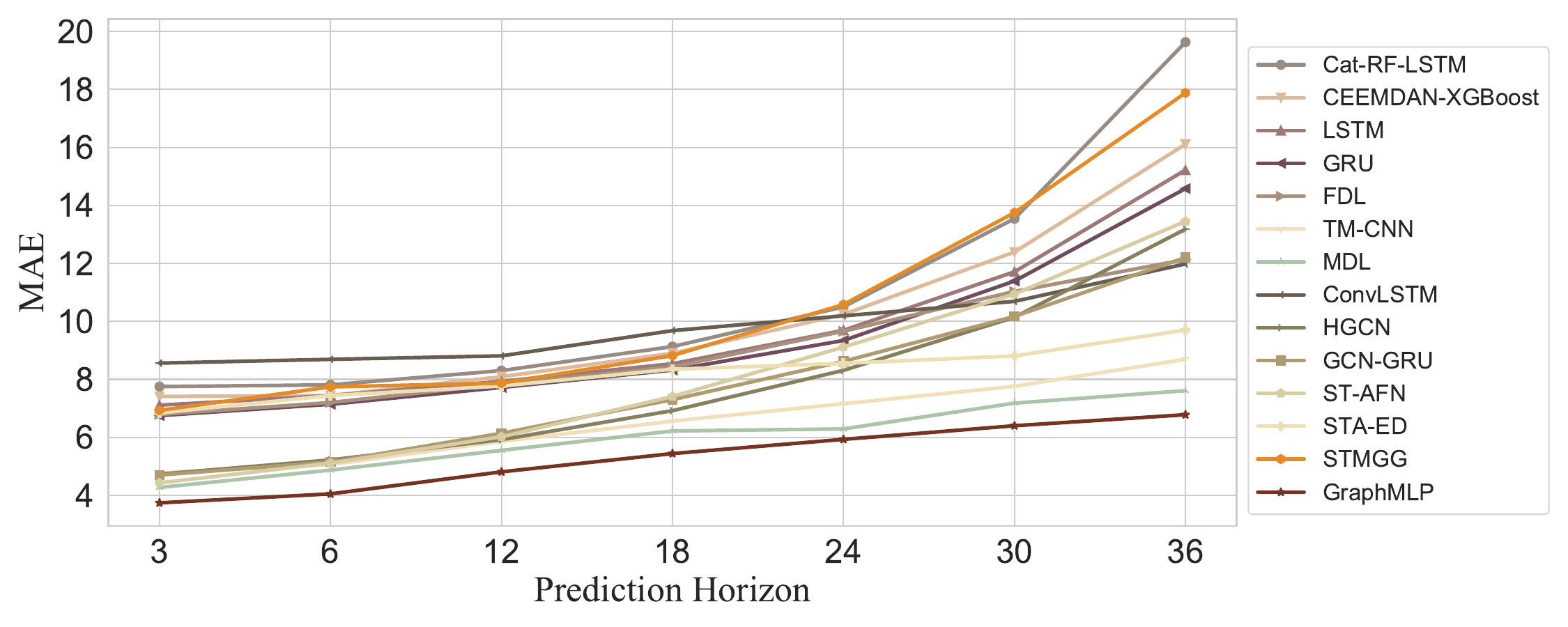}}
\newline
\hspace*{-0.7cm} 
\subfloat[RMSE on PeMS dataset]{
\hspace{0.7cm}
\includegraphics[width=3.1in]{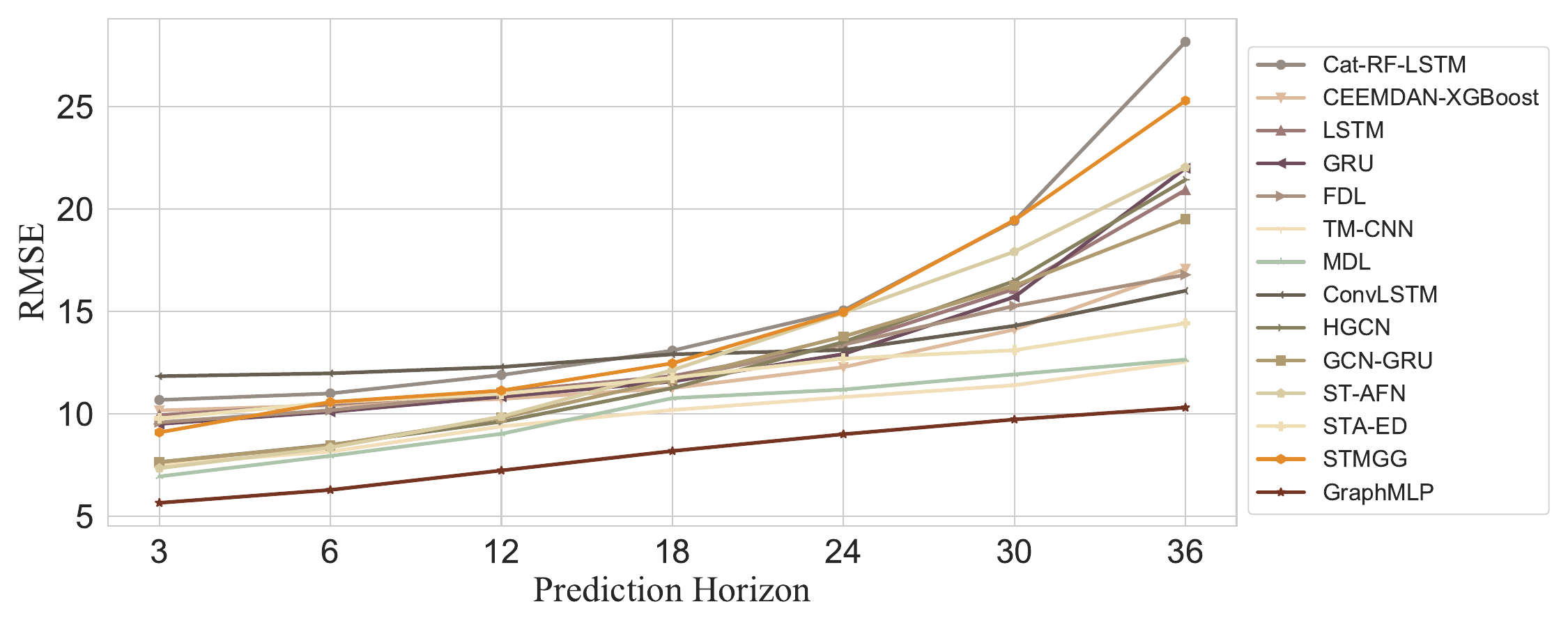}}
\newline
\hspace*{-0.7cm} 
\subfloat[MAE on HuaNan dataset]{
\hspace{0.7cm}
\includegraphics[width=3.1in]{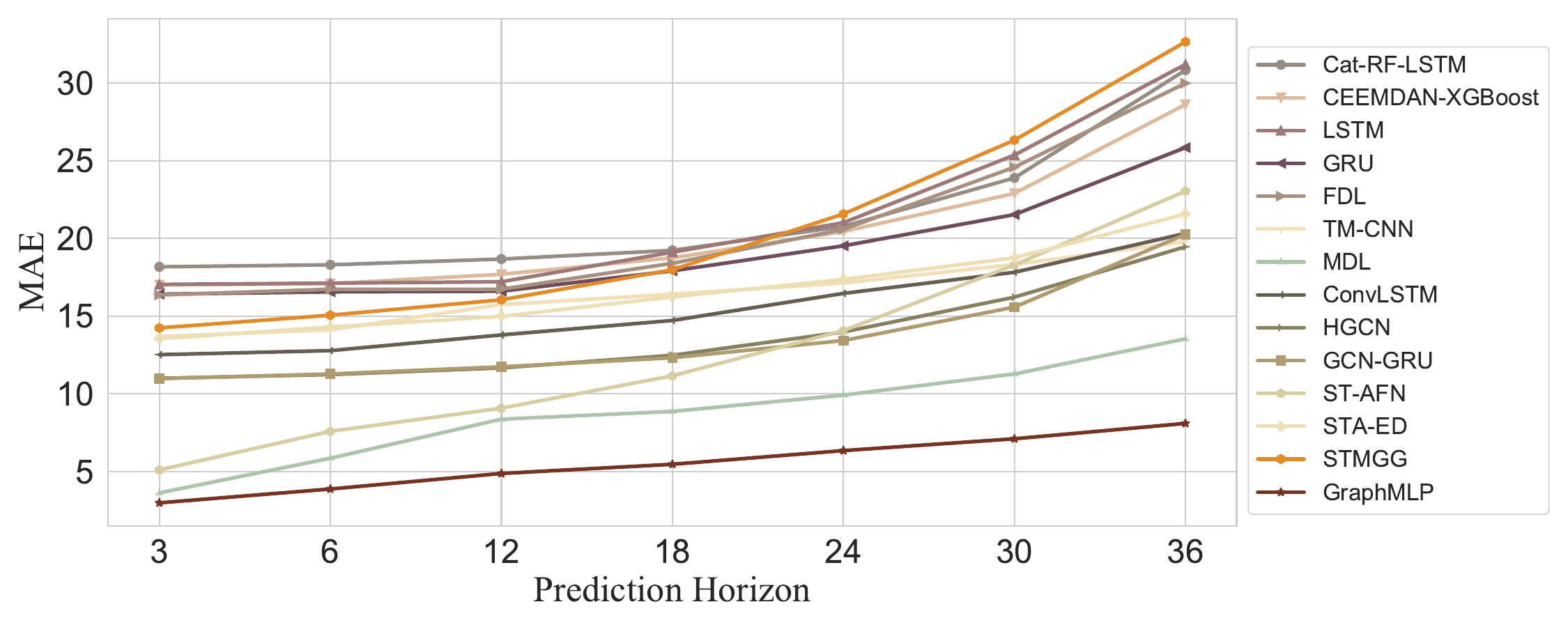}}
\newline
\hspace*{-0.7cm} 
\subfloat[RMSE on HuaNan dataset]{
\hspace{0.7cm}
\includegraphics[width=3.1in]{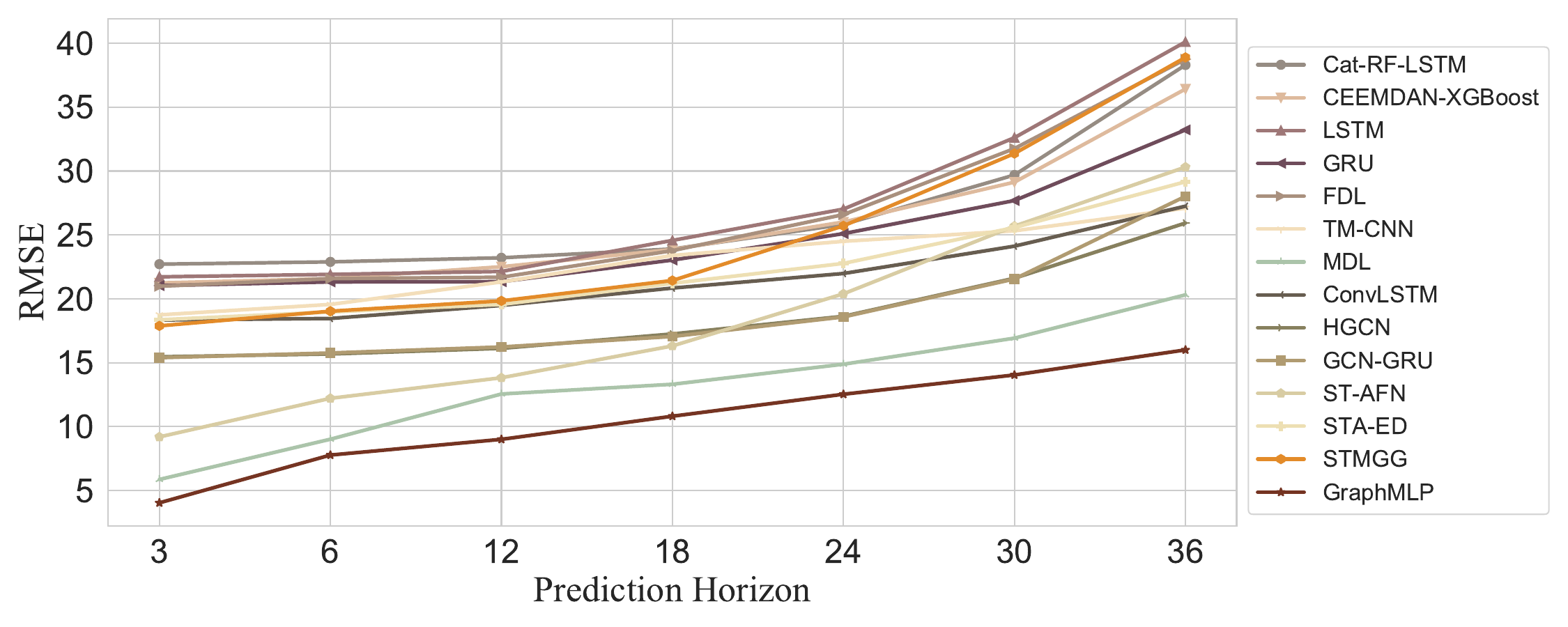}}
\caption{\label{fig:Long_term} The performance of lane-level traffic prediction models over long-term horizons.} 
 \vspace{-0.35cm}
\end{figure}

\subsection{Performance on Long-Term Prediction}
Due to the highly dynamic nature of lane networks, the three common prediction horizons (3, 6, 12) used in road-level traffic prediction are not sufficient for lane-level requirements. Therefore, the model's performance over longer prediction horizons is an important metric for lane-level traffic prediction. We conducted comparisons over seven prediction horizons (from 3 to 36) with an input window of 50 on both the PeMS and HuaNan datasets. \textbf{Figure \ref{fig:Long_term}} displays the variations in MAE and RMSE metrics on both datasets.

From subfigures (a) and (b), the changes in MAE and RMSE on the PeMS dataset can be observed. Most models show a similar trend in both metrics, with Cat-RF-LSTM and STMGG exhibiting a higher rate of change, indicating poorer performance over longer periods. Models based on LSTM and GRU are next, also showing a significant increase in error, while those based on convolution and MLP, such as TM-CNN, MDL, ConvLSTM, and GraphMLP, exhibit more gradual changes. Overall, ConvLSTM, MDL, and STA-ED have the flattest slopes, but due to their larger initial errors, their long-term performance is not as good as the GraphMLP model.

Subfigures (c) and (d) show the performance on the HuaNan dataset, where the overall trend among models is similar to that on the PeMS dataset, but with greater variability. It can be seen that on the HuaNan dataset, TM-CNN, MDL, ConvLSTM, STA-ED, and GraphMLP still maintain relatively flat slopes. However, GraphMLP shows a slight increase in the RMSE metric, yet it remains the model with the smallest prediction error.

\subsection{Effectiveness v.s. Efficiency}
\begin{figure*}[ht]
\vspace{-0.65cm}
\captionsetup[subfloat]{labelfont=scriptsize,textfont=scriptsize}
\centering
\subfloat{\includegraphics[width=1\textwidth]{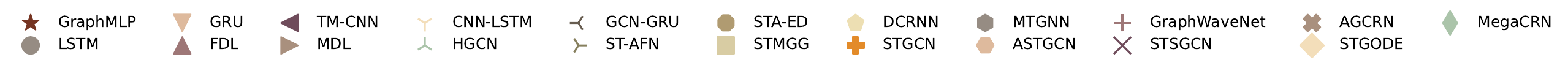}}
  \vspace{-0.25cm}
\newline
  \subfloat[Comparison of MAE on PeMS dataset \\with Horizon=3]{\includegraphics[width=0.25\textwidth]{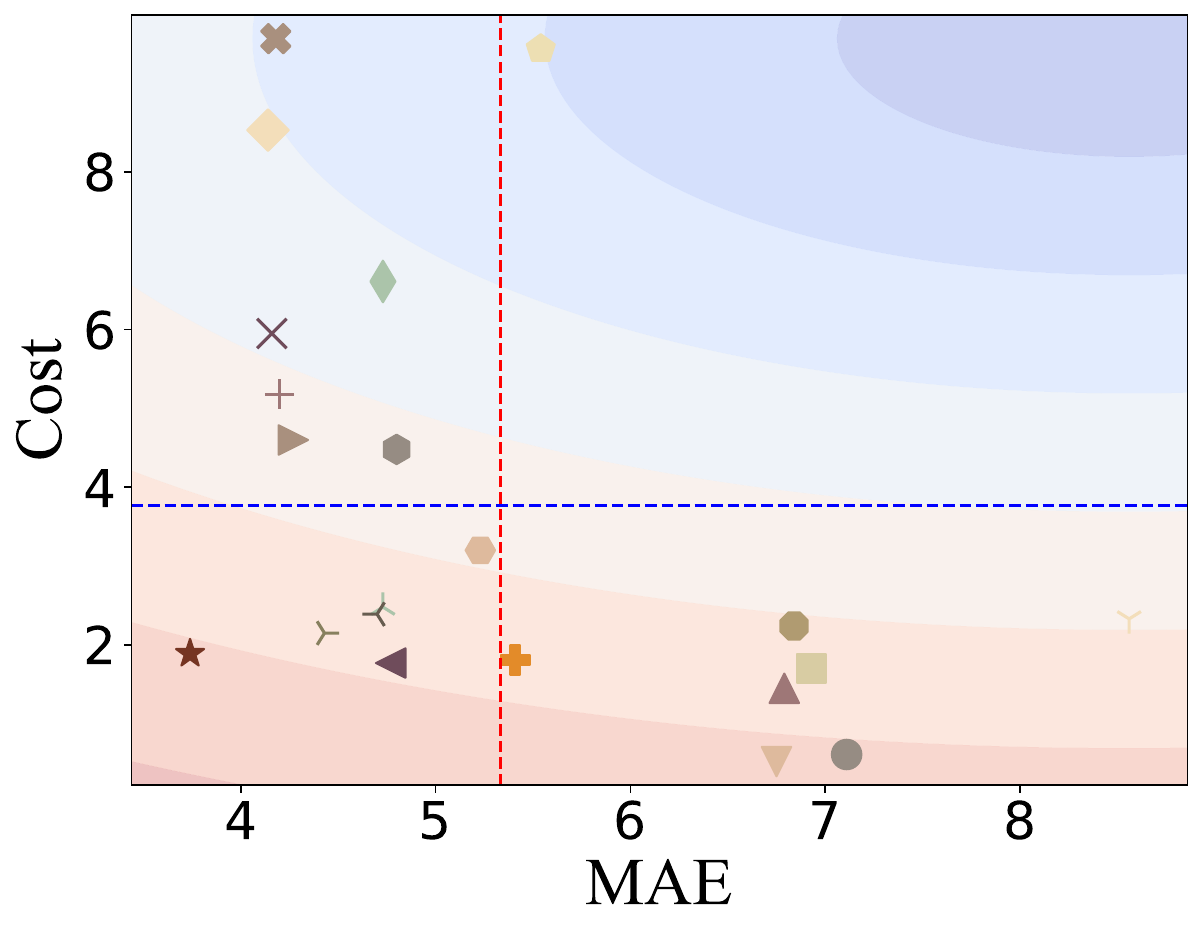}}
 \hfill 	
  \subfloat[Comparison of RMSE on PeMS dataset \\ with Horizon=3]{\includegraphics[width=0.25\textwidth]{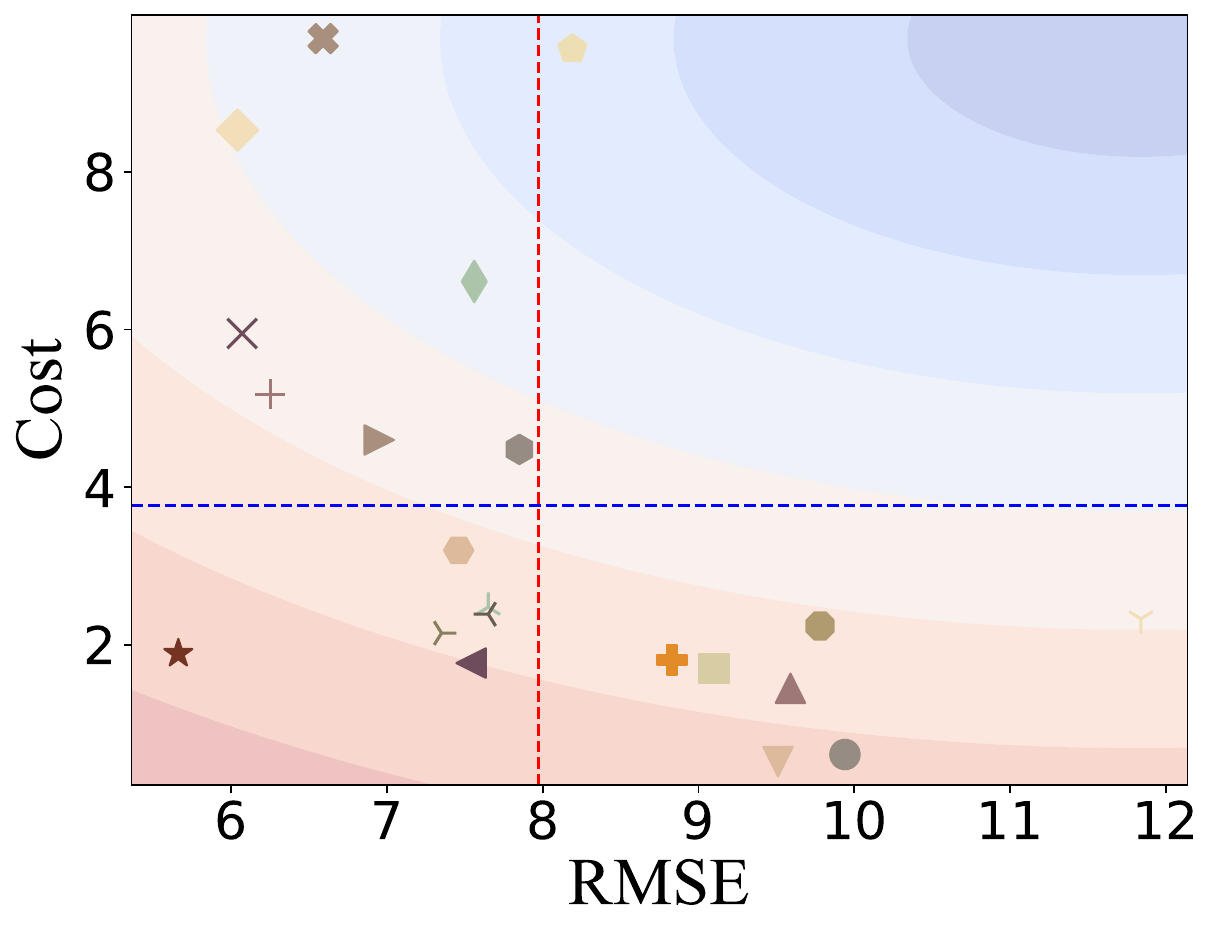}}
 \hfill	
  \subfloat[Comparison of MAE on HuaNan dataset \\with Horizon=3]{\includegraphics[width=0.25\textwidth]{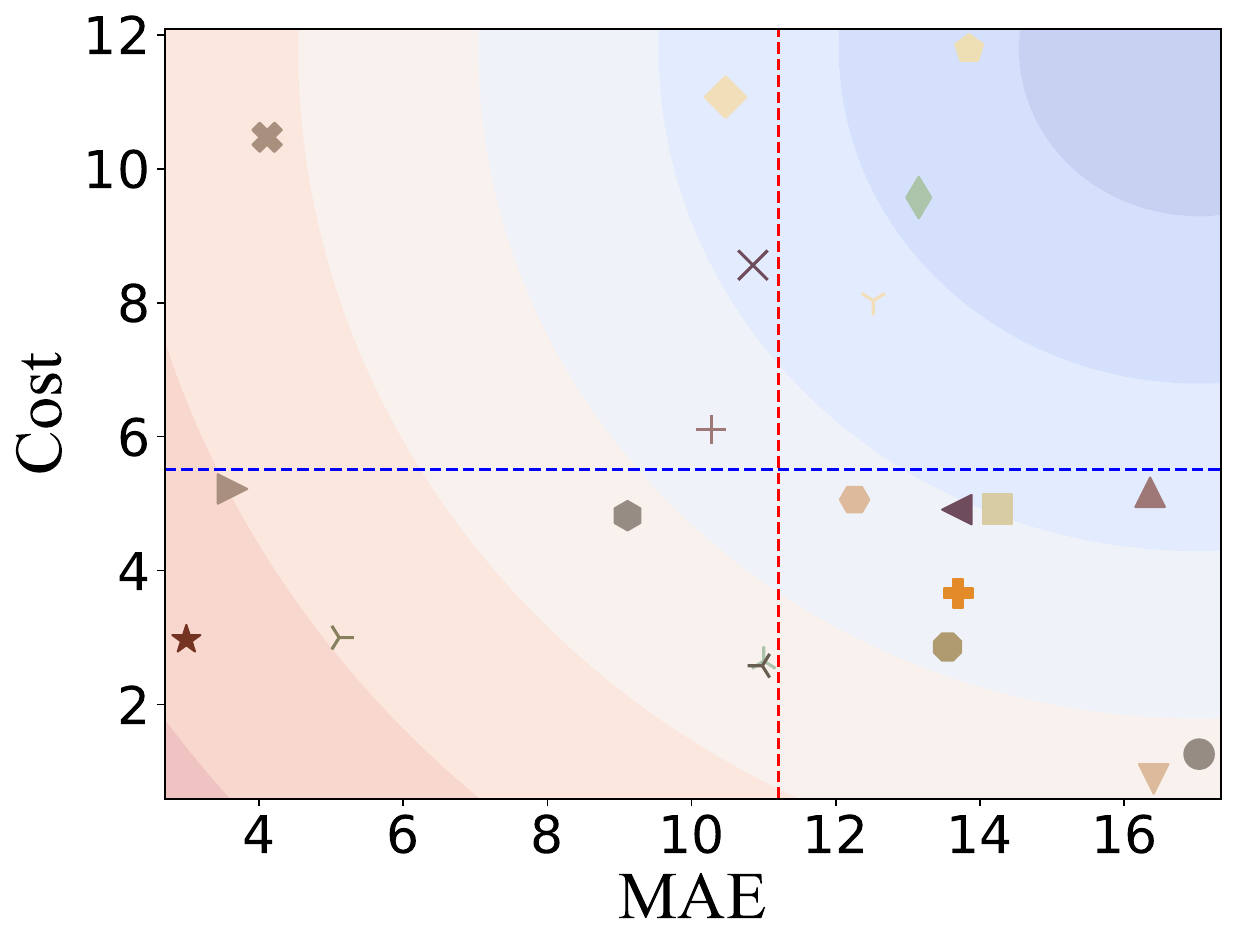}}
 \hfill 	
  \subfloat[Comparison of RMSE on HuaNan dataset \\with Horizon=3]{\includegraphics[width=0.25\textwidth]{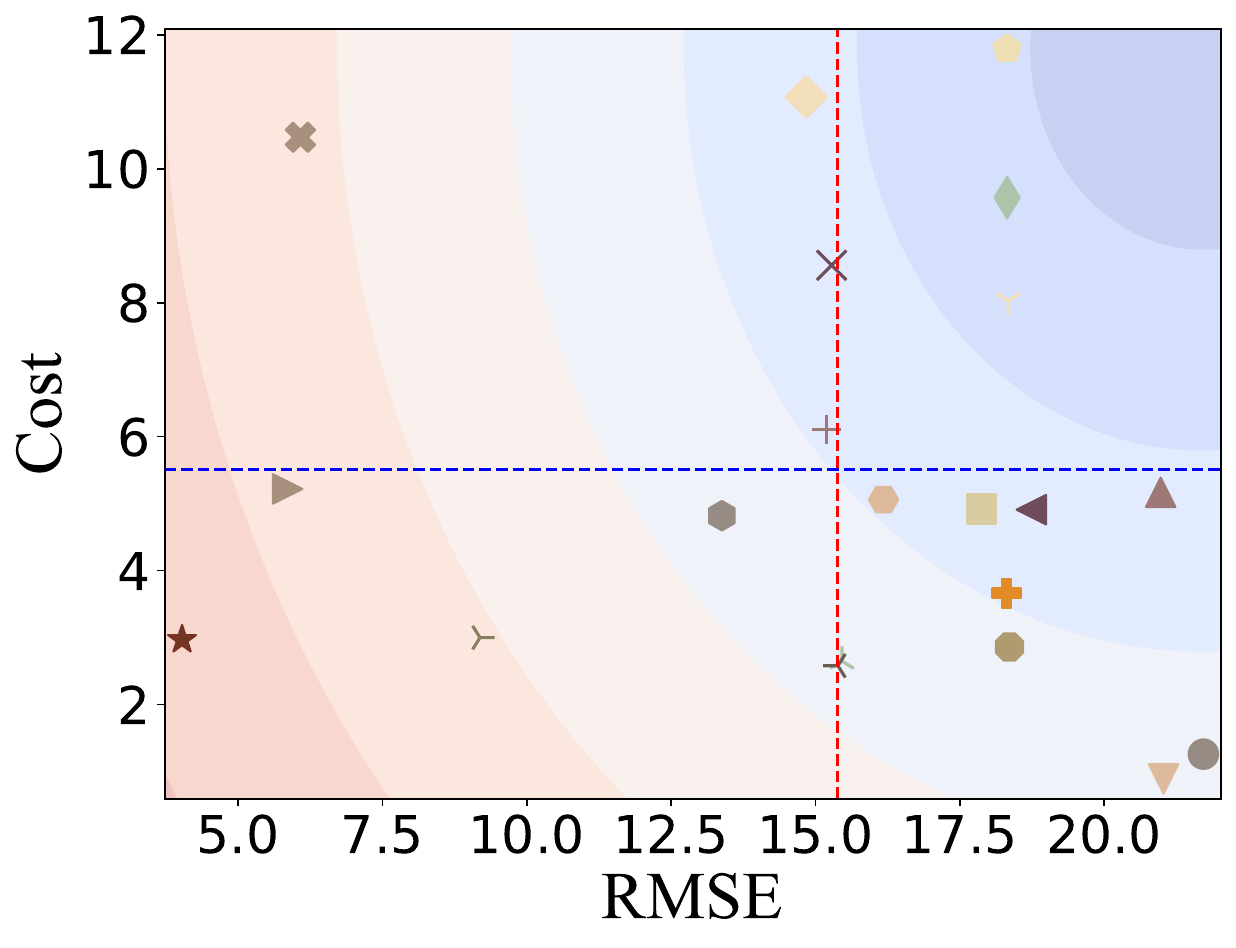}}
 \hfill
 \newline
  \subfloat[Comparison of MAE on PeMS dataset \\with Horizon=6]{\includegraphics[width=0.25\textwidth]{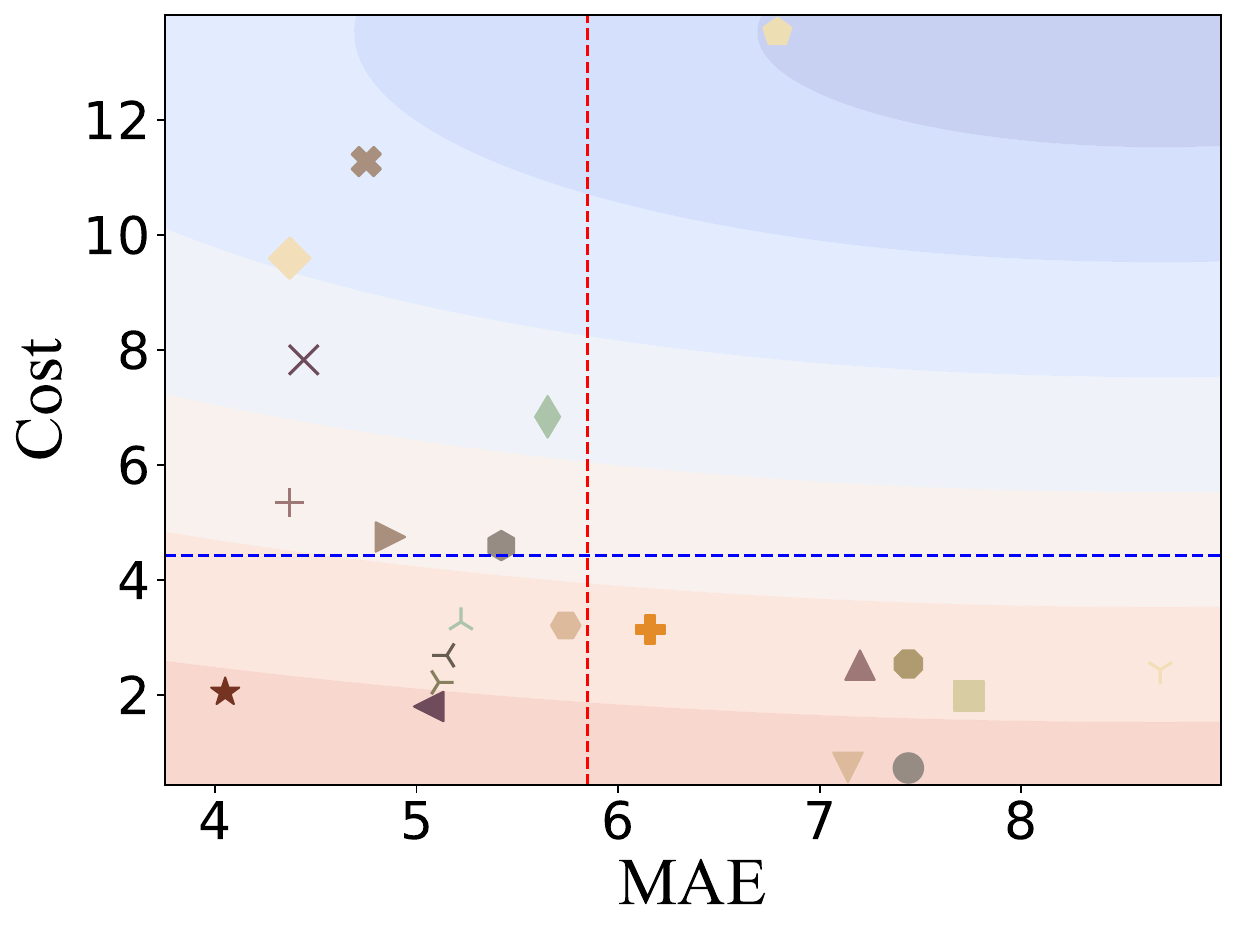}}
 \hfill 	
  \subfloat[Comparison of RMSE on PeMS dataset \\ with Horizon=6]{\includegraphics[width=0.25\textwidth]{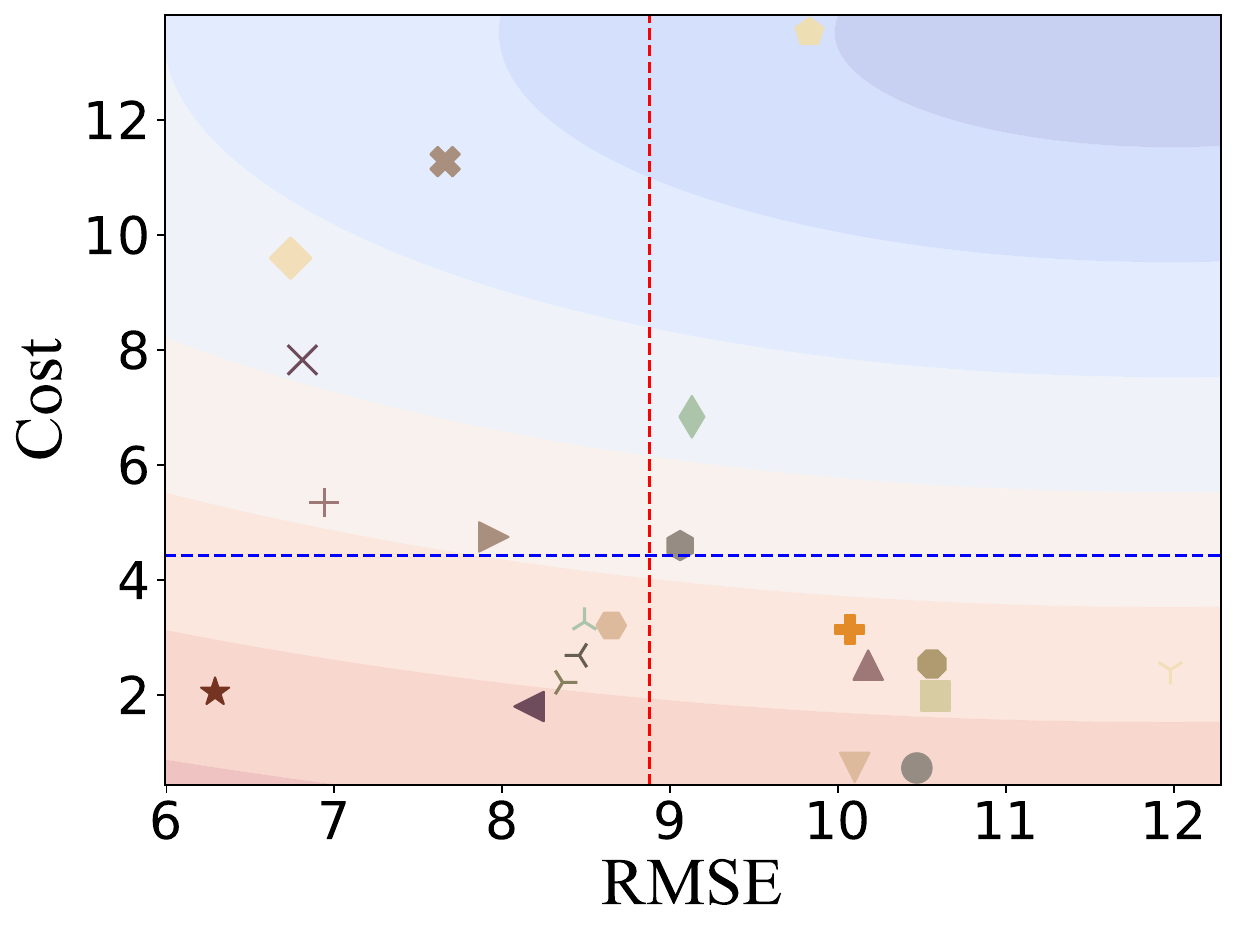}}
 \hfill	
  \subfloat[Comparison of MAE on HuaNan dataset \\with Horizon=6]{\includegraphics[width=0.25\textwidth]{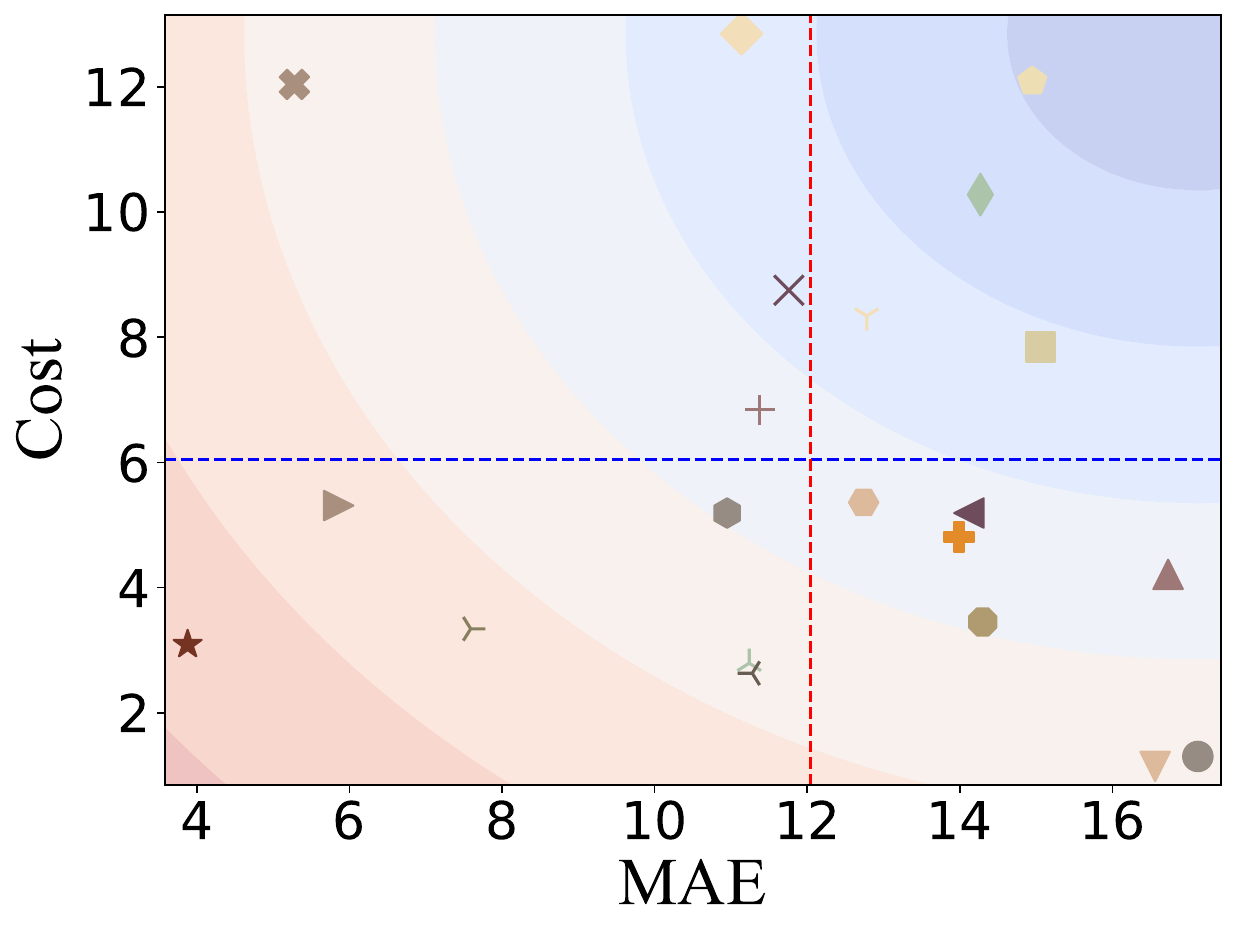}}
 \hfill 	
  \subfloat[Comparison of RMSE on HuaNan dataset \\with Horizon=6]{\includegraphics[width=0.25\textwidth]{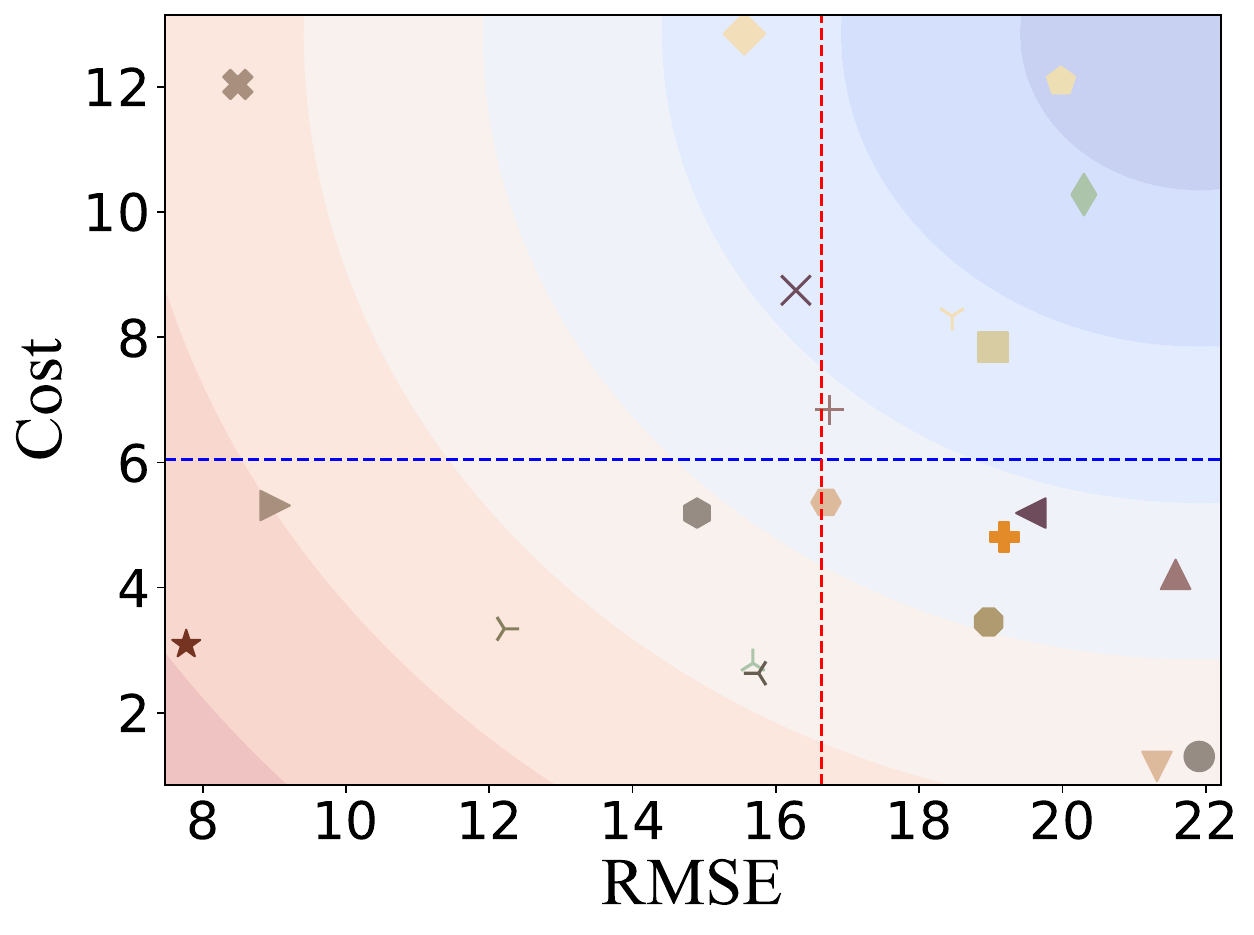}}
 \newline
 \subfloat[Comparison of MAE on PeMS dataset \\with Horizon=12]{\includegraphics[width=0.25\textwidth]{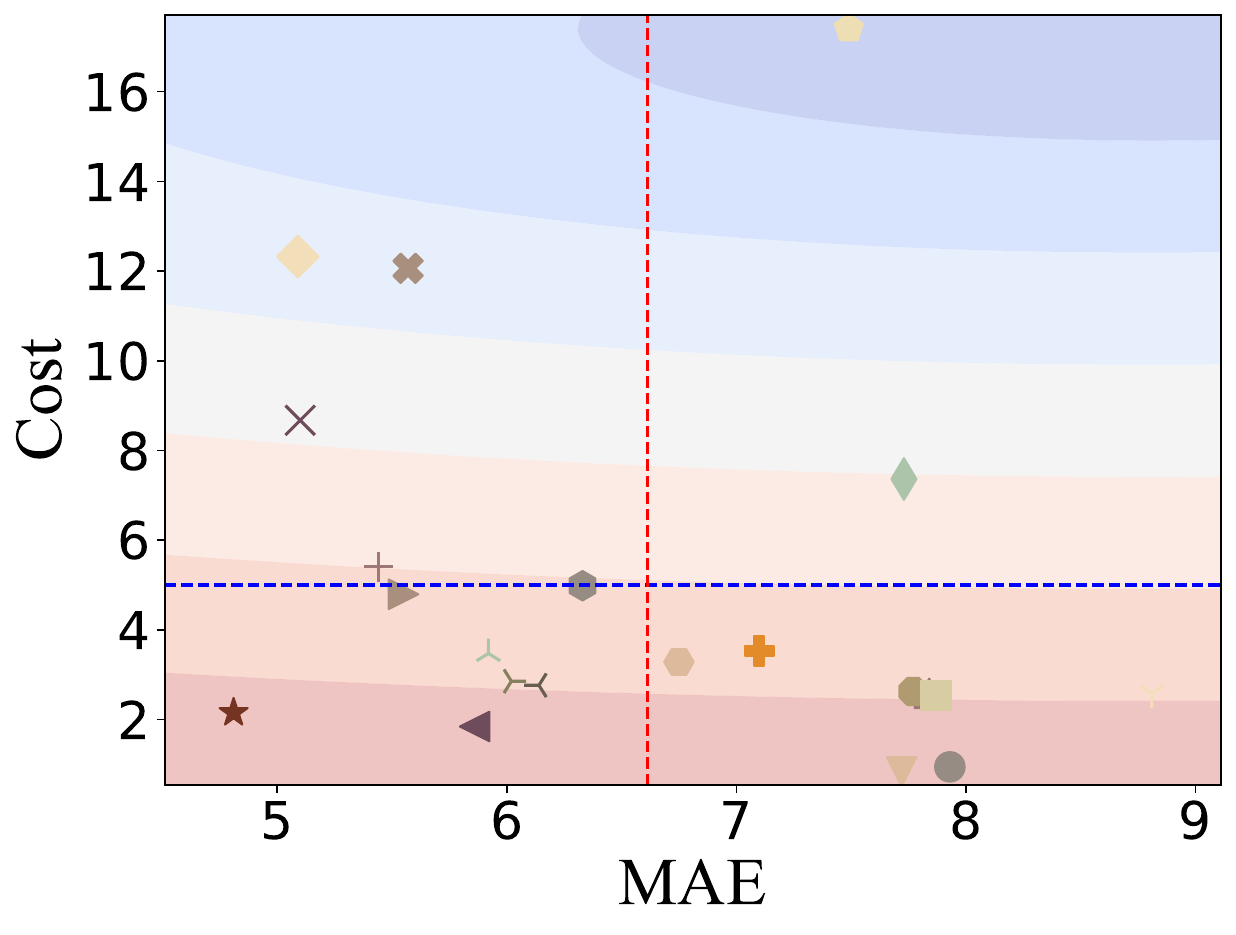}}
 \hfill 	
  \subfloat[Comparison of RMSE on PeMS dataset \\ with Horizon=12]{\includegraphics[width=0.25\textwidth]{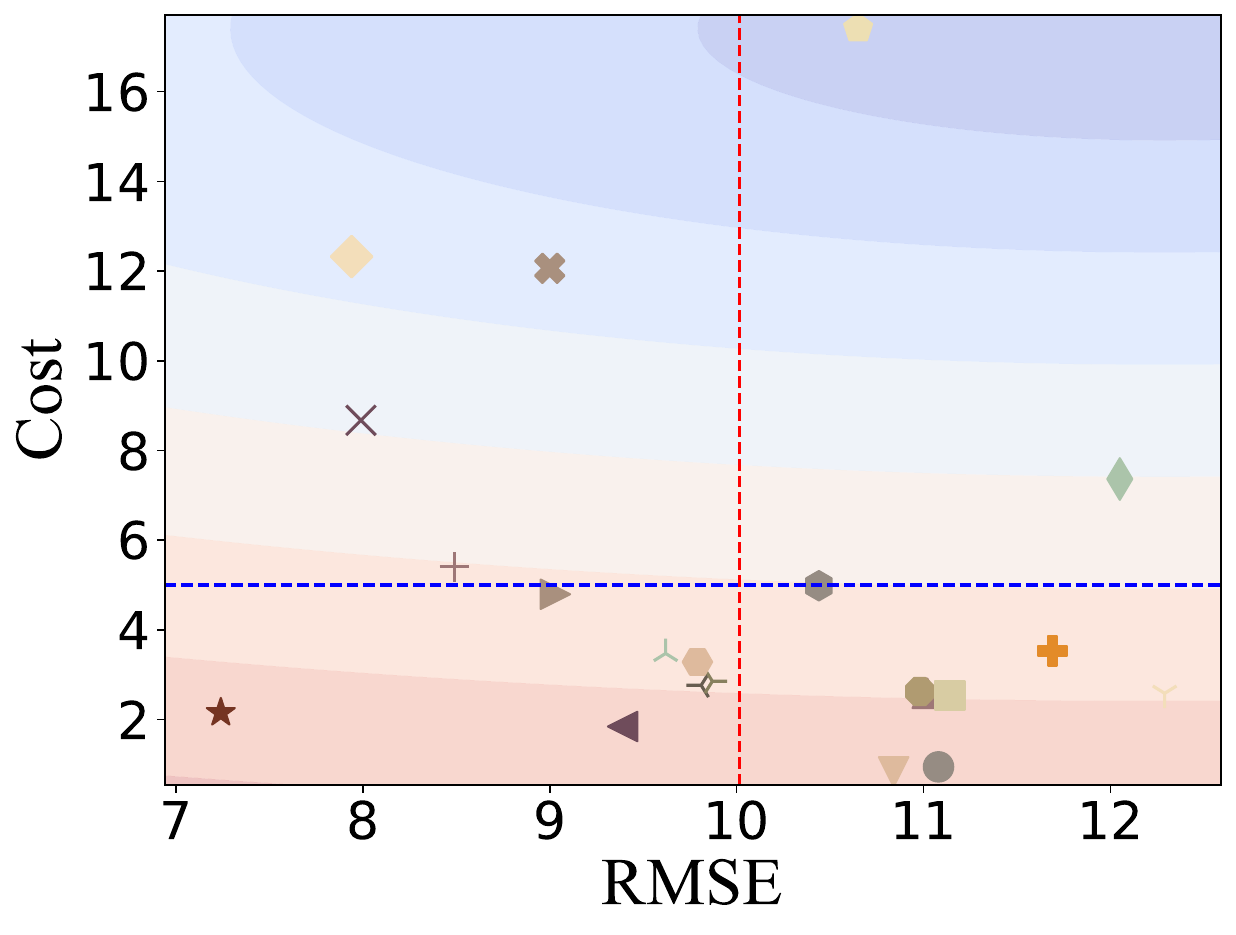}}
 \hfill	
  \subfloat[Comparison of MAE on HuaNan dataset \\with Horizon=12]{\includegraphics[width=0.25\textwidth]{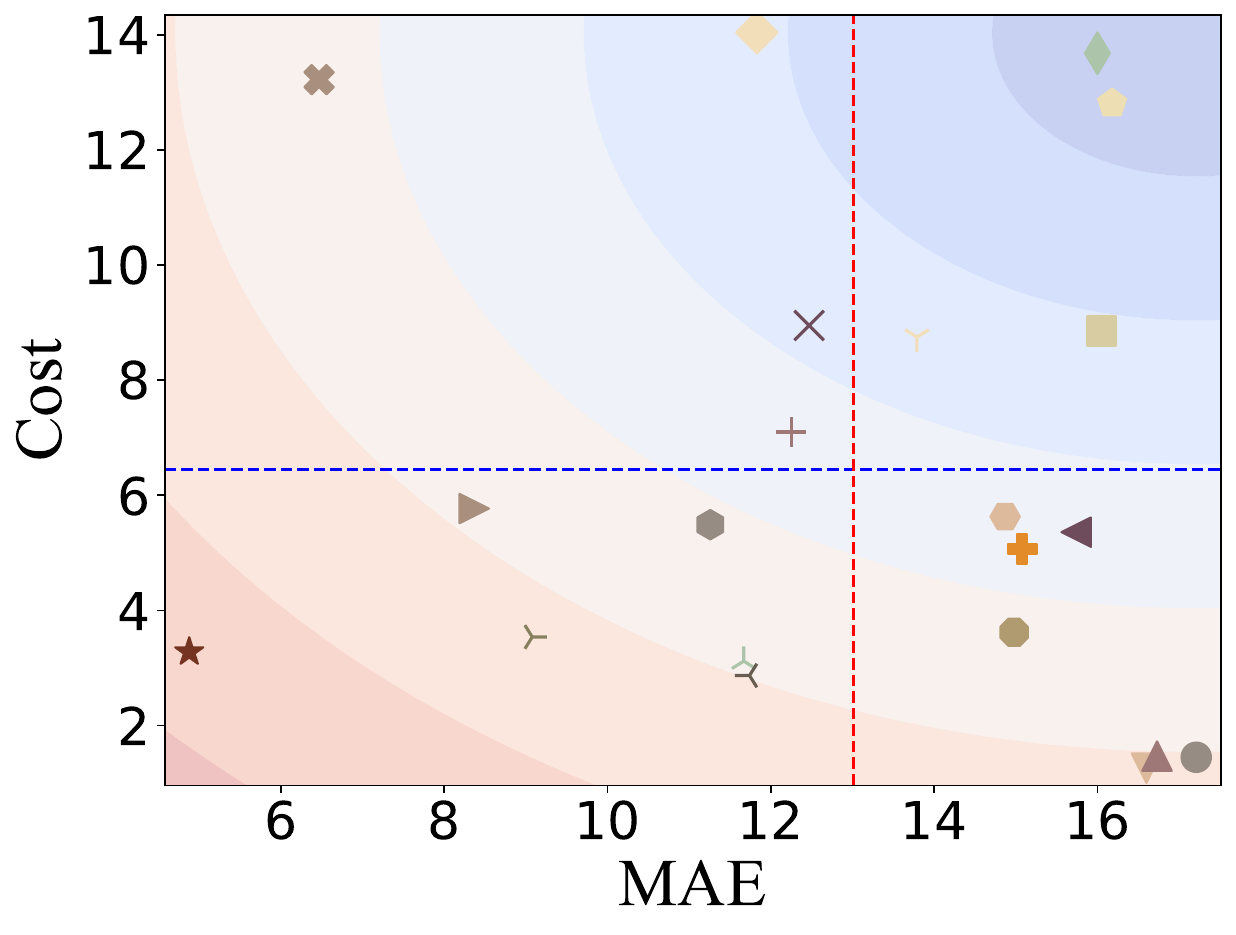}}
 \hfill 	
  \subfloat[Comparison of RMSE on HuaNan dataset \\with Horizon=12]{\includegraphics[width=0.25\textwidth]{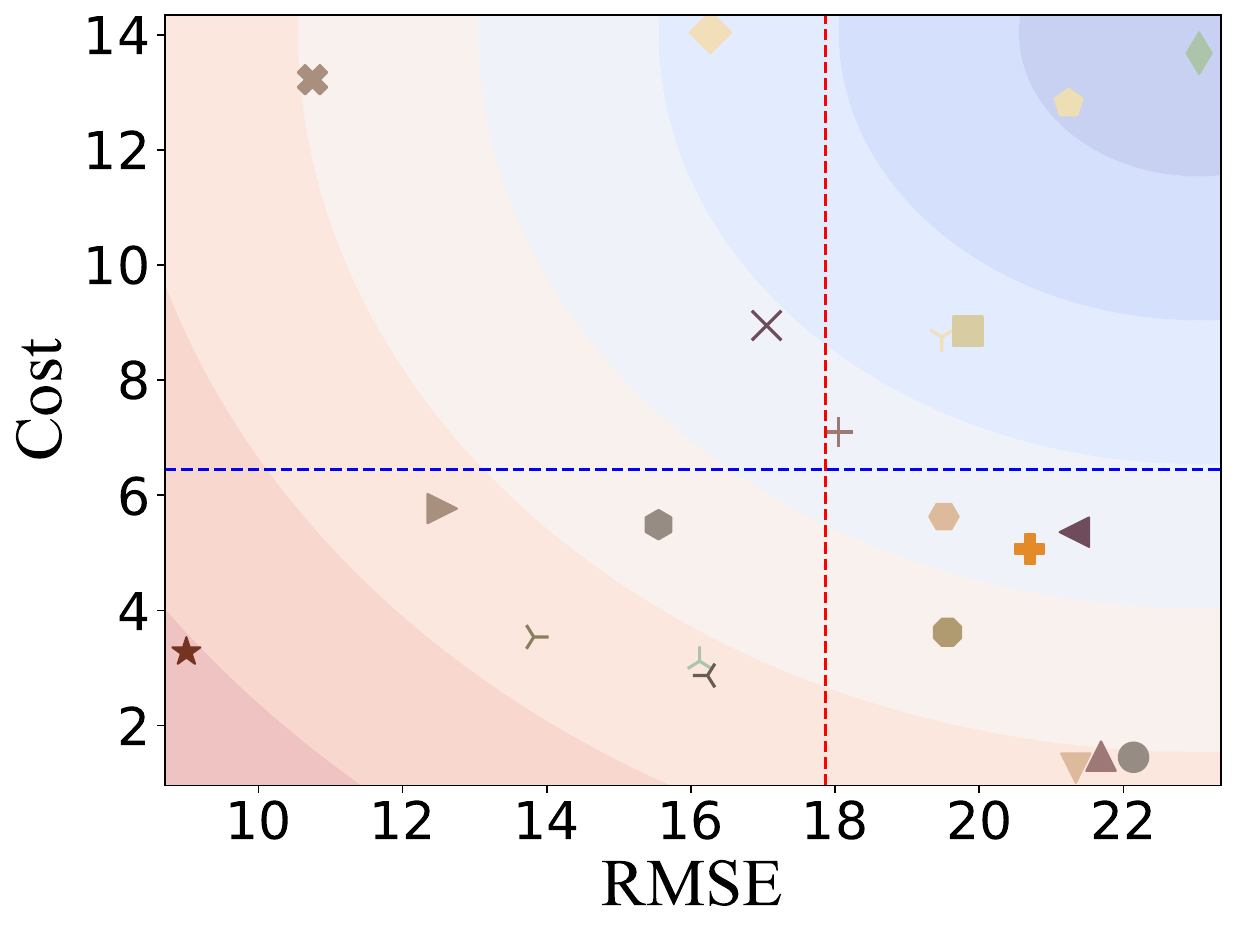}}
 \hfill
\caption{\label{fig:cost}Visual analysis of model performance based on training cost and prediction accuracy. The red and blue lines represent the mean values of the respective metrics.}
\vspace{-0.25cm}
\end{figure*}

To address the high dynamism, training time consumption is an essential metric to consider for practical model application. Hence, to facilitate a convenient comparison of the balance between efficacy and efficiency among models, we visualized the comparative results of all models on two datasets across three prediction scales in \textbf{Figure \ref{fig:cost}}. Models positioned closer to the bottom-left corner indicate a better balance between time and accuracy, whereas those in the top-right corner imply longer training times and greater prediction errors. The area constituted by the left side of the red line and the bottom of the blue line represents models that surpass the average in both training consumption and predictive accuracy, indicating superior comprehensive capabilities.

It is evident in all the results that GraphMLP consistently occupies a position closest to the bottom-left corner, demonstrating that its MLP architecture and parallel processing approach can achieve more accurate predictions with less training time consumption. Besides GraphMLP, models like ST-AFN, HGCN, and GCN-GRU also consistently fall within the bottom-left area, indicating their robust overall performance. This comparison also reveals that although spatio-temporal models based on graph structures perform well in predictions, most also entail higher time costs, particularly models like DCRNN and MegaCRN.

\subsection{Ablation Study}
Although our primary contribution does not lie in proposing a new model for lane-level traffic prediction, we deem it necessary to conduct a series of ablation studies to validate the impact of each component of the GraphMLP model on prediction accuracy and time consumption. \textbf{Table \ref{tab:Ablation}} presents the comparative results of the Instance Normalization module, the Dynamic Graph Network module, the Independent Temporal MLP module, and the complete model on both the PeMS and HuaNan datasets.

Overall, on both datasets, each component has a similar impact, with the omission of any module resulting in increased prediction error. The absence of the Independent Time MLP Network module leads to a larger error compared to the other two modules, indicating its central importance to the model. From the perspective of training consumption, the Dynamic Graph Network module occupies relatively more time. However, its parallel training with the Independent Time MLP Network module makes this acceptable in comparison to the reduced error rates.

\begin{table}[htbp]
\setlength\tabcolsep{3pt}
  \centering
   \vspace{-0.25cm}
  \caption{Ablation Study}
    \begin{tabular}{rllcrlc}
    \toprule
    \multicolumn{1}{l}{Datasets} & Variants &     &     & \multicolumn{1}{l}{Variants} &     &  \\
    \midrule
        & w/o & MAE & 5.15  & \multicolumn{1}{l}{w/o} & MAE & 4.45  \\
        & Instance & RMSE & 8.47  & \multicolumn{1}{l}{Dynamic} & RMSE & 7.13  \\
        & Normalization & MAPE & 19.61\% & \multicolumn{1}{l}{Graph} & MAPE & 17.64\% \\
    \multicolumn{1}{l}{PeMS} &     & Cost & 1.96  &     & Cost & 1.55  \\
\cmidrule{2-7}        & w/o & MAE & 5.22  & \multicolumn{1}{l}{w/o} & MAE & 4.05  \\
        & Indeoendent & RMSE & 8.59  & \multicolumn{1}{l}{None} & RMSE & 6.29  \\
        & Temporal & MAPE & 22.57\% &     & MAPE & 16.25\% \\
        & MLP & Cost & 1.90  &     & Cost & 2.05  \\
    \midrule
        & w/o & MAE & 5.02  & \multicolumn{1}{l}{w/o} & MAE & 4.80  \\
        & Instance & RMSE & 8.63  & \multicolumn{1}{l}{Dynamic} & RMSE & 7.31  \\
        & Normalization & MAPE & 20.80\% & \multicolumn{1}{l}{Graph} & MAPE & 19.23\% \\
    \multicolumn{1}{l}{HuaNan} &     & Cost & 2.91  &     & Cost & 2.37  \\
\cmidrule{2-7}        & w/o & MAE & 6.48  & \multicolumn{1}{l}{w/o} & MAE & 3.88  \\
        & Indeoendent & RMSE & 9.60  & \multicolumn{1}{l}{None} & RMSE & 7.77  \\
        & Temporal & MAPE & 25.16\% &     & MAPE & 17.03\% \\
        & MLP & Cost & 2.56  &     & Cost & 3.09  \\
    \bottomrule
    \end{tabular}%
  \label{tab:Ablation}%
  \vspace{-0.25cm}
\end{table}%

\subsection{Case Study}
\begin{figure}
 \vspace{-0.25cm}
 \captionsetup[subfloat]{labelfont=scriptsize,textfont=scriptsize}
 \centering
 \subfloat[Topological position of lane segment $l_{4,6}$]{
 \includegraphics[width=1.5in]{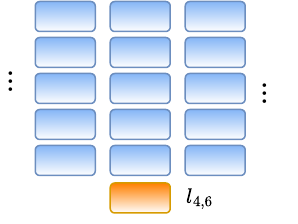}}  
 \subfloat[Prediction error of lane segment $l_{4,6}$]{
 \includegraphics[width=1.8in]{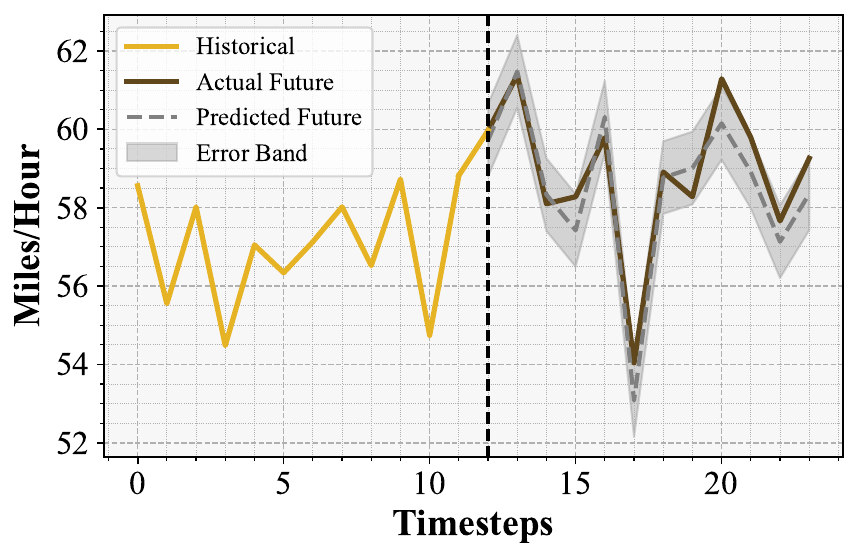}}  

 \caption{\label{fig: case} Case study on the topological structure and prediction visualization of lane-level traffic prediction on irregular roads.}
 \vspace{-0.35cm}
\end{figure}

To further demonstrate the characteristics of lane-level traffic prediction and intuitively evaluate the performance of the simple baseline model, GraphMLP in complex scenarios, we conduct a case study on a representative segment from the PeMSF dataset, which features irregular lane configurations. Specifically, we select the sixth lane segment $l_{4,6}$ under the fourth sensor and visualize its topological position and prediction results under a historical window of $T=12$ and a prediction horizon of $z=12$, as shown in \textbf{Figure~\ref{fig: case}}.

\textbf{Subfigure (a)} shows the topology around $l_{4,6}$, where entrance lanes and other asymmetric structures are clearly present. These irregular configurations are difficult to capture using grid-based models, whereas the graph-based approach flexibly represents lane connectivity, including dynamic structures such as entrance ramps and construction lanes, offering a more realistic and expressive modeling capability. \textbf{Subfigure (b)} compares the historical, actual future, and predicted speeds of the selected lane segment. GraphMLP effectively tracks traffic dynamics, with predictions closely matching the ground truth. The narrow error band suggests strong fitting ability and good generalization. This case highlights the practicality of graph-based representations for modeling irregular lane structures and demonstrates the adaptability of GraphMLP, as a baseline model, across diverse traffic scenarios.

\section{Conclusion and Future Work Discussion\label{section 6}}

In this paper, we conduct a detailed analysis of the unique challenges faced by lane-level traffic prediction compared to coarser-grained city and road-level traffic prediction and provide a comprehensive review and categorization of current research in the field. Based on this analysis, we define a unified spatial topology and prediction task for lane-level traffic prediction and introduce a concise and efficient baseline model, GraphMLP. This model employs adaptively generated dynamic graph convolution and patch-based independent temporal MLP networks for efficient prediction. For further empirical analysis, we offer three public datasets from two real road networks with both regular and irregular numbers of lanes. We have also replicated previously unpublished model codes and employed comprehensive evaluation metrics for a fair comparison of the accuracy and efficiency of existing models. Through this work, we aim to further advance the field of lane-level traffic prediction, providing richer perspectives and tools for traffic prediction tasks.

From the analysis presented in this paper, we believe current research still faces several limitations, such as the difficulty in modeling fine-grained and rapidly changing lane-level dynamics, limited generalizability across diverse road types, and the high computational cost of real-time inference. Therefore, future research in lane-level traffic prediction should focus on the following key areas to drive technological advancement and practical applications:

\textbf{Efficiency Design:} Future research should prioritize improving the efficiency of lane-level traffic prediction models, particularly in terms of training time and computational resource consumption. By optimizing model architectures, adopting more efficient training methods, or utilizing model compression techniques, the computational complexity of the models can be reduced, thereby enhancing their response speed and processing capacity in real-world applications. This would not only improve the performance of real-time traffic prediction but also enable lane-level prediction to be applied more broadly in intelligent transportation systems.

{\textbf{Road-Lane Interactions:}  As the complexity of traffic networks increases, future research needs to explore the interactions between different lanes in greater depth. In real-world traffic scenarios, changes in flow, traffic mobility, and lane-switching behaviors are closely interrelated. Therefore, effectively modeling these interactions at the lane level and integrating spatiotemporal information is key to improving prediction accuracy. Meanwhile, such modeling also opens opportunities for discovering latent traffic structures and bottlenecks through interpretable data mining. Future studies can explore closer road-lane interactions to further capture and model the complex dependencies between lanes.

\textbf{Multi-Granularity Collaboration:} Future lane-level traffic prediction should not only focus on predictions at a single granularity level but also consider the collaborative functioning of multiple granularities (e.g., road-level, region-level, and city-level). Multi-granularity models can leverage traffic information at different levels to achieve more comprehensive and accurate predictions. Especially in complex urban environments, the coordination between coarse-grained road-level predictions and fine-grained lane-level predictions will significantly enhance overall prediction performance.

\textbf{Large-Scale Traffic Models:}  Research on lane-level traffic prediction needs to better integrate into the development of large-scale traffic models, particularly in the context of analyzing large-scale road networks and long-duration datasets. As data volumes grow and model complexities increase, efficiently training and optimizing these large models will be an important direction for future research. Techniques such as pre-trained models and transfer learning can improve model generalization, reduce dependence on vast amounts of labeled data, and enhance the predictive capabilities of large-scale intelligent transportation systems.

Lane-level traffic prediction is a more complex and detailed task. Future research should focus not only on model innovation and performance optimization but also on practical deployment in real-world traffic systems to improve efficiency, safety, and adaptabilit, such as providing lane-level guidance for autonomous vehicles, supporting dynamic lane control policies like reversible (tidal) lanes, enabling proactive congestion mitigation at fine granularity, and assisting urban traffic signal optimization. Integrating external factors like traffic signals and lane-changing behavior will be crucial for advancing intelligent traffic management, enabling more accurate predictions, optimizing traffic flow, and supporting autonomous driving systems.


\newpage

\bibliographystyle{IEEEtran}
\bibliography{ref}
\vspace{-1cm}
\begin{IEEEbiography}[{\includegraphics[width=1in,height=1.25in,clip,keepaspectratio]{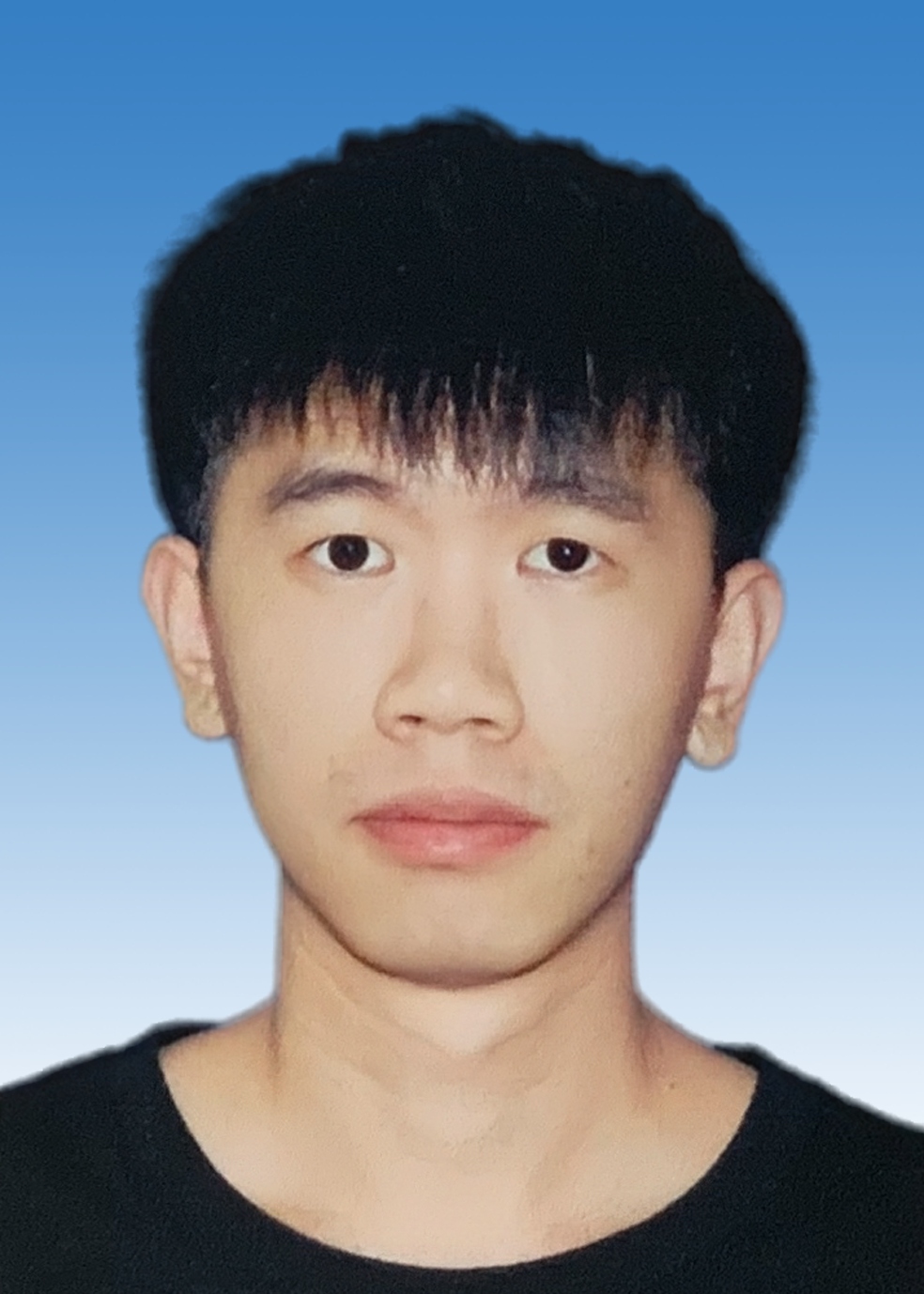}}]{Shuhao Li}
(Student Member, IEEE) is currently pursuing a Ph.D. degree at the School of Computer Science, Fudan University, Shanghai, China. He obtained his Master's degree in Computer Technology from Guangzhou University, Guangzhou, China, in 2023. His research interests include intelligent transportation systems, spatio-temporal data mining, anomaly detection, and related areas.
\end{IEEEbiography}
\begin{IEEEbiography}[{\includegraphics[width=1in,height=1.25in,clip,keepaspectratio]{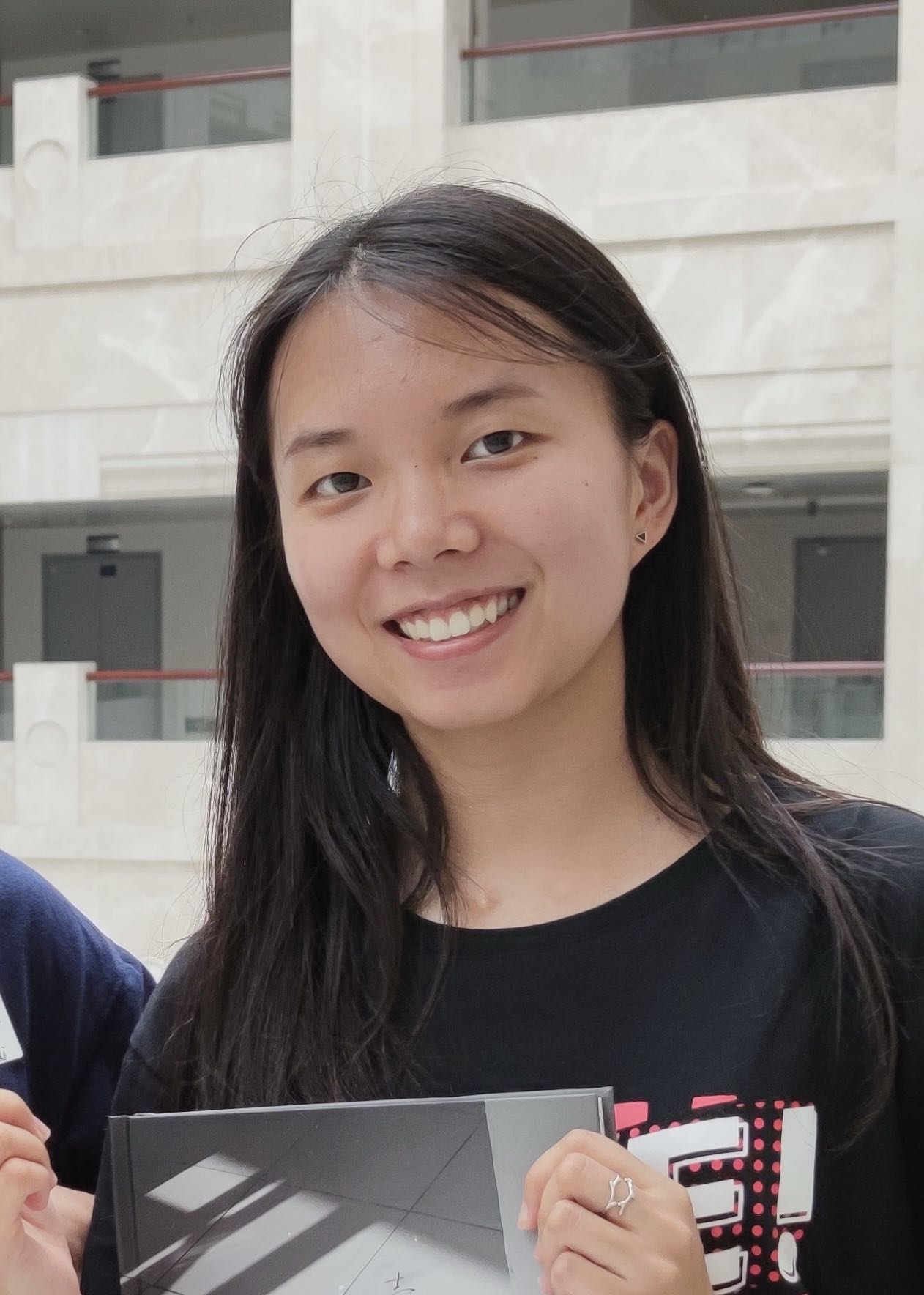}}]{Yue Cui}
received a Bachelor’s degree from the University of Electronic Science and Technology of China in 2020. She is a Ph.D. candidate at the Hong Kong University of Science and Technology. She is interested in developing effective and trustworthy algorithms for data science and artificial intelligence. She has published over 15 peer-reviewed papers in top-tier conferences and journals in the area of data mining and trustworthy AI, including KDD, WWW, VLDB, ICDE, TOIS, etc.
\end{IEEEbiography}

\begin{IEEEbiography}[{\includegraphics[width=1in,height=1.25in,clip,keepaspectratio]{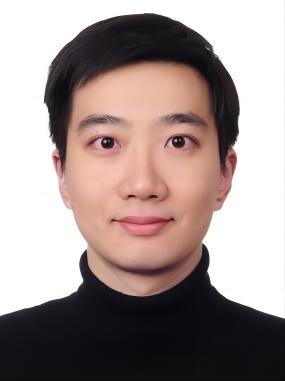}}]{Jingyi Xu}
received the M.Sc. degree in Data Science from the University of Exeter, UK, in 2020.
He is currently a second-year Ph.D. student at Fudan University. His research interests include artificial intelligence and data mining. His current research area is related to the deep learning approach for time series analysis and focuses on the application of both supervised and unsupervised paradigms.
\end{IEEEbiography}

\begin{IEEEbiography}[{\includegraphics[width=1in,height=1.25in,clip,keepaspectratio]{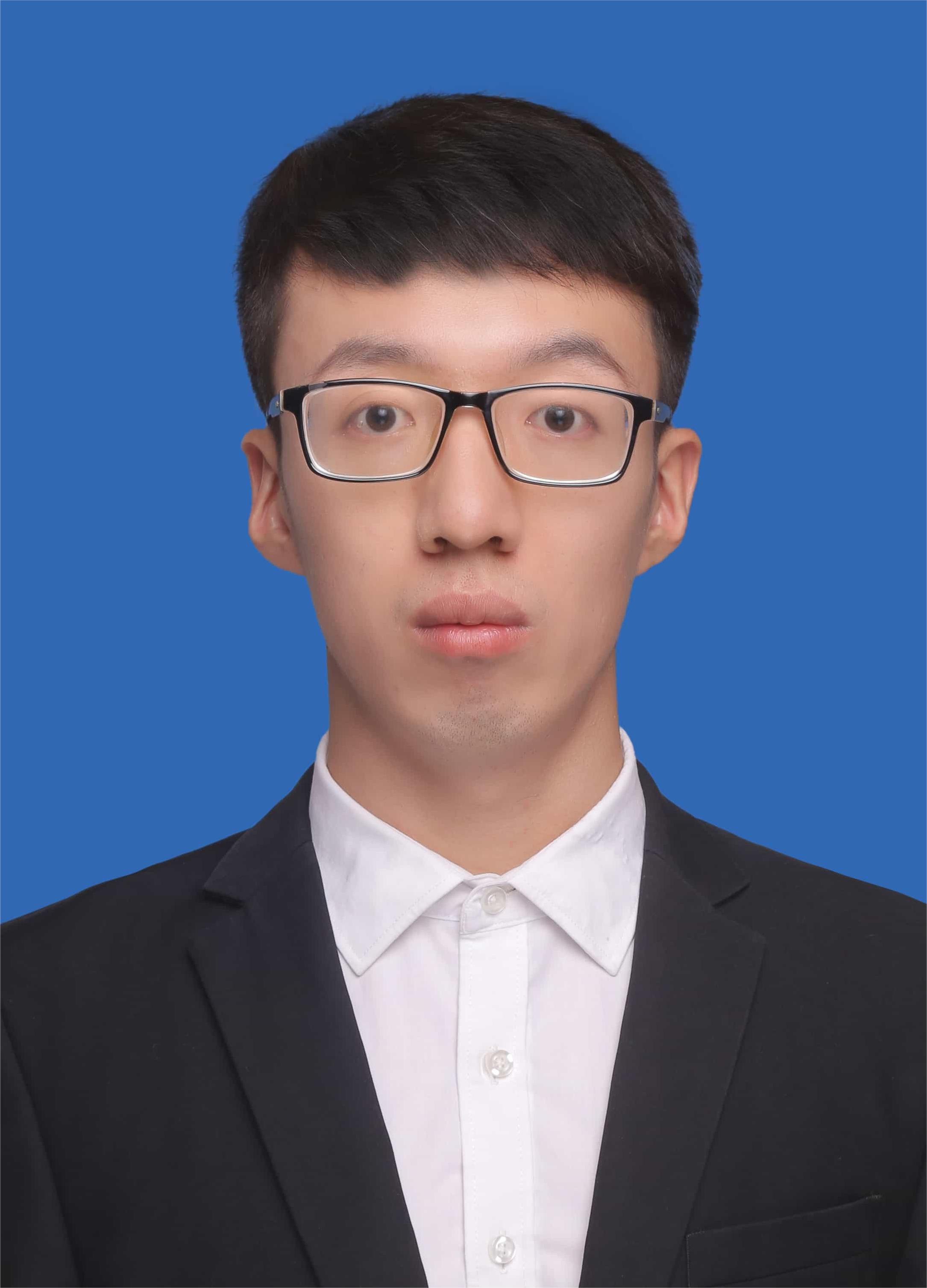}}]{Libin Li}
received the B.E. degree from Northeastern University, Shenyang, China, in 2022. He is currently pursuing an M.S. degree in Computer Technology from Guangzhou University, Guangzhou, China. His research interests include spatio-temporal data mining, traffic network analysis, and intelligent transportation systems.
\end{IEEEbiography}
\begin{IEEEbiography}[{\includegraphics[width=1in,height=1.25in,clip,keepaspectratio]{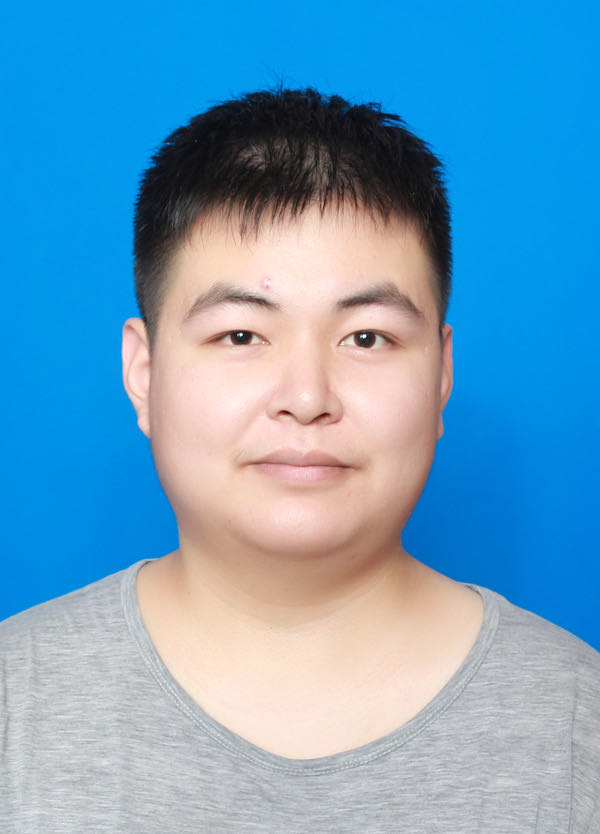}}]{Lingkai Meng}
is currently pursuing a Ph.D. degree at Antai College of Economics and Management, Shanghai Jiao Tong University, Shanghai, China. He obtained his Master's degree in Computer Technology from Guangzhou University, Guangzhou, China, in 2023. His research interests include graph data mining, community search, structural clustering, and data compression.
\end{IEEEbiography}

\begin{IEEEbiography}[{\includegraphics[width=1in,height=1.25in,clip,keepaspectratio]{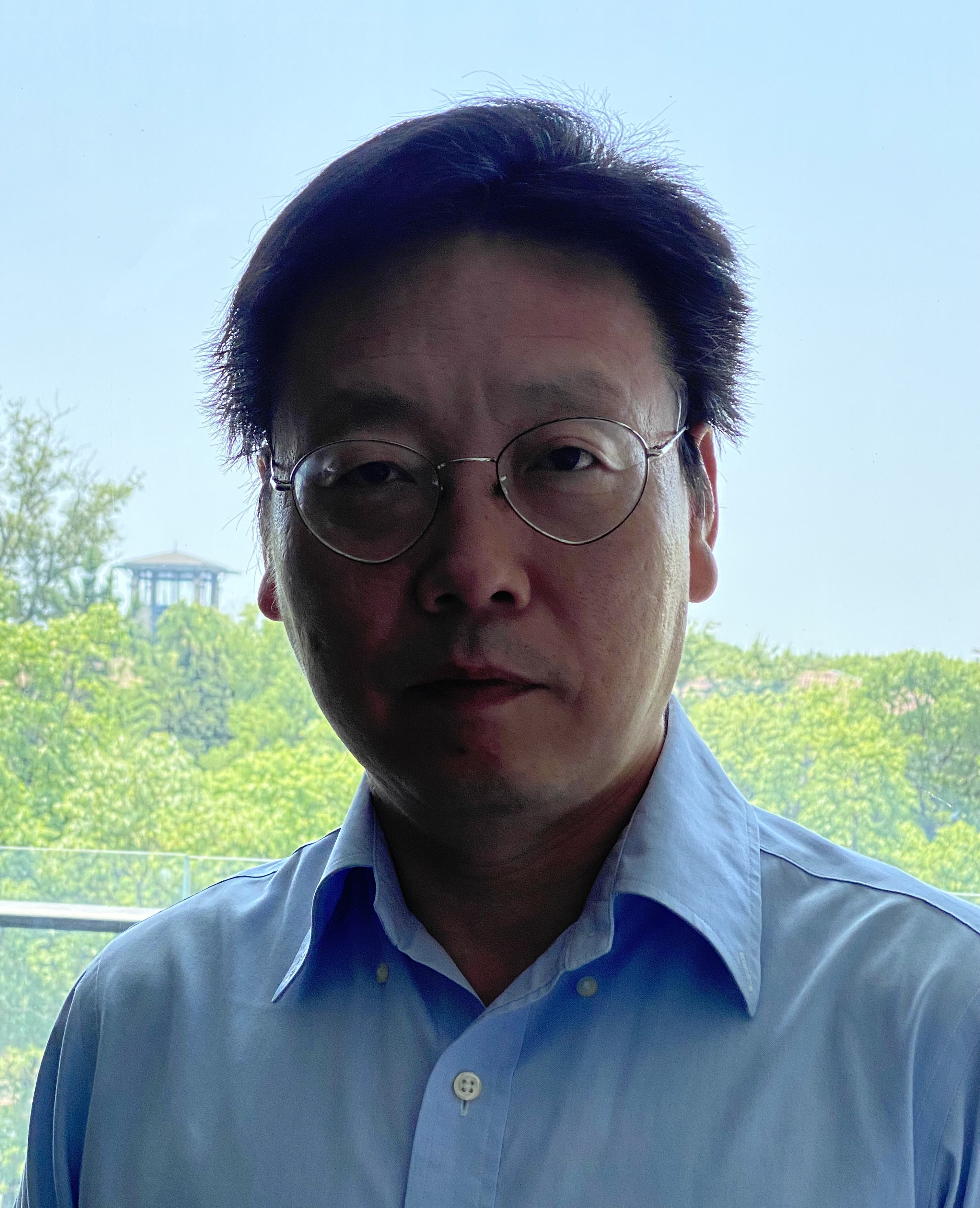}}]{Weidong Yang}
received a Ph.D. degree in software engineering from Xidian University in 1999. He was a Post-Doctoral Researcher from the School of Computer Science, Fudan University, Shanghai, China, from 1999 to 2001. He is a Professor with the School of Computer Science, Fudan University. His research interests include big data, knowledge engineering, database and data mining, and software engineering.
\end{IEEEbiography}

\begin{IEEEbiography}[{\includegraphics[width=1in,height=1.25in,clip,keepaspectratio]{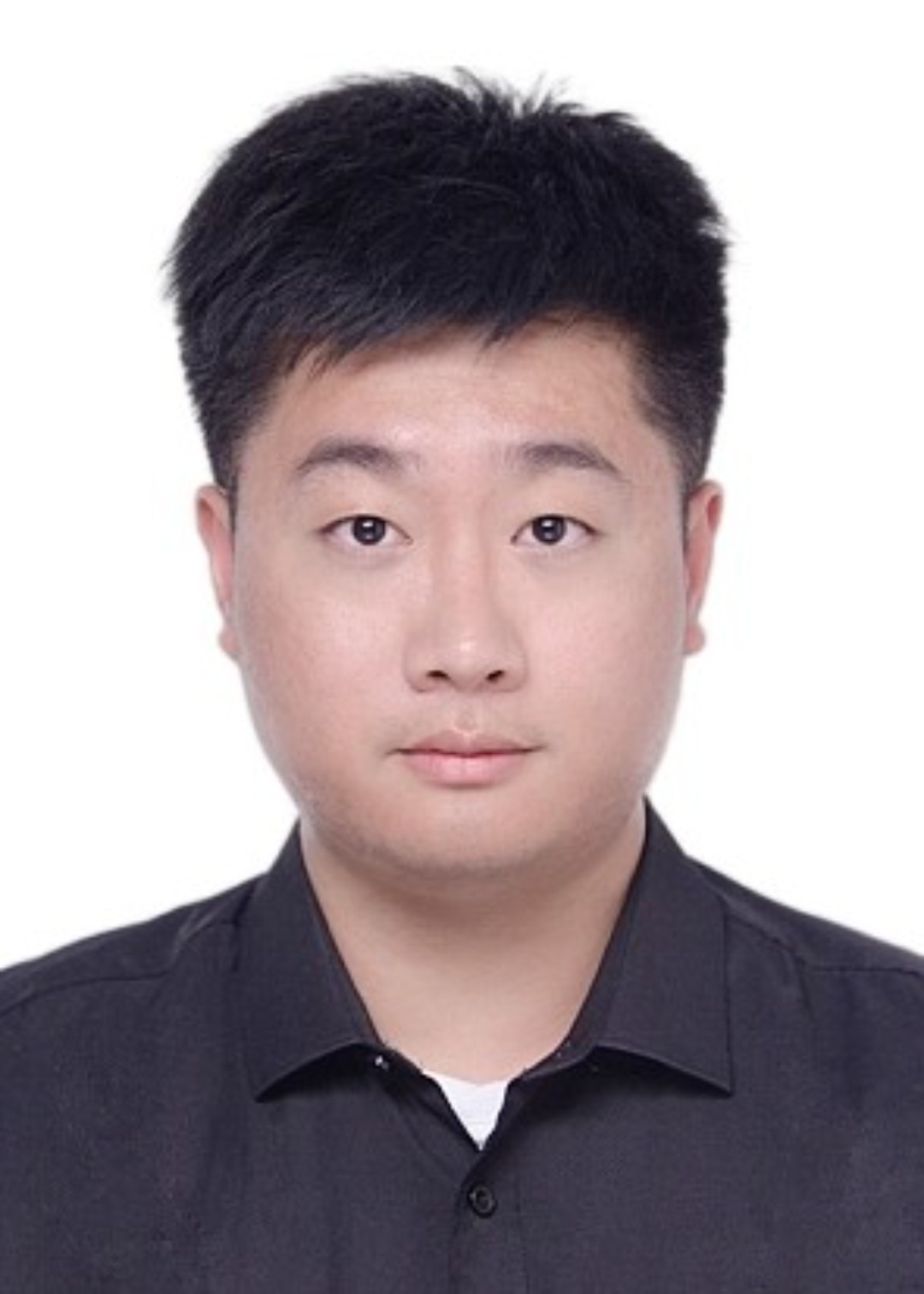}}]{Fan Zhang} is a Professor at Guangzhou University. He was a research associate at the University of New South Wales. He received the BEng degree from Zhejiang University in 2014, and the PhD from the University of Technology Sydney in 2018. His research interests include graph management and analysis. Since 2017, he has published more than 20 papers in top venues as the first/corresponding author, e.g., SIGMOD, KDD, PVLDB, TheWebConf, ICDE, TKDE, and VLDB Journal. His academic homepage is https://fanzhangcs.github.io/.
\end{IEEEbiography}

\begin{IEEEbiography}[{\includegraphics[width=1in,height=1.25in,clip,keepaspectratio]{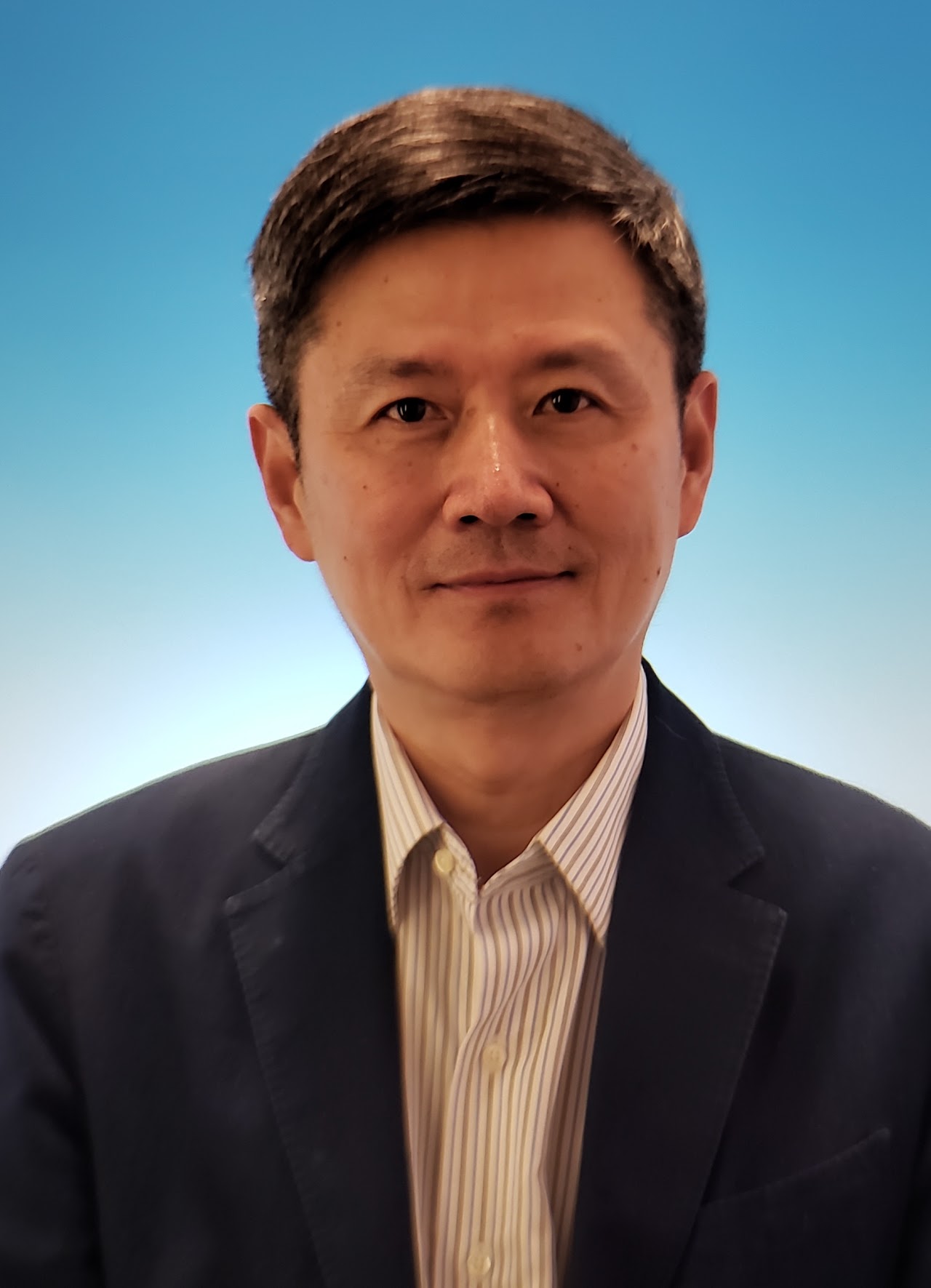}}]{Xiaofang Zhou}
(Fellow, IEEE) is Otto Poon Professor of Engineering and Chair Professor of Computer Science and Engineering at the Hong Kong University of Science and Technology. He received his Bachelor's and Master's degrees in Computer Science from Nanjing University, in 1984 and 1987 respectively, and his PhD degree in Computer Science from the University of Queensland in 1994. His research is focused on finding effective and efficient solutions for managing, integrating, and analyzing very large amounts of complex data for business, scientific and personal applications. His research interests include spatial and multimedia databases, high-performance query processing, web information systems, data mining, data quality management, and machine learning. He is a Fellow of IEEE.
\end{IEEEbiography}


\end{document}